\documentclass[10pt,twocolumn,letterpaper]{article} 

\usepackage{wacv} 

\makeatletter
\@namedef{ver@everyshi.sty}{}
\makeatother

\usepackage{times}
\usepackage{epsfig}
\usepackage{graphicx}
\usepackage{amsmath}
\usepackage{amssymb}
\usepackage{extarrows}
\usepackage{algorithm}
\usepackage{algpseudocode}
\usepackage{float}

\usepackage{bm}
\usepackage{subcaption}
\usepackage{todonotes}
\usepackage{tcolorbox}
\usepackage{tikz}
\usepackage{tikz-3dplot}
\usepackage{tikzscale}
\usetikzlibrary{arrows,shapes,chains,matrix,positioning,spy}
\usetikzlibrary{scopes,decorations,shadows,backgrounds,fit}
\usetikzlibrary{decorations.pathreplacing,calc,3d,quotes,decorations.text}
\usetikzlibrary{positioning}
\usepackage{pgfmath}
\usepackage{threeparttable}
\usepackage{paralist}            

\usepackage{array}
\usepackage{mdwmath}
\usepackage{mdwtab}
\usepackage{eqparbox}
\usepackage{fixltx2e}
\usepackage{stfloats}
\usepackage{url}
\usepackage{diagbox}
\usepackage{verbatim}
\usepackage{booktabs}
\usepackage{multirow}
\usepackage{mathtools}
\usepackage{breqn}
\usepackage{cite}
\usepackage{appendix}
\usepackage{pifont}

\def\wacvPaperID{547}
\wacvfinalcopy 

\def\etal{{\it et al. }}
\newcommand{\floor}[1]{\lfloor #1 \rfloor}
\newcommand{\ceil}[1]{\lceil #1 \rceil}

\newcommand\numberthis{\addtocounter{equation}{1}\tag{\theequation}}
\def\versus{{\it vs. }}
\newcommand{\ourmethod}{3DFaceFill}

\graphicspath{{figs/}}

\newcommand{\xmark}{\ding{55}}

\makeatletter
\def\thickhline{%
  \noalign{\ifnum0=`}\fi\hrule \@height \thickarrayrulewidth \futurelet
   \reserved@a\@xthickhline}
\def\@xthickhline{\ifx\reserved@a\thickhline
               \vskip\doublerulesep
               \vskip-\thickarrayrulewidth
             \fi
      \ifnum0=`{\fi}}
\makeatother

\newlength{\thickarrayrulewidth}
\ifwacvfinal
\fi

\ifwacvfinal
\usepackage[breaklinks=true,bookmarks=false]{hyperref}
\else
\usepackage[pagebackref=true,breaklinks=true,colorlinks,bookmarks=false]{hyperref}
\fi
\usepackage[capitalise]{cleveref}

\begin{document}

\title{\ourmethod: An Analysis-By-Synthesis Approach to Face Completion}

\author{Rahul Dey \qquad Vishnu Naresh Boddeti\\
Michigan State University\\
{\tt\small \{deyrahul, vishnu\}@msu.edu}
}

\setlength{\floatsep}{10pt plus 2.0pt minus 2.0pt}

\twocolumn[{
\renewcommand\twocolumn[1][]{#1}%
\maketitle
\begin{center}
    \vspace{-1em}
    \centering
    \captionsetup{font=small}
    \begin{minipage}{\textwidth}
    \begin{tikzpicture}
    \node (a0) {A:};
    \node[right of=a0, node distance=1.3cm] (a1) {\includegraphics[width=.113\linewidth]{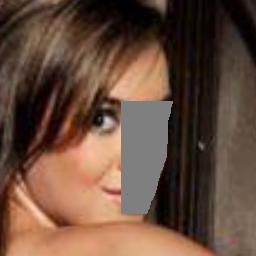}};
    \node[right of=a1, node distance=2.05cm] (a2) {\includegraphics[width=.113\linewidth]{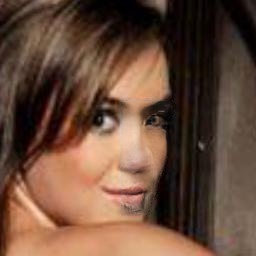}};
    \node[right of=a2, node distance=2.05cm] (a3) {\includegraphics[width=.113\linewidth]{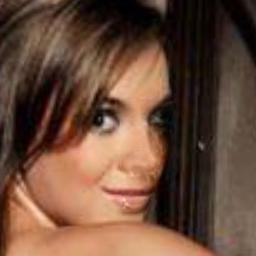}};
    \node[right of=a3, node distance=2.05cm] (a4) {\includegraphics[width=.113\linewidth]{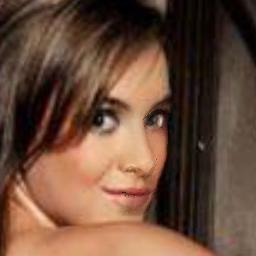}};
    
    \node[right of=a4, node distance=1.3cm] (b0) {B:};
    \node[right of=b0, node distance=1.3cm] (b1) {\includegraphics[width=.113\linewidth]{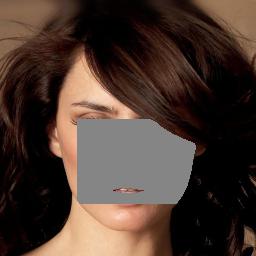}};
    \node[right of=b1, node distance=2.05cm] (b2) {\includegraphics[width=.113\linewidth]{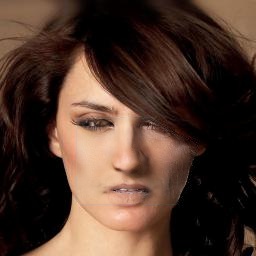}};
    \node[right of=b2, node distance=2.05cm] (b3) {\includegraphics[width=.113\linewidth]{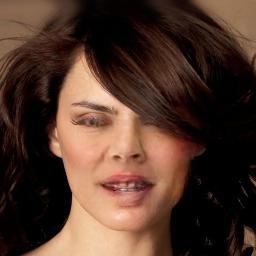}};
    \node[right of=b3, node distance=2.05cm] (b4) {\includegraphics[width=.113\linewidth]{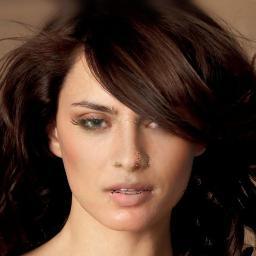}};
    
    \node[below of=a0, node distance=2.1cm] (c0) {C:};
    \node[right of=c0, node distance=1.3cm] (c1) {\includegraphics[width=.113\linewidth]{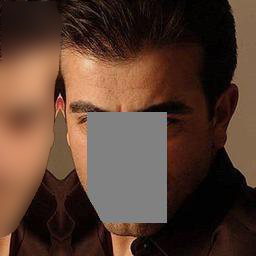}};
    \node[right of=c1, node distance=2.05cm] (c2) {\includegraphics[width=.113\linewidth]{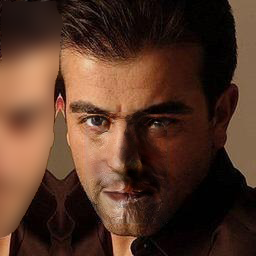}};
    \node[right of=c2, node distance=2.05cm] (c3) {\includegraphics[width=.113\linewidth]{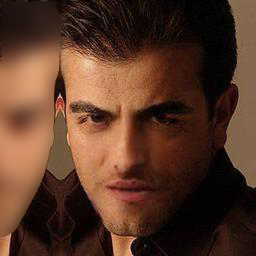}};
    \node[right of=c3, node distance=2.05cm] (c4) {\includegraphics[width=.113\linewidth]{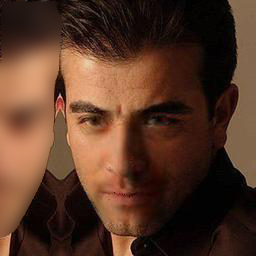}};
    \node[below of=c1, node distance=1.3cm] (c1_text) {\small (a) Input};
    \node[below of=c2, node distance=1.3cm] (c2_text) {\small (b) DeepFillv2};
    \node[below of=c3, node distance=1.3cm] (c3_text) {\small (c) PICNet};
    \node[below of=c4, node distance=1.3cm] (c4_text) {\small (d) \ourmethod};
    
    \node[right of=c4, node distance=1.3cm] (d0) {D:};
    \node[right of=d0, node distance=1.3cm] (d1) {\includegraphics[width=.113\linewidth]{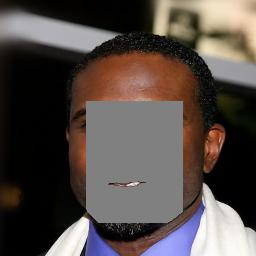}};
    \node[right of=d1, node distance=2.05cm] (d2) {\includegraphics[width=.113\linewidth]{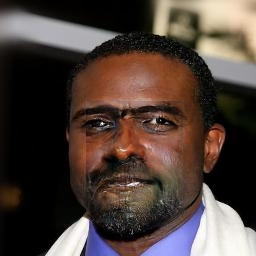}};
    \node[right of=d2, node distance=2.05cm] (d3) {\includegraphics[width=.113\linewidth]{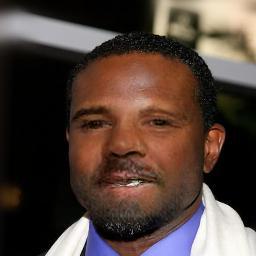}};
    \node[right of=d3, node distance=2.05cm] (d4) {\includegraphics[width=.113\linewidth]{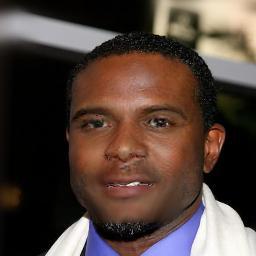}};
    \node[below of=d1, node distance=1.3cm] (d1_text) {\small (a) Input};
    \node[below of=d2, node distance=1.3cm] (d2_text) {\small (b) DeepFillv2};
    \node[below of=d3, node distance=1.3cm] (d3_text) {\small (c) PICNet};
    \node[below of=d4, node distance=1.3cm] (d4_text) {\small (d) \ourmethod};
    \end{tikzpicture}
    \end{minipage}
    \vspace{-0.8em}
    \captionsetup{type=figure}\captionof{figure}{\textbf{Inpainting of face images under diverse conditions by \ourmethod{} and existing approaches}. By modeling the image formation process \ourmethod{} is able to generate more geometrically consistent and photorealistic completions across diverse scenarios such as non-frontal poses (A), light and dark complexions (B,D), non-uniform facial illumination (\eg illumination is different on two sides of the nose in C) and in cases where the baselines tend to distort face components (\eg nose in B).\label{fig:teaser}}
\end{center}
}]
\thispagestyle{empty}

\begin{abstract}
\vspace{-0.5em}



Existing face completion solutions are primarily driven by end-to-end models that directly generate 2D completions of 2D masked faces. By having to implicitly account for geometric and photometric variations in facial shape and appearance, such approaches result in unrealistic completions, especially under large variations in pose, shape, illumination and mask sizes. To alleviate these limitations, we introduce \ourmethod, an analysis-by-synthesis approach for face completion that explicitly considers the image formation process. It comprises three components, (1) an encoder that disentangles the face into its constituent 3D mesh, 3D pose, illumination and albedo factors, (2) an autoencoder that inpaints the UV representation of facial albedo, and (3) a renderer that resynthesizes the completed face. By operating on the UV representation, \ourmethod{} affords the power of correspondence and allows us to naturally enforce geometrical priors (\eg facial symmetry) more effectively. Quantitatively, \ourmethod{} improves the state-of-the-art by up to 4dB higher PSNR and ~25\% better LPIPS for large masks. And, qualitatively, it leads to demonstrably more photorealistic face completions over a range of masks and occlusions while preserving consistency in global and component-wise shape, pose, illumination and eye-gaze.

\end{abstract}

\section{Introduction} \label{sec:introduction}
End-to-end image completion methods i.e., models that generate 2D completions directly from 2D masked images, have witnessed remarkable progress in recent years. These approaches rely primarily on architectural advances in neural network designs to implicitly account for photometric and geometric variations in image appearance. And even those that explicitly include scene geometry in their formulation do so largely in 2D. Consequently, object-based image completions from such methods often suffer from poor photorealism, especially under large variations in pose, shape, illumination of objects in the image and the inpainting mask. For example, in the context of faces, Fig. \ref{fig:teaser} shows face images having extreme poses (\ref{fig:teaser}.A), illumination variations across the face (\ref{fig:teaser}.C) and diverse appearances and shapes. Current state-of-the-art methods such as DeepFillv2 \cite{freeforminpainting} and PICNet \cite{zheng2019pluralistic}, both of which operate end-to-end on 2D image representations, often fail in preserving facial symmetry and the variations of the aforementioned factors (pose, illumination, texture, shape) while inpainting.

\begin{figure}[t]

\tikzstyle{block} = [rectangle, draw, fill=blue!20, text centered]
\centering
\resizebox{\columnwidth}{!}{
\begin{tikzpicture}[node distance=3cm, auto,>=latex', thick]

\node[label={[anchor=north west,text=white]north west:{$\mathbf{I}_m$}}] (masked_face) {\includegraphics[width=.16\linewidth]{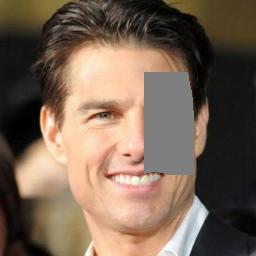}};
\node[below of=masked_face, node distance=1.5cm, block, draw=none, fill=white, minimum width=1cm, minimum height=0.5cm] (mask) {Mask};

\node[above of=masked_face, node distance=2cm, label={[anchor=north west,text=white]north west:{$\mathbf{M}_f$}}] (face_mask) {\includegraphics[width=.16\linewidth]{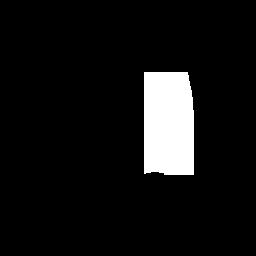}};

\node[right of=face_mask, node distance=2.4cm, block, fill=lightgray, fill opacity=0.5, text opacity=1, minimum height=0.5cm, minimum width=0.2cm, inner xsep=0pt] (segmenter) {\begin{tabular}{c}Face\\Segmenter\end{tabular}};

\node[right of=segmenter, node distance=2.4cm, label={[anchor=north west,text=white]north west:{$\mathbf{I}$}}] (orig) {\includegraphics[width=.16\linewidth]{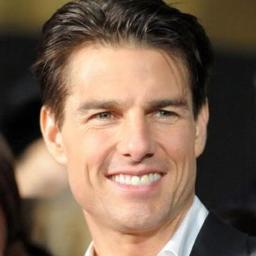}};
\node[above of=orig, node distance=0.9cm] (orig_text) {Original};

\draw[dashed,->] (orig) -- (segmenter);
\draw[dashed,->] (segmenter) -- (face_mask);
\draw[dashed,->] (face_mask) -- (masked_face);
\draw[->] (mask) -- (masked_face);

\node[right of=masked_face, node distance=2.6cm, block, fill=blue!60, fill opacity=0.5, text opacity=1, minimum width=0.5cm, minimum height=0.5cm, inner xsep=0pt] (3dmm) {\begin{tabular}{c}3DMM\end{tabular}};
\draw[->] (masked_face) -- node [midway,above] {Iter 1} (3dmm);

\node[right of=3dmm, node distance=2.4cm, block, fill=pink, fill opacity=0.5, text opacity=1, minimum width=0.5cm, minimum height=0.5cm, inner xsep=0pt] (inpainter) {\begin{tabular}{c}ALBEDO\\INPAINTER\end{tabular}};
\draw[->] (3dmm) -- (inpainter);

\node[right of=inpainter, node distance=2.7cm, block, fill={rgb:red,1;green,4;blue,3}, fill opacity=0.5, text opacity=1, minimum width=0.5cm, minimum height=1cm] (renderer) {RENDERER};
\draw[->] (inpainter) -- (renderer);

\node[right of=renderer, node distance=2.5cm, label={[anchor=north west,text=white]north west:{$\hat{\mathbf{I}}$}}] (completed) {\includegraphics[width=.16\linewidth]{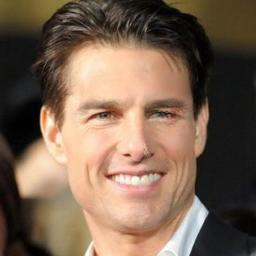}};
\node[below of=completed, node distance=0.9cm] (out_text) {Output};
\draw[->] (renderer) -- (completed);

\coordinate[right of=renderer, node distance=1.3cm] (renderer_right);
\coordinate[below of=renderer_right, node distance=1.4cm] (renderer_above);
\draw[] (renderer_right) -- (renderer_above);
\draw[->] (renderer_above) -| node[pos=0.25,below] {Iter $> 1$} (3dmm);

\end{tikzpicture}}
\caption{\label{fig:overview} \textbf{Overview:} \ourmethod{} is an iterative inpainting approach where the masked face is disentangled into its 3D shape, pose, illumination and partial albedo by the 3DMM module, following which the partial albedo is inpainted and finally the completed image is rendered. During inference (only), the completed image is fed back through the whole pipeline in subsequent iterations, while using the initial mask for albedo inpainting. During training, a pre-trained model segments the image into face, hair and background for constraining the mask to lie only on the face. This segmentation is optionally used during inference if necessary.}
\end{figure}
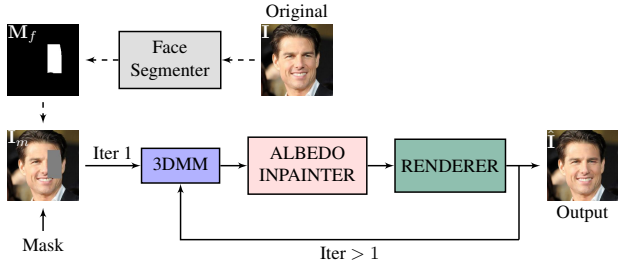

Several attempts have been made to customize generic image inpainting solutions for structured objects such as faces. General image inpainting approaches typically employ a CNN autoencoder as the inpainter and train it using a combination of photometric and adversarial losses \cite{contextencoderpathak, iizukagloballylocally, gencontextualattn, zheng2019pluralistic}. Face specific completion methods \cite{genfacecompletion, geometryawarefacecomp} employ additional losses such as landmark loss, perceptual loss and face parsing loss. However, these approaches still do not account for all factors in the image formation process like illumination and pose variations and as such fail to effectively impose geometric priors such as facial symmetry. Moreover, the implicit enforcement of geometric priors is still done in 2D as opposed to in 3D. This is a significant limitation as faces are inherently symmetric 3D objects and their projections on 2D images are often affected by the aforementioned factors of pose, illumination, shape \etc.

In contrast to the foregoing, this paper advocates for an analysis-by-synthesis approach for face completion that explicitly accounts for the 3D structure of faces i.e., shape and albedo, and image formation factors i.e., pose and illumination. The key insight of our solution is that performing face completion on the UV representation, as opposed to the 2D pixel representation, allows us to effectively leverage the power of correspondence and ultimately lead to geometrically and photometrically accurate face completion (see Fig.\ref{fig:teaser}). Our approach (see Fig. \ref{fig:overview}), dubbed \ourmethod{}, comprises of three components that are iteratively executed. First, the masked face image is disentangled into its constituent geometric and photometric factors. Second, an autoencoder performs inpainting on the UV representation of facial albedo. Lastly, the completed face is re-synthesized by a differentiable renderer. Our specific contributions are:

\vspace{2pt}
\noindent\textbf{--} We propose \ourmethod{}, a simple yet very effective face completion model that explicitly disentangles photometric and geometric factors and perform inpainting in the UV representation of facial albedo while preserving the associated facial shape, pose and illumination.

\vspace{2pt}    
\noindent\textbf{--} We propose a 3D symmetry-aware network architecture and a symmetry loss for the inpainter to propagate albedo features from the visible to symmetric masked regions of the UV representation. Enforcing the symmetry prior in 3D, as opposed to 2D, allows \ourmethod{} to more effectively leverage and preserve facial \emph{symmetry} while inpainting.
    
\vspace{2pt}
\noindent\textbf{--} Given our trained model, we propose a simple refinement process at inference by \emph{iteratively} reprocessing the face completion through the model. This process enables us to address the ``chicken-and-egg" problem of simultaneously inferring both the photometric and geometric factors and completion of the face from a masked image. The procedure is especially effective for heavily masked faces, improving the PSNR by up to 1dB.
    
\vspace{2pt}
\noindent\textbf{--} Extensive benchmarking on several datasets and unconstrained in-the-wild images results in \ourmethod{} producing photorealistic and geometrically consistent face completions over a range of masks and real occlusions, especially in terms of pose, lighting, and attributes such as eye-gaze and shape of nose along with a quantitative improvement of upto 4dB PSNR and 25\% in LPIPS\cite{lpips}.

\section{Related Work}

\noindent\textbf{Image Inpainting:} Earlier image inpainting approaches\cite{imageinpaintingbertalmio, criminisi2004region, patchmatchbarnes, hays2007scene} used diffusion or patch based methods to fill in the missing regions. This produced sharp results but often lacked semantic consistency. Recent techniques employ a CNN autoencoder along with a GAN loss to generate semantically consistent and realistic completions \cite{contextencoderpathak, semanticimageinpainting, iizukagloballylocally}. More recent methods focus on architectural enhancements to improve inpainting for variable and free form masks. These include a more refined discriminator in PatchGAN \cite{pix2pix}, contextual attention by DeepFill \cite{gencontextualattn} and gated convolutions \cite{partialconv, freeforminpainting}. In contrast, we adopt vanilla CNN architectures and instead rely on a more accurate analysis-by-synthesis modeling approach. Recently, Zheng \etal \cite{zheng2019pluralistic} generated multiple completions by sampling from a conditional distribution. Though this is a topic of interest, it is orthogonal to the goals of the current paper.

\vspace{5pt}
\noindent\textbf{Face Completion:} Face completion is a more challenging variant of image completion because of the complexity and diversity of faces. To address this, many approaches impose additional geometric and photometric priors in the form of face related losses \cite{geometryawarefacecomp, genfacecompletion, faceinpaintingan, zhang2017demeshnet, li2020learning, 3dmmganfacecomp}. A recent approach called DSA \cite{oracleattention} uses oracle-learned attention maps and component-wise discriminators to generate high-fidelity completions. While it often generates photorealistic completions in well-lit frontal faces, it still relies on implicitly learned priors which are insufficient to enforce correct geometry in challenging poses and illuminations. All these approaches rely on novel architectural advances and loss functions while \ourmethod{} focuses on more explicit and precise modeling of the image-formation process.

Concurrently, Deng \etal \cite{uvgan} completed self-occluded UV texture to synthesize new face views. This assumes that the full face image and at least half of the UV texture is always visible. In contrast, we go beyond self-occlusion and instead, perform 3D factorization on the masked face and complete its \textit{albedo} for \emph{masked face completion}. Furthermore, since texture is not always symmetric due to illumination variations, \cite{uvgan} needs synthetically completed texture maps for training; whereas our model performs completion on albedo which is further disentangled from both geometry as well as illumination allowing us to effectively enforce symmetry prior, without needing synthetically completed UV-maps for training, as it bears out in our experiments. A few recent works have also attempted to leverage symmetry for face completion \cite{zhang2018symmetry, li2020symmetry}. However, these approaches employ complex symmetry registration operations, which require huge computational resources; moreover these operations are often susceptible to large geometric variations.

\section{Approach}

\begin{figure*}[t]
\centering

\begin{subfigure}{0.73\textwidth}
\centering

\tikzstyle{format} = [draw, thin, fill=blue!20]
\tikzstyle{decision} = [diamond, draw, fill=blue!20, text width=4.5em, text badly centered, node distance=2cm, inner sep=0pt]
\tikzstyle{block} = [rectangle, draw, fill=blue!20, text centered]
\tikzstyle{disc_layer} = [block, opacity=.5, rotate=90, minimum height=4pt, inner sep=0, fill=teal!60]
\tikzstyle{line} = [draw, -latex']
\tikzstyle{circ} = [draw,circle,inner sep=0pt]
\tikzstyle{trap} = [trapezium, draw, fill=green!20, text centered]
\tikzstyle{trap_nob} = [trapezium, fill=green!20, trapezium angle=68]

\centering
\begin{tikzpicture}[node distance=3cm, auto,>=latex', thick]

\node[label={[anchor=north west,text=white]north west:{$\mathbf{I}_m$}}] (inp1) {\includegraphics[width=.08\linewidth]{figs/presentation/tom_cruise/input.jpg}};
\node[above of=inp1, node distance=0.9cm] (inp1_text) {\footnotesize Input};

\node[right of=inp1, node distance=2.0cm, shape border rotate=270, trap, fill=blue!70!green!50!white, minimum height=0.6cm] (enc1) {$\mathbf{E}$};
\draw[->] (inp1) -- node [pos=0.4,above] {\scriptsize Iter 1} (enc1);

\coordinate[right of=enc1, node distance=0.6cm] (enc1_break) {};
\node[right of=enc1_break, block, rotate=90, fill=red, opacity=.5, node distance=0.5cm, minimum width=0.75cm, minimum height=5pt, inner sep=0] (pose) {};
\node[above right of=pose, node distance=0.4cm] (pose_text) {$\mathbf{p}$};
\node[above of=pose, block, rotate=90, opacity=.5, node distance=.88cm, minimum width=1cm, minimum height=5pt, inner sep=0] (fs) {};
\node[above right of=fs, node distance=0.4cm] (fs_text) {$\mathbf{f_S}$};
\node[below of=pose, block, rotate=90, fill=red!20, opacity=.5, node distance=0.73cm, minimum width=0.75cm, minimum height=5pt, inner sep=0]
(l) {};
\node[above right of=l, node distance=0.4cm] (l_text) {$\mathbf{l}$};

\draw[->] (enc1) -- (enc1_break) |- (fs);
\draw[->] (enc1) -- (enc1_break) |- (pose);
\draw[->] (enc1) -- (enc1_break) |- (l);

\node[right of=pose, node distance=1cm] (sh_bases) {$\mathcal{H}_b$};
\node[right of=l, node distance=1cm, circ, minimum size=.25cm] (sh_mult) {$\times$};
\node[right of=sh_mult, node distance=1.1cm] (shade) {\includegraphics[width=0.06\linewidth]{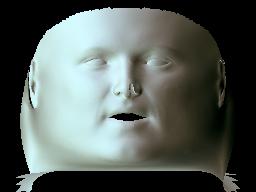}};
\node[above of=shade, node distance=0.65cm] (shade_text) {$\mathbf{C}^{uv}$};
\draw[->] (l) -- (sh_mult);
\draw[->] (sh_bases) -- (sh_mult);
\draw[->] (sh_mult) -- (shade);

\node[above of=sh_bases, node distance=1.65cm] (bfm_bases) {$\mathcal{S}_b$};
\node[right of=fs, node distance=1cm, circ, minimum size=.25cm] (bfm_mult) {$\times$};
\node[right of=bfm_mult, node distance=1.1cm] (shape) {\includegraphics[trim={100 100 100 100},clip,width=.06\linewidth]{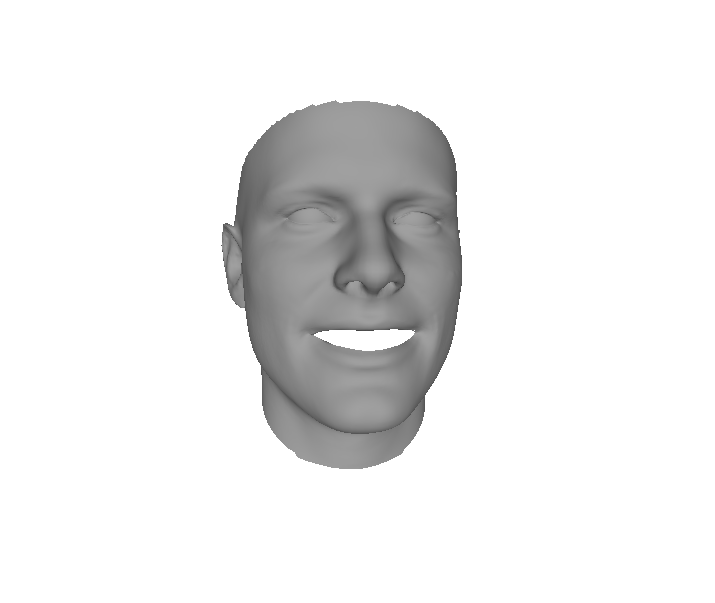}};
\node[above of=shape, node distance=0.65cm] (shape_text) {$\mathbf{S}$};
\draw[->] (fs) -- (bfm_mult);
\draw[->] (bfm_bases) -- (bfm_mult);
\draw[->] (bfm_mult) -- (shape);
\coordinate[above of=pose, node distance=0.2cm] (above_pose);
\node[right of=above_pose, node distance=2.9cm, scale=4] (3dmm_curly) {\}};

\coordinate[above right of=pose_text, node distance=0.4cm] (pose_text_above_right);
\node [block,dashed,minimum width=4.8cm,minimum height=3.6cm,fill=blue!60,right of=pose_text_above_right,node distance=0.15cm,opacity=0.2,label={[anchor=north,text opacity=0.5]north:\small{3D FACTORIZATION}}] {}; 
\coordinate[right of=shape_text, node distance=6.0cm] (separator1) {};

\node[right of=shape_text, node distance=2.2cm] (albedo_out) {\includegraphics[width=.08\linewidth]{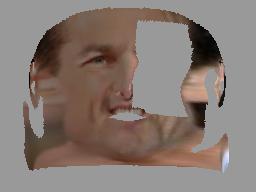}};
\node[above of=albedo_out, node distance=0.65cm] (alb_out_text) {$\mathbf{A}^{uv}_m$};
\node[right of=albedo_out, node distance=2cm,label={[anchor=north west,text=white]north west:\small{$\mathbf{M}^{uv}$}}] (mask_uv) {\includegraphics[width=.08\linewidth]{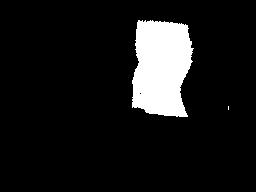}};
\node[above of=mask_uv, node distance=0.65cm] (mask_uv_text) {$\mathbf{M}^{uv}$};

\node[right of=albedo_out, node distance=1cm] (occ_alb_mult) {$\odot$};
\node[below of=occ_alb_mult, node distance=0.6cm, scale=4, rotate=90] (curly_2) {\{};
\node[below of=curly_2, node distance=1.5cm] (curly2_bottom) {};
\node[left of=curly2_bottom, node distance=.75cm, block, opacity=.5, minimum width=1cm, minimum height=4pt, inner sep=0] (alb_recon_first1) {};
\node[right of=curly2_bottom, node distance=.75cm, block, opacity=.5, minimum width=1cm, minimum height=4pt, inner sep=0] (alb_recon_first2) {};
\node[above of=alb_recon_first2, node distance=0.65cm] (flip) {\includegraphics[width=.02\linewidth]{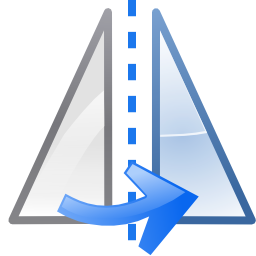}};
\node[right of=flip, node distance=0.7cm] {\scriptsize{\begin{tabular}{c} Horizontal \\ flip \end{tabular}}};
\coordinate[below of=curly_2, node distance=0.25cm] (curly2_bottom);
\draw[->] (curly2_bottom) -| (flip);
\draw[->] (flip) -- (alb_recon_first2);
\draw[->] (curly2_bottom) -| (alb_recon_first1);
\node[below of=curly2_bottom, node distance=1.7cm] (alb_recon_concat) {$\oplus$};
\draw[->] (alb_recon_first1) |- (alb_recon_concat);
\draw[->] (alb_recon_first2) |- (alb_recon_concat);

\node[below of=alb_recon_concat, node distance=.3cm, block, opacity=.5, minimum width=1.5cm, minimum height=4pt, inner sep=0] (alb_recon2) {};
\node[below of=alb_recon2, node distance=.25cm, block, opacity=.5, minimum width=1.2cm, minimum height=4pt, inner sep=0] (alb_recon3) {};
\node[below of=alb_recon3, node distance=.25cm, block, opacity=.5, minimum width=1cm, minimum height=4pt, inner sep=0] (alb_recon4) {};
\node[below of=alb_recon4, node distance=.25cm, block, opacity=.5, minimum width=.75cm, minimum height=4pt, inner sep=0] (alb_recon5) {};
\node[below of=alb_recon5, node distance=.25cm, block, opacity=.5, minimum width=.75cm, minimum height=4pt, inner sep=0] (alb_recon6) {};
\node[below of=alb_recon6, node distance=.25cm, block, opacity=.5, minimum width=1cm, minimum height=4pt, inner sep=0] (alb_recon7) {};
\node[below of=alb_recon7, node distance=.25cm, block, opacity=.5, minimum width=1.2cm, minimum height=4pt, inner sep=0] (alb_recon8) {};
\node[below of=alb_recon8, node distance=.25cm, block, opacity=.5, minimum width=1.5cm, minimum height=4pt, inner sep=0] (alb_recon9) {};
\node[below of=alb_recon9, node distance=.25cm, block, opacity=.5, minimum width=1.75cm, minimum height=4pt, inner sep=0] (alb_recon10) {};

\coordinate[right of=alb_recon5,node distance=0.5cm] (alb_recon5_right);
\node [block, dashed, minimum width=4.1cm, minimum height=4.1cm, above of=alb_recon5_right, node distance=0.6cm, fill=pink, opacity=0.3, label={[anchor=east,text opacity=0.7,label distance=0.2cm]east:\small{\begin{tabular}{c} ALBEDO \\ INPAINTER \\ $\mathcal{G}$
\end{tabular}}}] {}; 

\node[below of=alb_recon10, node distance=0.75cm,label={[anchor=north west,text=white]north west:\small{$\hat{\mathbf{A}}^{uv}$}}] (alb_comp) {\includegraphics[width=.08\linewidth]{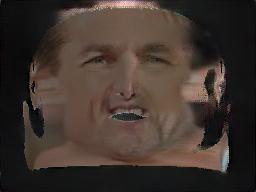}};
\node[right of=alb_comp, node distance=2.0cm,label={[anchor=north west,text=white]north west:\small{$\hat{\mathbf{A}}^{uv}_{flip}$}}] (alb_flip) {\scalebox{-1}[1]{\includegraphics[width=.08\linewidth]{figs/presentation/tom_cruise/albedo_out.jpg}}};
\draw[<->] (alb_comp) -- node [midway,below] {\scriptsize Flip} (alb_flip);
\coordinate (l_sym_point) at ($(alb_comp)!0.5!(alb_flip)$);
\node[below of=l_sym_point,node distance=0.7cm,rotate=90,scale=3] (l_sym_curly) {\{};
\node[below of=l_sym_curly,node distance=0.3cm] (l_sym_loss) {\small $\mathcal{L}_{sym}$};
\node[left of=l_sym_curly,node distance=1.5cm] (a_loss) {\small $\mathcal{L}_{A}$};

\coordinate[left of=alb_recon7,node distance=1.7cm] (shape_ren);
\coordinate[right of=above_pose, node distance=3.1cm] (3dmm_curly_point);
\coordinate[right of=3dmm_curly_point, node distance=0.3cm] (3dmm_curly_point2);
\coordinate[below of=3dmm_curly_point2, node distance=2.6cm] (rendering_inp_point);
\draw[] (3dmm_curly_point) -| (rendering_inp_point);
\draw[->] (rendering_inp_point) -- ++(-0.5,0);

\node[left of=shape_ren, node distance=1cm, trap, rotate=90, fill={rgb:red,1;green,4;blue,3}, fill opacity=0.5, text opacity=1] (rendering) {\textbf{Renderer}};
\node[left of=rendering, node distance=1.5cm, label={[anchor=north west,text=white]north west:\small{$\hat{\mathbf{I}}$}}] (final_output) {\includegraphics[width=.08\linewidth]{figs/presentation/tom_cruise/completed.png}};
\node[below of=final_output, node distance=0.75cm] (alb_out_text) {\footnotesize Completed};
\node[below of=final_output, node distance=1.6cm, label={[anchor=north west,text=white]north west:\small{$\mathbf{I}_{gt}$}}] (ground_truth) {\includegraphics[width=.08\linewidth]{figs/presentation/tom_cruise/original.jpg}};
\node[below of=ground_truth, node distance=0.75cm] (alb_out_text) {\footnotesize Groundtruth};

\node (output_center) at ($(final_output)!0.5!(ground_truth)$) {};
\node[left of=output_center, node distance=1cm, disc_layer, minimum width=1.5cm] (img_disc1) {};
\node[left of=img_disc1, node distance=.25cm, disc_layer, minimum width=1.2cm] (img_disc2) {};
\node[left of=img_disc2, node distance=.25cm, disc_layer, minimum width=1cm] (img_disc3) {};
\node[left of=img_disc3, node distance=.25cm, disc_layer, minimum width=.75cm] (img_disc4) {};
\node[left of=img_disc4, node distance=0.6cm] (real_fake2) {\small $\mathcal{L}_{GAN}$};
\node[left of=final_output, node distance=2.2cm] (img_disc_text) {\footnotesize{\begin{tabular}{c}
     PyramidGAN
\end{tabular}}};

\node[below of=img_disc1, node distance=1.1cm] (i_loss) {\small $\mathcal{L}_{I}$};

\coordinate[below=1cm of shape_ren.west, node distance=1cm] (shape_bottom);
\coordinate[below of=rendering_inp_point, node distance=0.8cm] (alb_comp_left);
\draw[] (alb_comp) -| (alb_comp_left);
\draw[->] (alb_comp_left) -- ++(-0.5cm,0);
\draw[->] (rendering) -- (final_output);

\coordinate[above=0.25cm of final_output] (final_output_top);
\draw[dashed] (final_output) -- (final_output_top);
\draw[dashed,->] (final_output_top) -| node [pos=0.25,above] {\scriptsize Iter 2} (enc1);
\end{tikzpicture}

\caption{\textbf{Architecture}}
\label{fig:architecture}
\end{subfigure}
\begin{subfigure}{0.21\textwidth}
    \centering
    \begin{subfigure}{\linewidth}
        \includegraphics[width=\columnwidth]{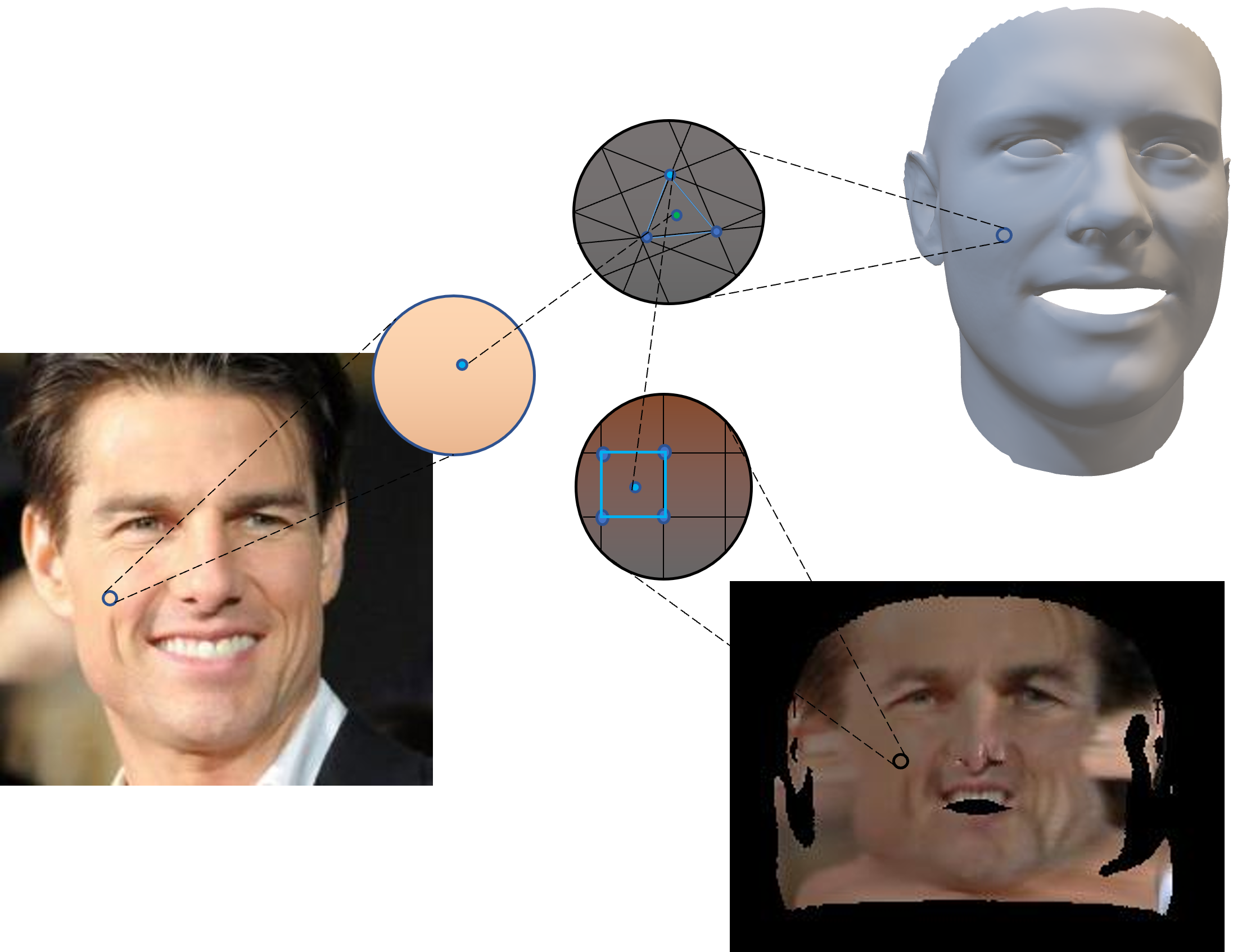}
        \caption{\textbf{UV Sampling}}
        \label{fig:uv_sampling}
    \end{subfigure}
    \vspace{5mm}
    \begin{subfigure}{\linewidth}
        \includegraphics[width=\columnwidth]{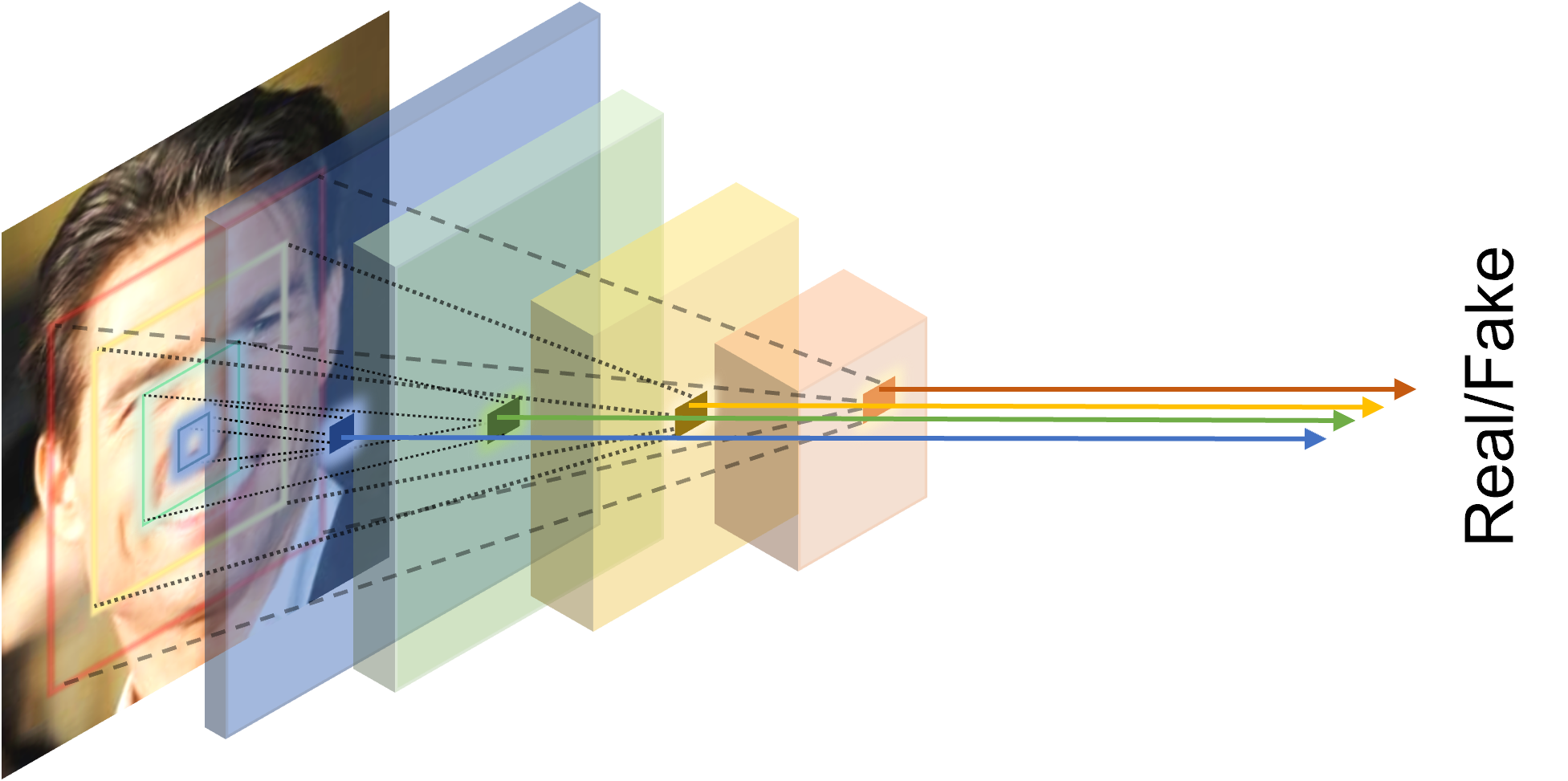}
        \caption{\textbf{PyramidGAN}}
        \label{fig:pyramidgan}
    \end{subfigure}
\end{subfigure}

\caption{\label{fig:arch_overview} \textbf{(a) Architecture:} Given a masked face $\mathbf{I}_m$, the 3DMM encoder extracts its shape $\mathbf{f_S}$, pose $\mathbf{p}$ and illumination $\mathbf{l}$ parameters, from which we obtain the full shape $\mathbf{S}$ and shade $\mathbf{C}^{uv}$ by linear combination of the corresponding bases. Then a partial albedo $\mathbf{A}^{uv}_{m}$ is obtained by first re-projecting the 3D mesh onto the masked image to obtain the UV-texture, as shown in \textbf{(b)} and then removing the shade from it $\mathbf{A}^{uv}_{m} = \mathbf{T}^{uv}_{m} \oslash \mathbf{C}^{uv}$. Finally, the albedo inpainter $\mathcal{G}$ completes the partial albedo as $\hat{\mathbf{A}}^{uv}$, conditioned on the UV-mask $\mathbf{M}^{uv}$, which is rendered along with the estimated shape, pose and shade to obtain the completed image $\hat{\mathbf{I}}$. To generate photorealistic completion, the completed and groundtruth images are evaluated by the proposed \textbf{(c) PyramidGAN} discriminator. \textbf{(b) UV Sampling:} 3D mesh is projected onto the face image to obtain per vertex RGB values $\mathbf{T}_{\mathbf{v}}(x,y,z)$. Each mesh face triangle $\mathbf{t=(v_1,v_2,v_3)}$ is mapped to a particular pixel in the UV space $\mathbf{T}^v_{m}(t) \rightarrow \mathbf{T}^{uv}_{m}(u,v)$ which allows us to sample the UV texture using barycentric interpolation.\vspace{-3mm}}
\end{figure*}

In this section, we first present an overview of our proposed 3D face completion approach (dubbed \ourmethod{}) followed by the details of each component. As shown in Fig.~\ref{fig:overview}, \ourmethod{} has three components: a 3DMM encoder, an albedo completion module and a renderer. Given a masked face, \ourmethod{} first resolves it into its constituent 3D shape, pose and illumination using the 3DMM encoder (Fig.~\ref{fig:arch_overview}). Then, we obtain the partial facial texture in the UV-domain by re-projecting the mesh onto the input image (Fig.~\ref{fig:uv_sampling}). We further remove the shading component to obtain an illumination-invariant partial albedo. The inpainter completes the partial albedo using symmetric and learned priors. Finally, the renderer combines the inpainted albedo with the estimated 3D factors to obtain the completed face. As a natural extension of the proposed approach, we use 3D factorization and completion in a complimentary way to further improve completion iteratively.

\subsection{3D Factorization}\label{sec:3dmm}
Existing face image completion approaches directly operate on 2D, which makes it non-trivial to enforce strong 3D geometric and photometric priors. This leads to poor face completion in challenging conditions of poses, geometry, lighting, \etc. This motivates us to adopt explicit 3D factorization of face images to disentangle the appearance and geometric components, to enable robust completion.


Essentially, the 3D factorization module is an inverse renderer $\Phi:\mathbf{I}\rightarrow (\mathbf{S}, \mathbf{p}, \mathbf{l}, \mathbf{A})$ that resolves a 2D face $\mathbf{I}$ into its constituent shape $\mathbf{S} \in \mathbb{R}^3$, pose $\mathbf{p}$, illumination $\mathbf{l}$ and albedo $\mathbf{A}$. Various 3DMM approaches like \cite{blanz1999morphable, egger2018occlusion, ganfit} can be a natural fit for this. However, they are not real time, leaving learning based 3D reconstruction approaches \cite{mofa, sfsnet2018sengupta, neuralfaceediting, tran2018extreme, tran2019learning, tran2019towards, unsup3dcvpr2020} as the obvious choices. While any of these approaches can potentially be used in our approach, for the purpose of this work, we adopt a simplified version of the nonlinear 3DMM presented by Tran \etal \cite{tran2019learning}. 

The 3D factorizaiton module consists of a 3DMM encoder and an albedo decoder (used only during training). The encoder $\mathcal{E}$ first resolves the image $\mathbf{I}$ in to its shape, albedo and illumination coefficients $(\mathbf{f_S}, \mathbf{f_A}, \mathbf{l})$ and pose $\mathbf{p}=(s,\mathbf{R},\mathbf{t})$. Using the shape coefficients, we obtain the full shape $\mathbf{S}$ by linear combination with the Basel Face Model's (BFM) bases \cite{bfm}. Similarly, we combine the illumination coefficients linearly with the spherical harmonics (SH) bases $\mathbf{H}_b$ \cite{sphericalharmonics} to obtain the surface shading $\mathbf{C}^{uv}$ (we assume \textit{Lambertian} surface reflectance). The decoder $\mathcal{D}_{\mathbf{A}}$ maps the albedo coefficients into the full UV-albedo $\mathcal{D}_{\mathbf{A}}:\mathbf{f_A} \rightarrow \mathbf{A}^{uv}$, which is then multiplied with the shade to obtain the texture $\mathbf{T}^{uv}=\mathbf{A}^{uv}\odot\mathbf{C}^{uv}$. A differentiable renderer $\mathcal{R}$ \cite{tran2019learning} then reprojects the estimated 3D factors into image $\mathbf{I}_{ren}$ using the Z-buffer technique:
\begin{equation}
    \label{eqn:image_form}
    \mathbf{I}_{ren} = \mathcal{R}\left( \mathbf{S}, \mathbf{A}, \mathbf{p}, \mathbf{l} \right)
\end{equation}

We train the module using masked images for robustness to partial inputs. For further details, refer the supplement.

\subsection{Albedo Completion Module} \label{sec:albedo_comp}
Architecturally, our albedo completion module is similar to other adversarially trained image-completion autoencoders \cite{contextencoderpathak, genfacecompletion, gencontextualattn}. However, ours has the unique advantage of being solely focused on recovering the missing albedo, which has been disentangled from other variations in shape, pose and illumination through 3D factorization and is largely symmetric in its UV-representation. UVGAN \cite{uvgan} performs a similar completion of self-occluded UV-texture extracted from fully-visible face images. However, because of the entangled illumination, they don't use symmetry and need a synthetically completed texture map for supervision, whereas we use symmetry as self-supervision. 

To this end, we discard the soft albedo obtained from the 3DMM albedo decoder and instead obtain the more realistic partial albedo from the input image in the UV space. This is done in two steps: first, we reproject the obtained 3D mesh onto the face image and use bilinear interpolation to sample the per-vertex texture (see Fig.~\ref{fig:uv_sampling}):
\begin{align*}
    \mathbf{T}^{\mathbf{v}}_m(x,y,z) = \sum_{\substack{p\in\{\floor{x},\ceil{x}\} \\ q\in\{\floor{y},\ceil{y}\}}} \mathbf{I}_m^{p,q}(1-|x-p|)(1-|y-q|)
\end{align*}

Then, we map the sampled partial texture $\mathbf{T}^{\mathbf{v}}_{m}$ onto the UV space using barycentric interpolation on the predefined mesh-to-uv mappings $\mathbf{T}^{\mathbf{v}}_{m}(\mathbf{v}_1,\mathbf{v}_2,\mathbf{v}_3) \rightarrow \mathbf{T}^{uv}_{m}(u,v)$. From the texture, we obtain the partial albedo by simply removing the estimated shade: $\mathbf{A}^{uv}_{m} = \mathbf{T}^{uv}_{m} \oslash \mathbf{C}^{uv}$, where $\oslash$ is the element-wise division operation. We perform similar operations to unwarp the mask $\mathbf{M}$ on-to the UV-space as $\mathbf{M}^{uv}$.

We use a U-Net \cite{unet} based autoencoder $\mathcal{G}$ to complete the partial albedo conditioned on the input mask, $\mathcal{G}: (\mathbf{A}^{uv}_{m}, \mathbf{M}^{uv}) \rightarrow (\hat{\mathbf{A}}^{uv}, \sigma^{uv}$), where $\hat{\mathbf{A}}^{uv}$ is the completed albedo and $\sigma^{uv}$ is the uncertainty of completion. In order to leverage the bilateral symmetry of the UV facial albedo as an attention map, we modify the U-Net architecture (henceforth referred to as Sym-UNet). This is specially helpful since we do not have access to the full groundtruth albedo maps for training. To do so, we split the first convolution layer $f_{1:2c}$ into two parts: $f_{1,1:c}$ and $f_{2,c+1:2c}$ with equal number of output channels $c$ (see Fig.~\ref{fig:overview}). The first filter operates on the input albedo as such $\mathbf{h}_1=f_1(\mathbf{A}^{uv}_{m})$. The second, instead, operates on the horizontally flipped albedo $\mathbf{h}_2=f_2(hflip(\mathbf{A}^{uv}_{m}))$. We then concatenate the activations $\mathbf{h}_1$ and $\mathbf{h}_2$ from these two filters and pass it through the rest of the network. During training, the first filter learns to extract features from the visible parts of the albedo while the second filter learns to extract features corresponding to the symmetrically opposite visible parts to apply on the occluded regions (see Sec. \textcolor{red}{3.2} in the supplementary). 

A naive approach of doing so, however, results in artifacts from the symmetrical counterparts to appear on the visible regions, making the network convergence difficult. Instead, we use gated convolutions \cite{freeforminpainting} (in all but the final layer), to ensure that such symmetric features are only transferred to the masked regions and do not create artifacts on the visible regions. We use group normalization\cite{groupnorm} and ELU activation\cite{elu} for all the feature layers and the final output is simply clipped between -1 and 1. We then render the completed albedo $\hat{\mathbf{A}}^{uv}$, along with the estimated shape, pose and illumination to obtain a completed image $\hat{\mathbf{I}}$ using eqn.~\ref{eqn:image_form}. Finally, we simply blend the input and completed images to obtain the output image: $\mathbf{I}_{out} = \mathbf{I} \odot (1 - \mathbf{M}) + \hat{\mathbf{I}} \odot \mathbf{M}$.

\vspace{5pt}
\noindent \textbf{PyramidGAN Discriminator:} To generate sharp and semantically realistic completions, we use a multi-scale PatchGAN discriminator \cite{wang2018high, shocher2019ingan}, which we refer to as the \textit{PyramidGAN}. The PyramidGAN evaluates the final output $\mathbf{I}_{out}$ at multiple locations and scales ranging from coarse and global to fine and local (refer to Fig.~\ref{fig:pyramidgan}). Features from each $l$-th downsampling layer of the PyramidGAN $\mathcal{D}_l$ are used to evaluate an average hinge loss \cite{freeforminpainting, juefei2018rankgan} for that layer. We then compute the average loss across all the layers as the total loss, thus giving equal weightage to each scale:
\begin{align}
    \label{eqn:gan_loss}
    \mathcal{L}_{\mathcal{G}} = &-\mathbb{E}_{p(z)}\left[\mathbb{E}_{l\in L} \left[ \mathcal{D}_l(\mathcal{G}(z) \right] \right] \numberthis\\
    \mathcal{L}_{\mathcal{D}} = &\mathbb{E}_{x}\left[\mathbb{E}_{l\in L} [\mathbf{1} - \mathcal{D}_l(x)]_+ \right] + \mathbb{E}_{p(z)}\left[\mathbb{E}_{l\in L} [\mathbf{1} + \mathcal{D}_l(\mathcal{G}(z)]_+ \right] \nonumber,
\end{align}

\vspace{5pt}
\noindent \textbf{Training Losses:} We train the albedo completion module with the following total loss:
\begin{align}
    \mathcal{L} &= \lambda_{1}\mathcal{L}_{A} + \lambda_{2}\mathcal{L}_{I} + \lambda_{3}\mathcal{L}_{sym} + \lambda_{4}\mathcal{L}_{GAN} + \lambda_{5}\mathcal{L}_{gp},
\end{align}
where $\mathcal{L}_{A} = \mathcal{L}_{\sigma}(||\hat{\mathbf{A}}^{uv} - \hat{\mathbf{A}}^{uv}_{gt}||_1, \sigma^{uv})$ and $\mathcal{L}_{I} = \mathcal{L}_{\sigma}(||\hat{\mathbf{I}} - \mathbf{I}_{gt}||_1, \sigma)$ are the pixel losses for the albedo and the image, respectively, $\mathcal{L}_{sym}$ is the symmetry loss, $\mathcal{L}_{GAN}$ is the GAN loss given in eqn.~\ref{eqn:gan_loss} and $\mathcal{L}_{gp}$ is the WGAN-GP loss as described in \cite{wgangp}. The albedo symmetry loss is carefully applied on the masked regions whose symmetric counterparts are visible, to supplement as supervised attention:
\begin{align}
    \mathcal{L}_{sym} = \mathcal{L}_{\sigma}\left((\mathbf{1}-\mathbf{M}^{uv})\mathbf{M}^{uv}_{flip} \odot ||\hat{\mathbf{A}}^{uv} - \hat{\mathbf{A}}^{uv}_{flip}||_1,\nonumber \sigma^{uv}\right)
\end{align}
Here, $\mathcal{L}_{\sigma} (\mathbf{x}, \sigma) = \frac{1}{D} \sum_{i}\frac{1}{2} x_i exp(-\sigma_i) + \frac{\sigma_i}{2}$ is the aleatoric uncertainty loss\cite{kendall2017uncertainties}. The loss coefficients are set to have similar magnitude for all the loss components. In this paper, the goal is to show the efficacy of explicit 3D consideration on the geometric and photometric accuracy of face completion. So, \emph{we withhold from using attention or face specific losses \cite{genfacecompletion, gencontextualattn, freeforminpainting, zheng2019pluralistic, oracleattention}} and leave them as future add-ons.

\vspace{5pt}
\noindent \textbf{Iterative Refinement:} 3D factorization is an important first step of our proposed approach, which itself leads to robust face completion in cases where 2D based methods fail. To make the 3D factorization itself robust to partial images, we train the 3DMM encoder on face images with randomly sized and randomly located masks. However, there is scope to further improve upon this and leverage the full power of our proposed two-step approach. To do this, we adopt a simple iterative refinement technique where face completion leads to improved 3D factorization and vice versa, as shown in Fig.~\ref{fig:overview}. During inference, the masked face is used to distill the 3D factors in the first iteration; while in the next iteration, the completed face itself forms the input for 3D analysis. This leads to iteratively refined 3D analysis (\emph{specially the 3D pose}) as well as face completion. Though one can repeat the iterative step many times, we experimentally found that two such iterations are usually sufficient.

\section{Experimental Evaluation}

\begin{figure*}
    \centering
    \begin{subfigure}{.49\textwidth}
        \centering
        \includegraphics[width=.49\linewidth,trim={0 10 60 70},clip]{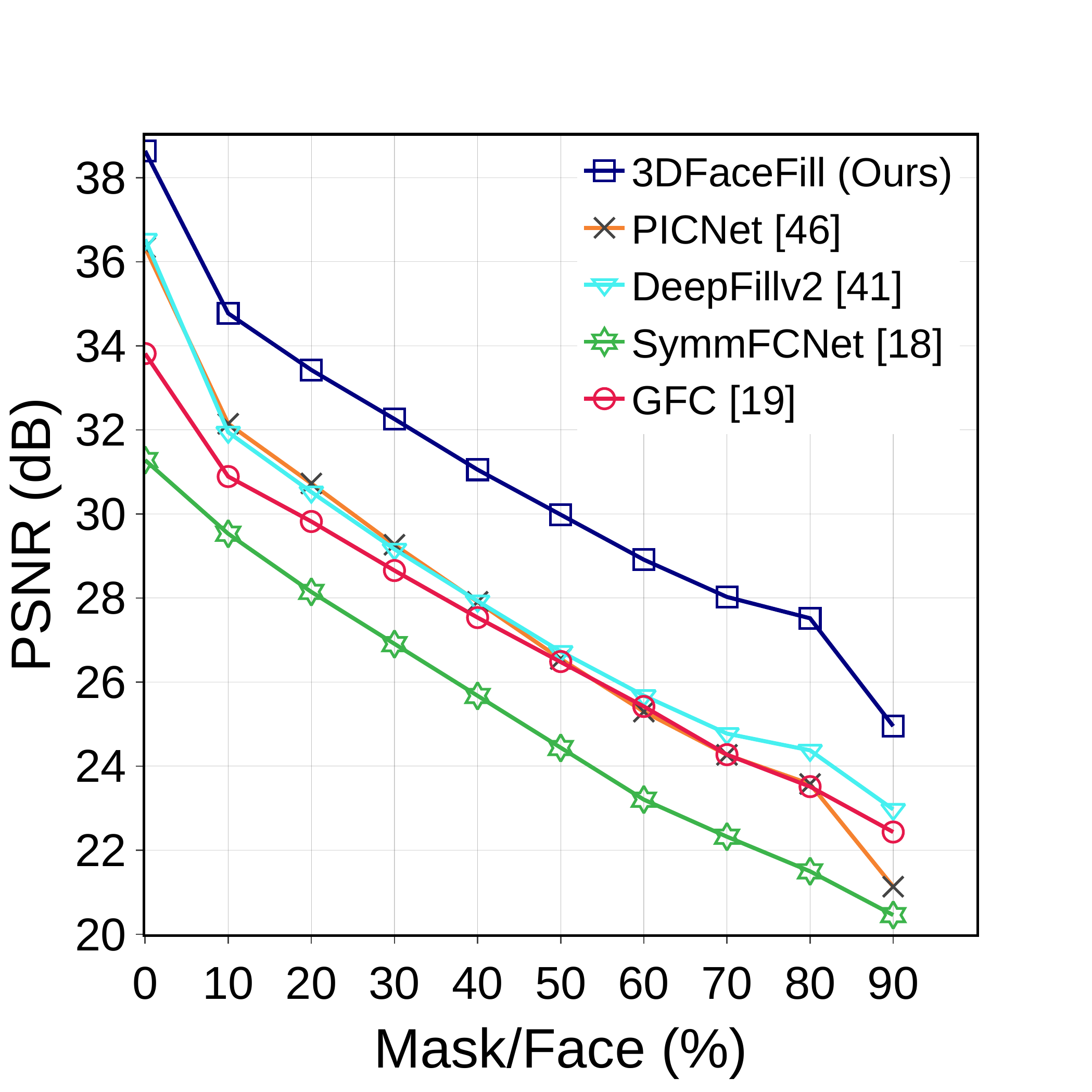}
        \includegraphics[width=.49\linewidth,trim={0 10 60 70},clip]{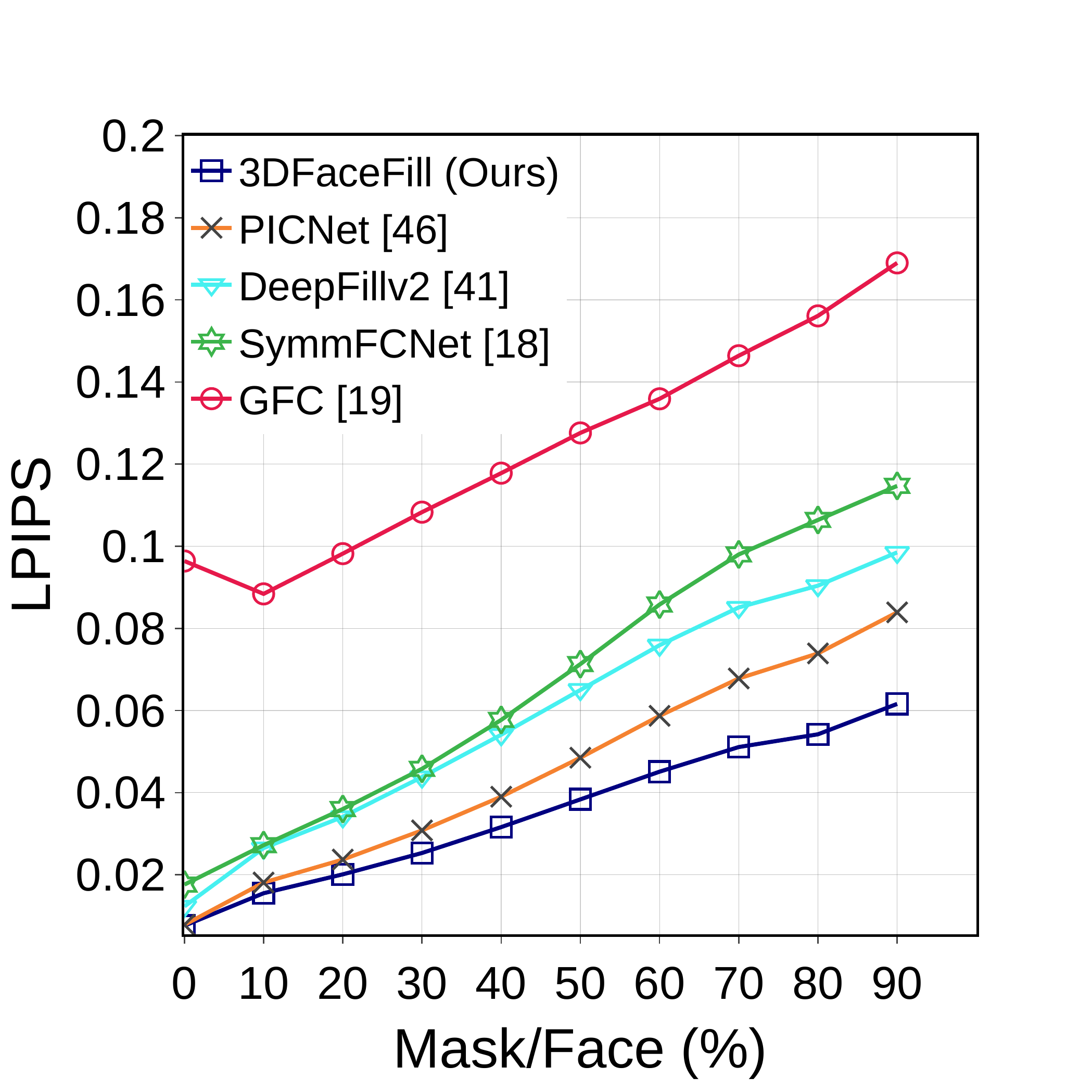}
        \caption{CelebA dataset \cite{celeba}}
        \label{fig:celeba_quant}
    \end{subfigure}
    \begin{subfigure}{.49\textwidth}
        \centering
        \includegraphics[width=.49\linewidth,trim={0 10 60 70},clip]{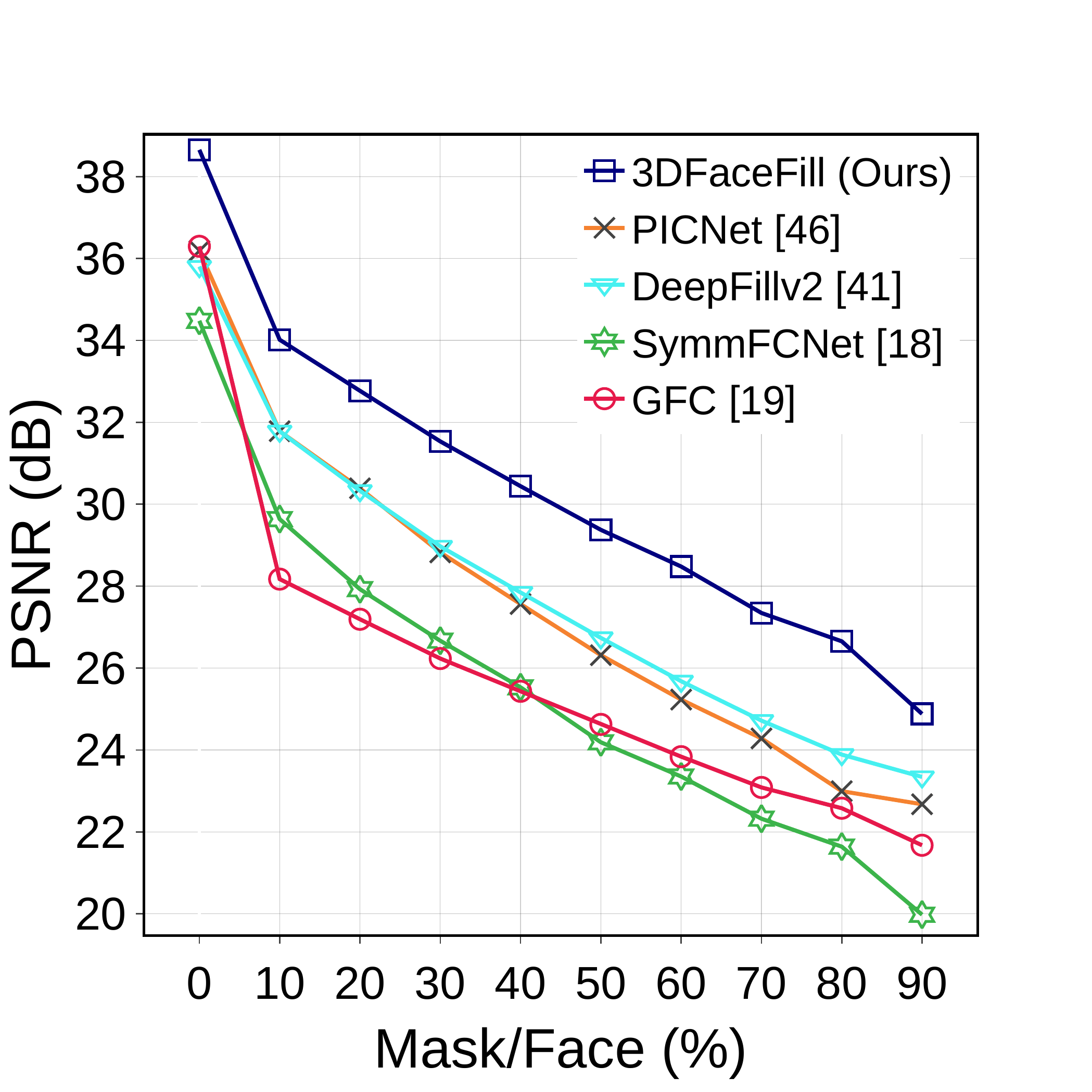}
        \includegraphics[width=.49\linewidth,trim={0 10 60 70},clip]{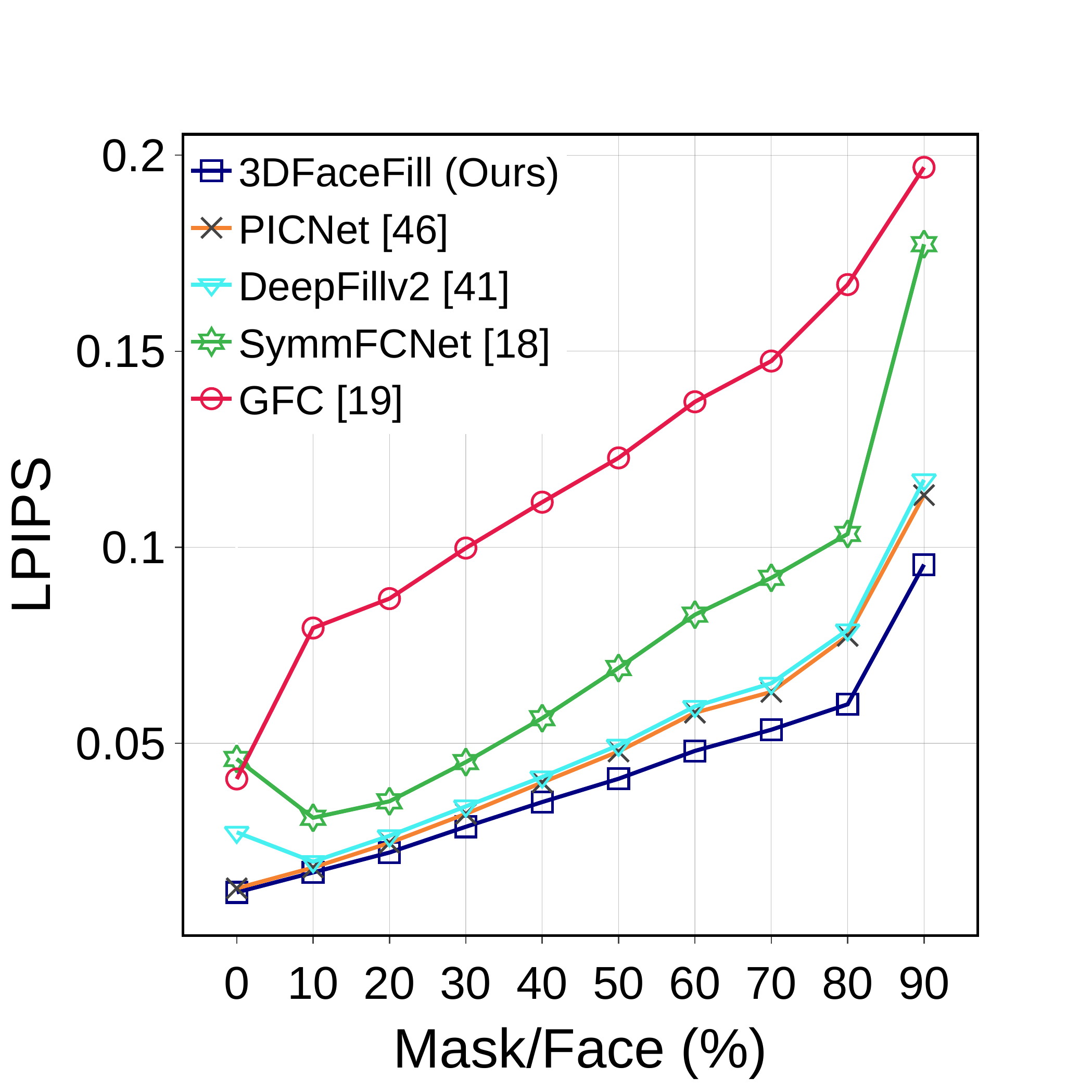}
        \caption{CelebA-HQ dataset \cite{CelebAMask-HQ}}
        \label{fig:celebahq_quant}
    \end{subfigure}
    \begin{subfigure}{.49\textwidth}
        \centering
        \includegraphics[width=.49\linewidth,trim={0 10 60 70},clip]{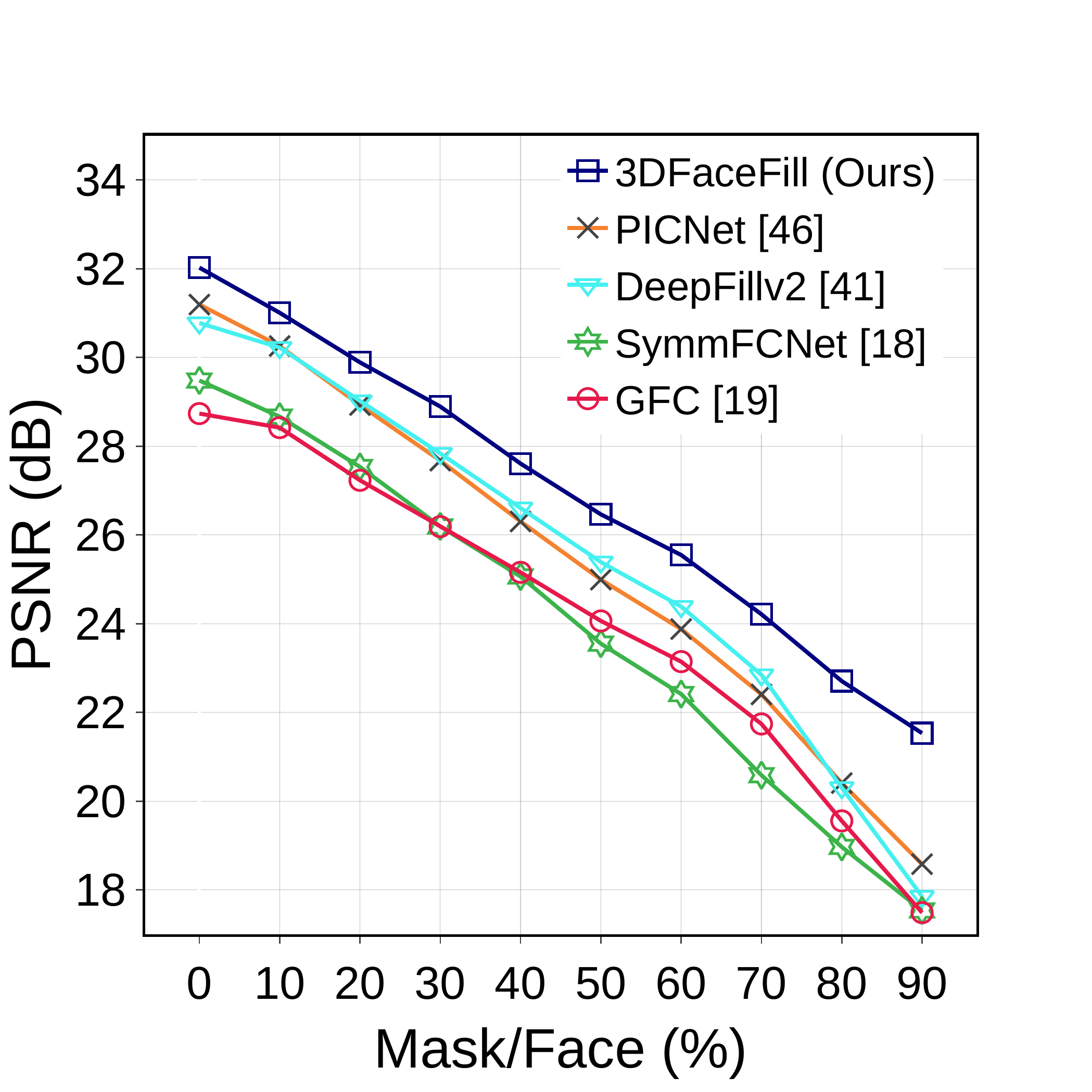}
        \includegraphics[width=.49\linewidth,trim={0 10 60 70},clip]{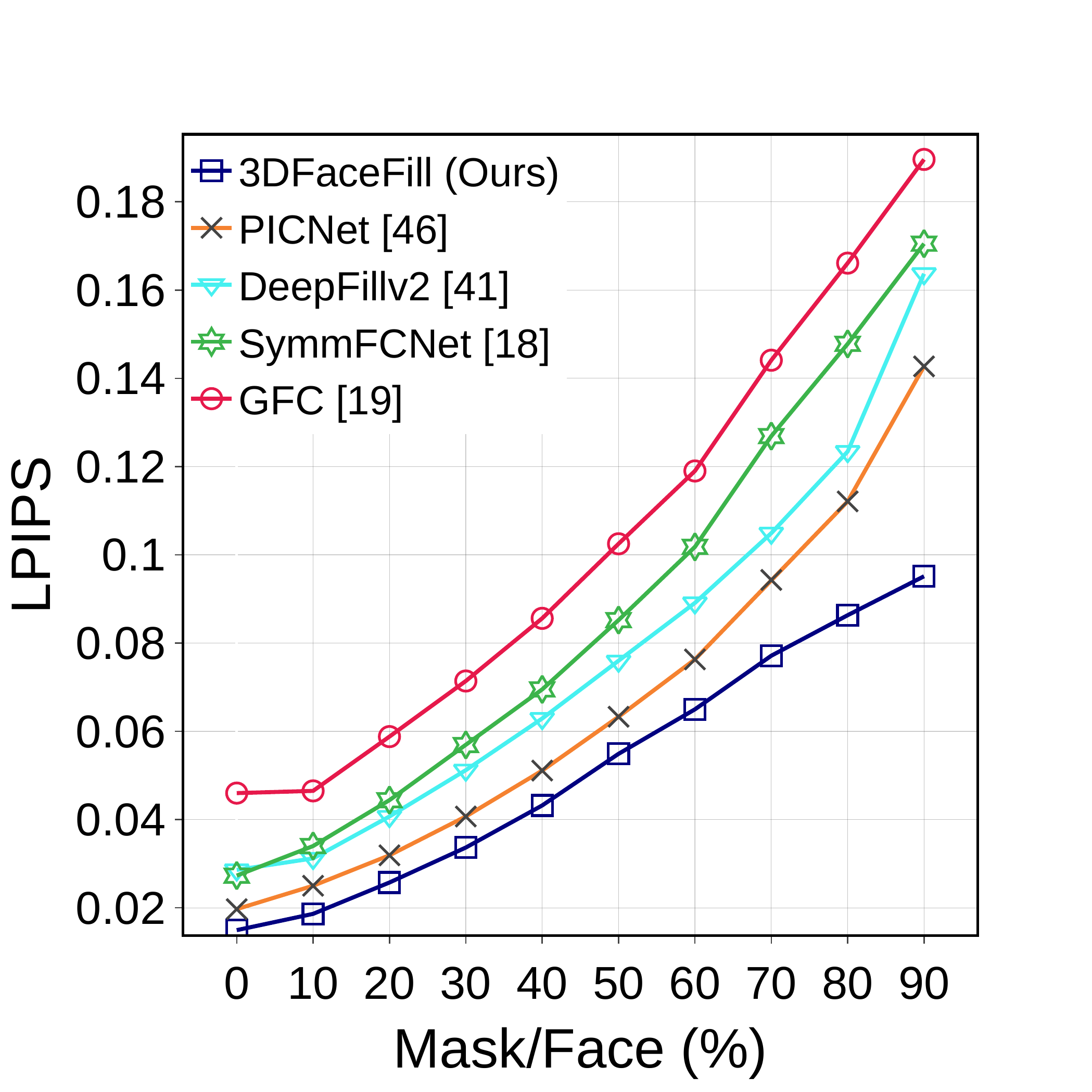}
        \caption{MultiPIE dataset \cite{multipie}}
        \label{fig:multipie_quant}
    \end{subfigure}
    \begin{subfigure}{.49\textwidth}
        \centering
        \newcolumntype{M}[1]{>{\raggedright\arraybackslash}m{#1}}
        \resizebox{.99\linewidth}{!}{\begin{tabular}{M{17mm}l@{\hspace{1\tabcolsep}}c@{\hspace{1\tabcolsep}}c@{\hspace{1\tabcolsep}}c@{\hspace{1\tabcolsep}}c@{\hspace{1\tabcolsep}}c}
            \hline
            \textbf{Dataset} & \textbf{Metric} & GFC \cite{genfacecompletion} & SymmFC \cite{li2020symmetry} & DeepFill \cite{freeforminpainting} & PIC \cite{zheng2019pluralistic} & \ourmethod{}\\
            \hline
            \multirow{3}{*}{\textbf{CelebA}} & \textbf{PSNR} $(\uparrow)$ & 27.0298 & 25.8817 & 28.2097 & 28.1262 & \textbf{30.4917}\\
            & \textbf{SSIM} $(\uparrow)$ & 0.9257 & 0.9273 & 0.9356 & 0.9424 & \textbf{0.9521}\\
            & \textbf{LPIPS} $(\downarrow)$ & 0.1134 & 0.0537 & 0.0499 & 0.0362 & \textbf{0.0326}\\
            \hline
            \multirow{3}{*}{\textbf{CelebAHQ}} & \textbf{PSNR} $(\uparrow)$ & 25.5836 & 25.6203 & 27.9885 & 27.7020 & \textbf{29.9398}\\
            & \textbf{SSIM} $(\uparrow)$ & 0.8895 & 0.9232 & 0.9311 & 0.9380 & \textbf{0.9492}\\
            & \textbf{LPIPS} $(\downarrow)$ & 0.1076 & 0.0535 & 0.0394 & 0.0376 & \textbf{0.0365}\\
            \hline
            \multirow{3}{*}{\textbf{MultiPIE}} & \textbf{PSNR} $(\uparrow)$ & 25.3805 & 25.1280 & 26.8225 & 26.5574 & \textbf{28.7515}\\
            & \textbf{SSIM} $(\uparrow)$ & 0.9127 & 0.9266 & 0.9391 & 0.9397 & \textbf{0.9553}\\
            & \textbf{LPIPS} $(\downarrow)$ & 0.0798 & 0.0645 & 0.0577 & 0.0472 & \textbf{0.0436}\\
            \hline
        \end{tabular}}
        \caption{Average metrics across all masks.}
        \label{tab:quantitative}
    \end{subfigure}
    \vspace{-2mm}
    \caption{\textbf{Quantitative Evaluation:} We perform face completion over (a) CelebA \cite{celeba}, (b) CelebA-HQ \cite{CelebAMask-HQ} and (c) MultiPIE \cite{multipie} datasets across a range (0-90\%) of mask to face area ratios and evaluate the PSNR and LPIPS \cite{lpips} metrics. In addition, we report the overall metrics across all mask-to-face are ratios in Table (d). \ourmethod{} consistently outperforms the baselines across all the datasets and mask ratios. \vspace{-3mm}}
    \label{fig:quantitative}
\end{figure*}

\begin{figure*}
    \footnotesize
    \centering
    
    \tikzstyle{block} = [rectangle, draw, fill=blue!20, text centered]
    \begin{tikzpicture}
    \node[rotate=90, text centered] (a0) {Darker Skin};
    \node[right of=a0, node distance=1.25cm] (a1) {\includegraphics[width=0.24\columnwidth,trim={20 20 20 20},clip]{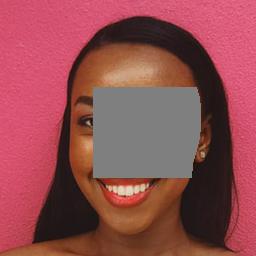}};
    \node[right of=a1, node distance=2.1cm] (a2) {\includegraphics[width=0.24\columnwidth,trim={20 20 20 20},clip]{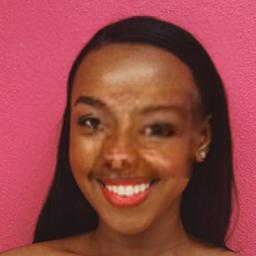}};
    \node[right of=a2, node distance=2.1cm] (a3) {\includegraphics[width=0.24\columnwidth,trim={20 20 20 20},clip]{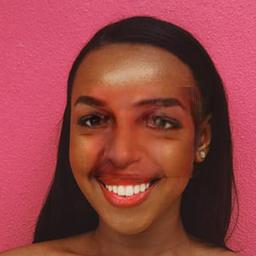}};
    \node[right of=a3, node distance=2.1cm] (a5) {\includegraphics[width=0.24\columnwidth,trim={20 20 20 20},clip]{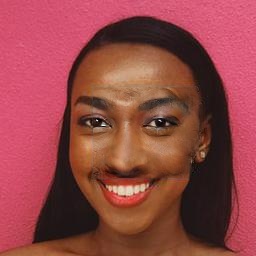}};
    \node[right of=a5, node distance=2.1cm] (a6) {\includegraphics[width=0.24\columnwidth,trim={20 20 20 20},clip]{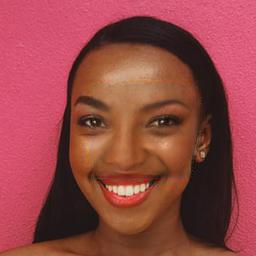}};
    \node[right of=a6, node distance=2.1cm] (a8) {\includegraphics[width=0.24\columnwidth,trim={20 20 20 20},clip]{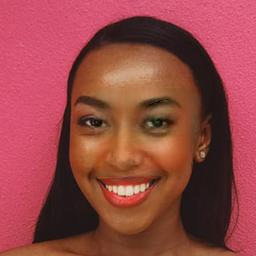}};
    \node[right of=a8, node distance=2.1cm] (a9) {\includegraphics[width=0.24\columnwidth,trim={20 20 20 20},clip]{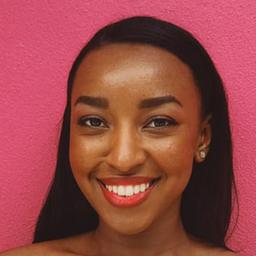}};

    \node[below of=a1, node distance=2.1cm] (g1) {\includegraphics[width=0.24\columnwidth,trim={20 20 20 20},clip]{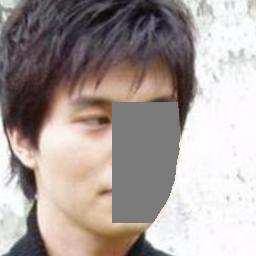}};
    \node[right of=g1, node distance=2.1cm] (g2) {\includegraphics[width=0.24\columnwidth,trim={20 20 20 20},clip]{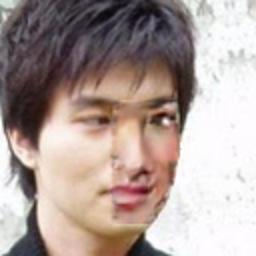}};
    \node[right of=g2, node distance=2.1cm] (g3) {\includegraphics[width=0.24\columnwidth,trim={20 20 20 20},clip]{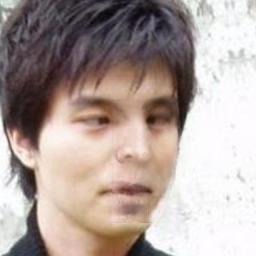}};
    \node[right of=g3, node distance=2.1cm] (g5) {\includegraphics[width=0.24\columnwidth,trim={20 20 20 20},clip]{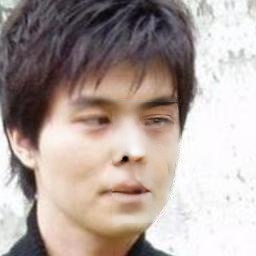}};
    \node[right of=g5, node distance=2.1cm] (g6) {\includegraphics[width=0.24\columnwidth,trim={20 20 20 20},clip]{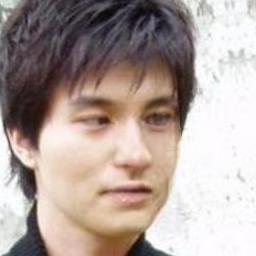}};
    \node[right of=g6, node distance=2.1cm] (g8) {\includegraphics[width=0.24\columnwidth,trim={20 20 20 20},clip]{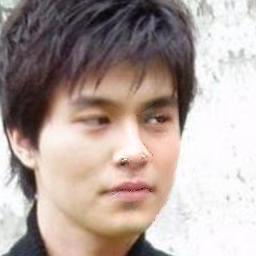}};
    \node[right of=g8, node distance=2.1cm] (g9) {\includegraphics[width=0.24\columnwidth,trim={20 20 20 20},clip]{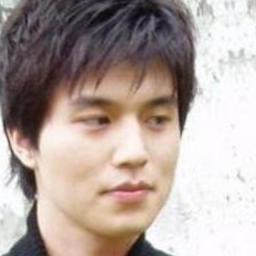}};
    \node[left of=g1, node distance=1.25cm, rotate=90, text centered] (g0) {Eye Gaze};
    
    
    \node[below of=g1, node distance=2.1cm] (i1) {\includegraphics[width=0.24\columnwidth,trim={20 20 20 20},clip]{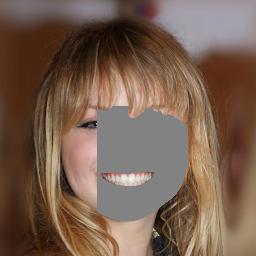}};
    \node[right of=i1, node distance=2.1cm] (i2) {\includegraphics[width=0.24\columnwidth,trim={20 20 20 20},clip]{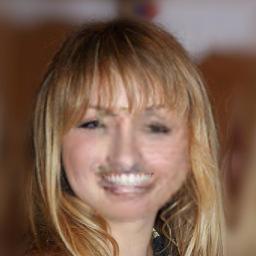}};
    \node[right of=i2, node distance=2.1cm] (i3) {\includegraphics[width=0.24\columnwidth,trim={20 20 20 20},clip]{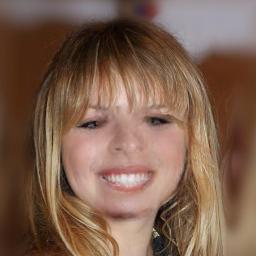}};
    \node[right of=i3, node distance=2.1cm] (i5) {\includegraphics[width=0.24\columnwidth,trim={20 20 20 20},clip]{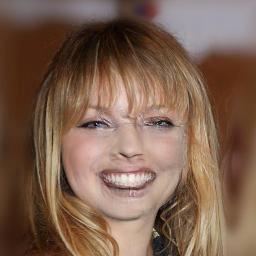}};
    \node[right of=i5, node distance=2.1cm] (i6) {\includegraphics[width=0.24\columnwidth,trim={20 20 20 20},clip]{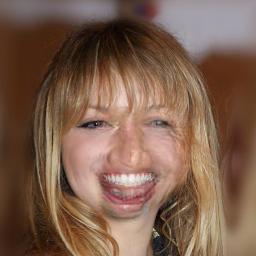}};
    \node[right of=i6, node distance=2.1cm] (i8) {\includegraphics[width=0.24\columnwidth,trim={20 20 20 20},clip]{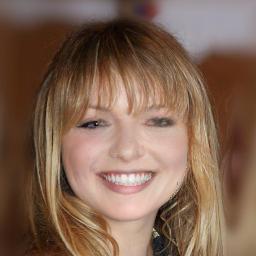}};
    \node[right of=i8, node distance=2.1cm] (i9) {\includegraphics[width=0.24\columnwidth,trim={20 20 20 20},clip]{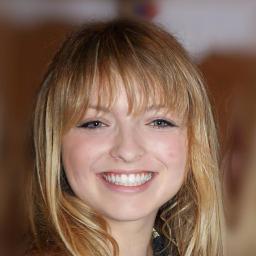}};
    \node[left of=i1, node distance=1.25cm, rotate=90, text centered] (i0) {Deformations};
    
    
    \node[below of=i1, node distance=2.1cm] (k1) {\includegraphics[width=0.24\columnwidth,trim={20 20 20 20},clip]{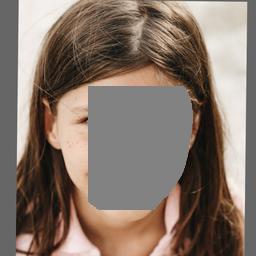}};
    \node[right of=k1, node distance=2.1cm] (k2) {\includegraphics[width=0.24\columnwidth,trim={20 20 20 20},clip]{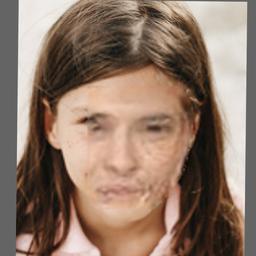}};
    \node[right of=k2, node distance=2.1cm] (k3) {\includegraphics[width=0.24\columnwidth,trim={20 20 20 20},clip]{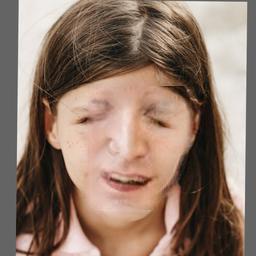}};
    \node[right of=k3, node distance=2.1cm] (k5) {\includegraphics[width=0.24\columnwidth,trim={20 20 20 20},clip]{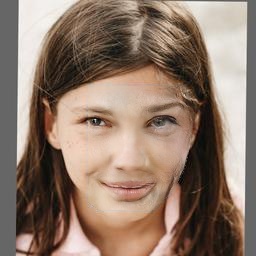}};
    \node[right of=k5, node distance=2.1cm] (k6) {\includegraphics[width=0.24\columnwidth,trim={20 20 20 20},clip]{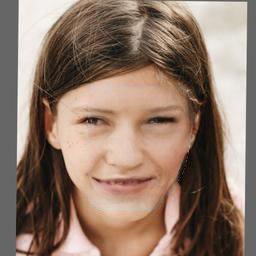}};
    \node[right of=k6, node distance=2.1cm] (k8) {\includegraphics[width=0.24\columnwidth,trim={20 20 20 20},clip]{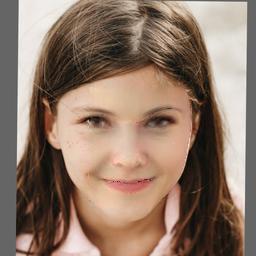}};
    \node[right of=k8, node distance=2.1cm] (k9) {\includegraphics[width=0.24\columnwidth,trim={20 20 20 20},clip]{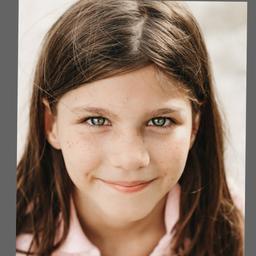}};
    \node[left of=k1, node distance=1.25cm, rotate=90, text centered] (k0) {Kids};

    \node[below of=k1, node distance=1.25cm] (o1) {\footnotesize Input};
    \node[below of=k2, node distance=1.25cm] (o2) {\footnotesize GFC \cite{genfacecompletion}};
    \node[below of=k3, node distance=1.25cm] (o3) {\footnotesize SymmFCNet \cite{li2020symmetry}};
    \node[below of=k5, node distance=1.25cm] (o5) {\footnotesize DeepFill \cite{freeforminpainting}};
    \node[below of=k6, node distance=1.25cm] (o6) {\footnotesize PIC \cite{zheng2019pluralistic}};
    \node[below of=k8, node distance=1.25cm] (o8) {\footnotesize \ourmethod{} (Ours)};
    \node[below of=k9, node distance=1.25cm] (o9) {\footnotesize Ground Truth};
    \end{tikzpicture}
    \vspace{-1mm}
    \caption{\textbf{Qualitative Evaluation:} Inpainting on faces from the CelebA \cite{celeba} and CelebA-HQ \cite{CelebAMask-HQ} test sets (except last row downloaded from the internet). Across a variety of scenarios, the completions from baselines often have artifacts while those from \ourmethod{} are significantly more photorealistic due to explicit modelling of the image formation process. More examples can be found in the supplementary.\vspace{-3mm}}
    \label{fig:qualitative}
\end{figure*}
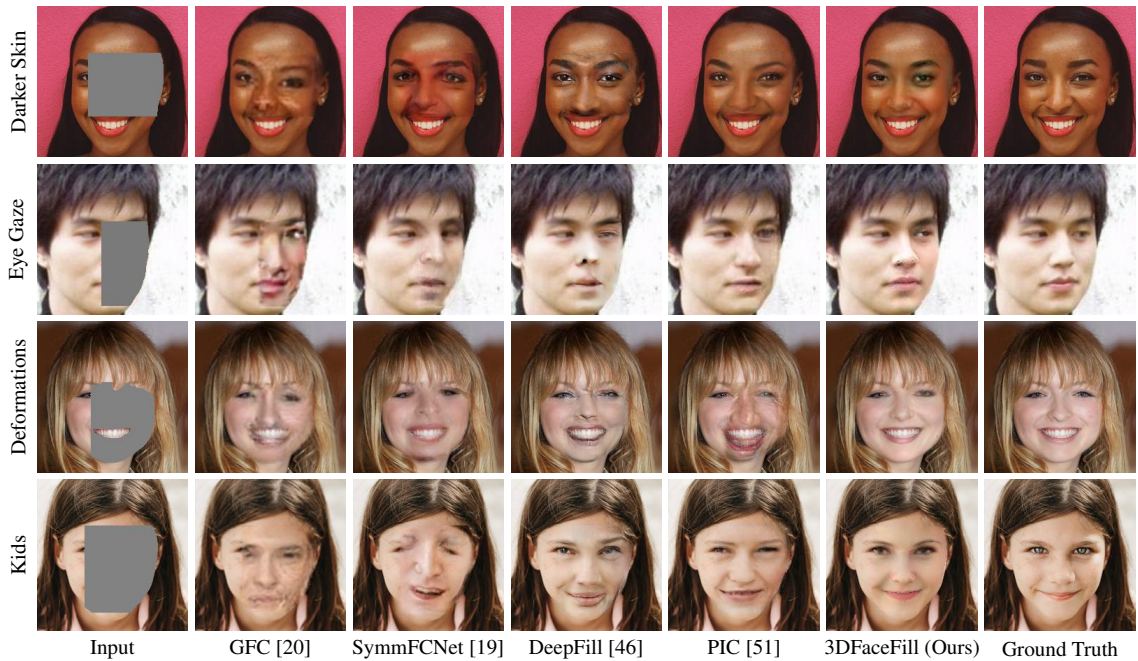

\begin{figure*}
    \footnotesize
    \centering
    \tikzstyle{block} = [rectangle, draw, fill=blue!20, text centered]
    \begin{tikzpicture}
        \node[rotate=90, text centered] (a00) {Input};
        \node[right of=a00, node distance=1.35cm] (a01) {\includegraphics[width=.1\textwidth,trim={30 30 30 30},clip]{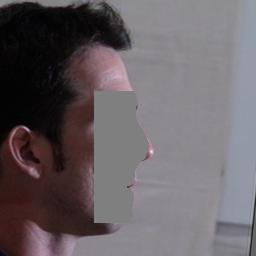}};
        \node[right of=a01, node distance=2.0cm] (a02) {\includegraphics[width=.1\textwidth,trim={30 30 30 30},clip]{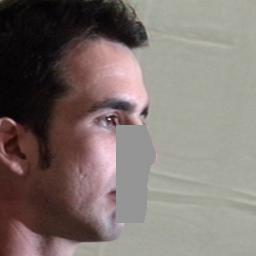}};
        \node[right of=a02, node distance=2.0cm] (a03) {\includegraphics[width=.1\textwidth,trim={30 30 30 30},clip]{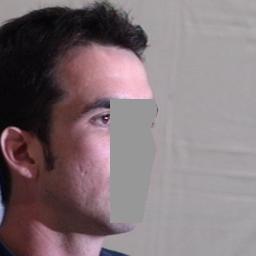}};
        \node[right of=a03, node distance=2.0cm] (a04) {\includegraphics[width=.1\textwidth,trim={30 30 30 30},clip]{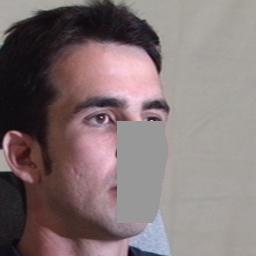}};
        \node[right of=a04, node distance=2.50cm] (a08) {\includegraphics[width=.1\textwidth,trim={30 30 30 30},clip]{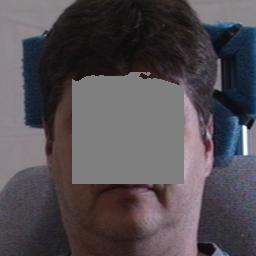}};
        \node[right of=a08, node distance=2.0cm] (a11) {\includegraphics[width=.1\textwidth,trim={30 30 30 30},clip]{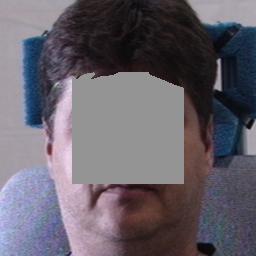}};
        \node[right of=a11, node distance=2.0cm] (a12) {\includegraphics[width=.1\textwidth,trim={30 30 30 30},clip]{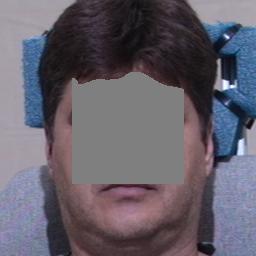}};
        \node[right of=a12, node distance=2.0cm] (a13) {\includegraphics[width=.1\textwidth,trim={30 30 30 30},clip]{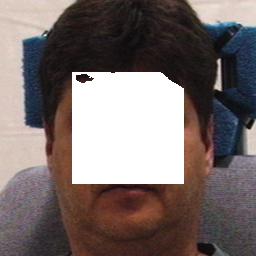}};
        
        \node[below of=a01, node distance=1.80cm] (b01) {\includegraphics[width=.1\textwidth,trim={30 30 30 30},clip]{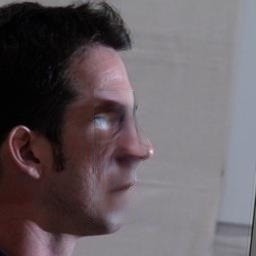}};
        \node[right of=b01, node distance=2.0cm] (b02) {\includegraphics[width=.1\textwidth,trim={30 30 30 30},clip]{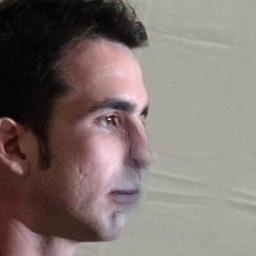}};
        \node[right of=b02, node distance=2.0cm] (b03) {\includegraphics[width=.1\textwidth,trim={30 30 30 30},clip]{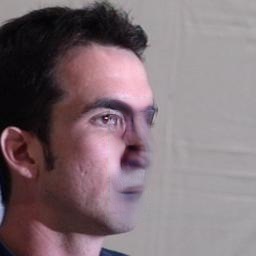}};
        \node[right of=b03, node distance=2.0cm] (b04) {\includegraphics[width=.1\textwidth,trim={30 30 30 30},clip]{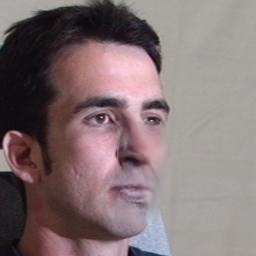}};
        \node[left of=b01, node distance=1.35cm, rotate=90, text centered] (b00) {DeepFillv2 \cite{freeforminpainting}};
        \node[right of=b04, node distance=2.50cm] (b08) {\includegraphics[width=.1\textwidth,trim={30 30 30 30},clip]{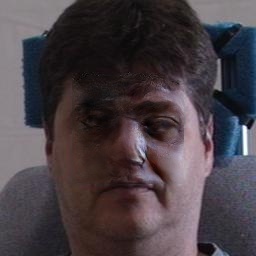}};
        \node[right of=b08, node distance=2.0cm] (b11) {\includegraphics[width=.1\textwidth,trim={30 30 30 30},clip]{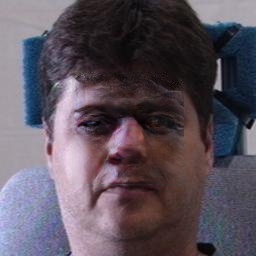}};
        \node[right of=b11, node distance=2.0cm] (b12) {\includegraphics[width=.1\textwidth,trim={30 30 30 30},clip]{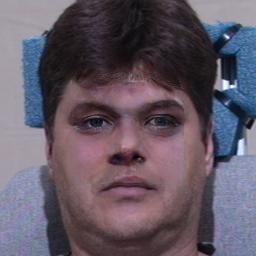}};
        \node[right of=b12, node distance=2.0cm] (b13) {\includegraphics[width=.1\textwidth,trim={30 30 30 30},clip]{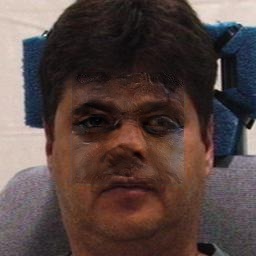}};
        
        \node[below of=b01, node distance=1.80cm] (c01) {\includegraphics[width=.1\textwidth,trim={30 30 30 30},clip]{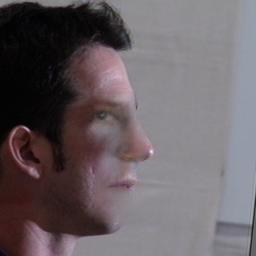}};
        \node[right of=c01, node distance=2.0cm] (c02) {\includegraphics[width=.1\textwidth,trim={30 30 30 30},clip]{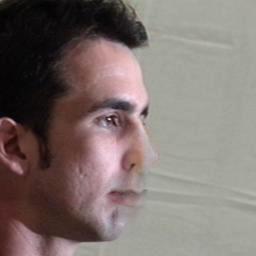}};
        \node[right of=c02, node distance=2.0cm] (c03) {\includegraphics[width=.1\textwidth,trim={30 30 30 30},clip]{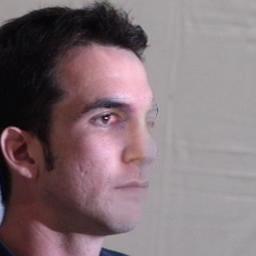}};
        \node[right of=c03, node distance=2.0cm] (c04) {\includegraphics[width=.1\textwidth,trim={30 30 30 30},clip]{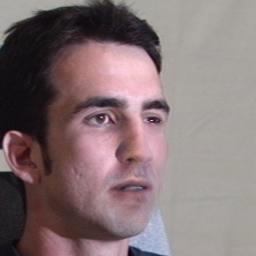}};
        \node[left of=c01, node distance=1.35cm, rotate=90, text centered] (c00) {PICNet \cite{zheng2019pluralistic}};
        \node[right of=c04, node distance=2.50cm] (c08) {\includegraphics[width=.1\textwidth,trim={30 30 30 30},clip]{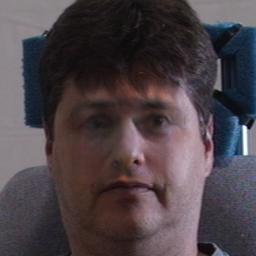}};
        \node[right of=c08, node distance=2.0cm] (c11) {\includegraphics[width=.1\textwidth,trim={30 30 30 30},clip]{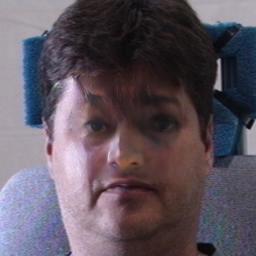}};
        \node[right of=c11, node distance=2.0cm] (c12) {\includegraphics[width=.1\textwidth,trim={30 30 30 30},clip]{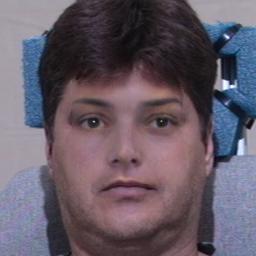}};
        \node[right of=c12, node distance=2.0cm] (c13) {\includegraphics[width=.1\textwidth,trim={30 30 30 30},clip]{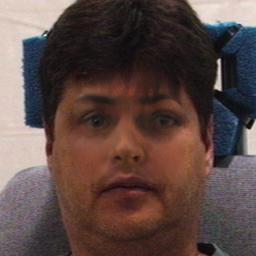}};
        
        \node[below of=c01, node distance=1.80cm] (d01) {\includegraphics[width=.1\textwidth,trim={30 30 30 30},clip]{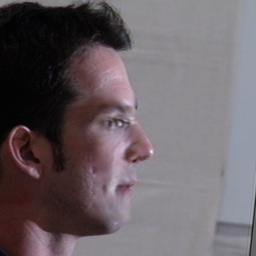}};
        \node[right of=d01, node distance=2.0cm] (d02) {\includegraphics[width=.1\textwidth,trim={30 30 30 30},clip]{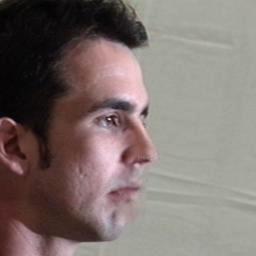}};
        \node[right of=d02, node distance=2.0cm] (d03) {\includegraphics[width=.1\textwidth,trim={30 30 30 30},clip]{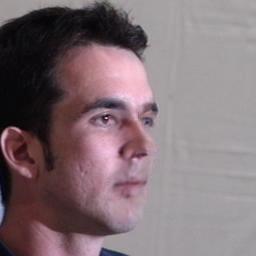}};
        \node[right of=d03, node distance=2.0cm] (d04) {\includegraphics[width=.1\textwidth,trim={30 30 30 30},clip]{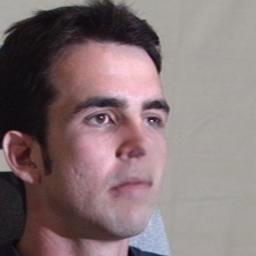}};
        \node[left of=d01, node distance=1.35cm, rotate=90, text centered] (d00) {\ourmethod{}};
        \node[right of=d04, node distance=2.50cm] (d08) {\includegraphics[width=.1\textwidth,trim={30 30 30 30},clip]{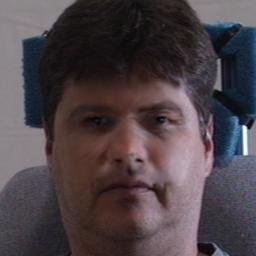}};
        \node[right of=d08, node distance=2.0cm] (d11) {\includegraphics[width=.1\textwidth,trim={30 30 30 30},clip]{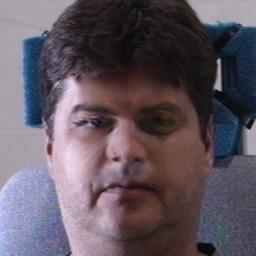}};
        \node[right of=d11, node distance=2.0cm] (d12) {\includegraphics[width=.1\textwidth,trim={30 30 30 30},clip]{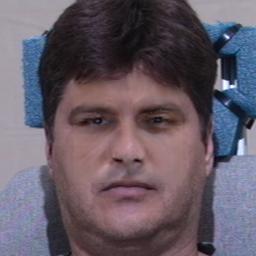}};
        \node[right of=d12, node distance=2.0cm] (d13) {\includegraphics[width=.1\textwidth,trim={30 30 30 30},clip]{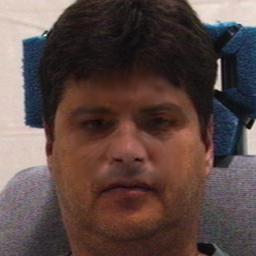}};
        
        \node[below of=d01, node distance=1.80cm] (e01) {\includegraphics[width=.1\textwidth,trim={30 30 30 30},clip]{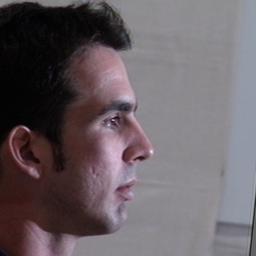}};
        \node[right of=e01, node distance=2.0cm] (e02) {\includegraphics[width=.1\textwidth,trim={30 30 30 30},clip]{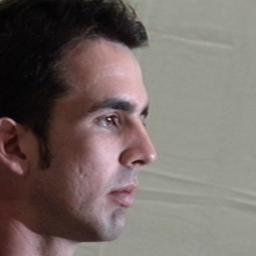}};
        \node[right of=e02, node distance=2.0cm] (e03) {\includegraphics[width=.1\textwidth,trim={30 30 30 30},clip]{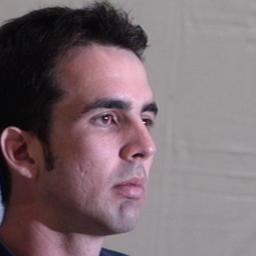}};
        \node[right of=e03, node distance=2.0cm] (e04) {\includegraphics[width=.1\textwidth,trim={30 30 30 30},clip]{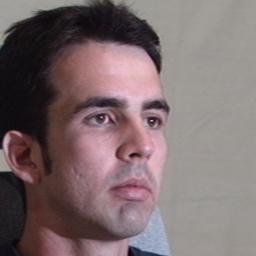}};
        \node[left of=e01, node distance=1.35cm, rotate=90, text centered] (d00) {Ground truth};
        \node[right of=e04, node distance=2.50cm] (e08) {\includegraphics[width=.1\textwidth,trim={30 30 30 30},clip]{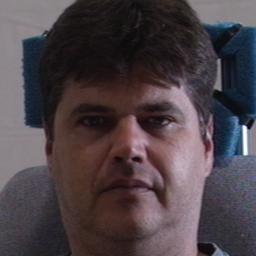}};
        \node[right of=e08, node distance=2.0cm] (e11) {\includegraphics[width=.1\textwidth,trim={30 30 30 30},clip]{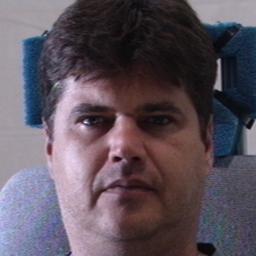}};
        \node[right of=e11, node distance=2.0cm] (e12) {\includegraphics[width=.1\textwidth,trim={30 30 30 30},clip]{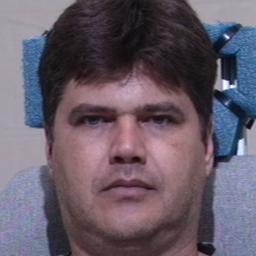}};
        \node[right of=e12, node distance=2.0cm] (e13) {\includegraphics[width=.1\textwidth,trim={30 30 30 30},clip]{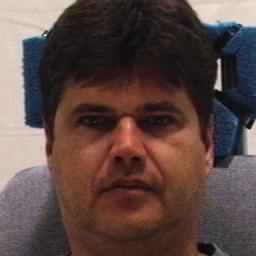}};
    \end{tikzpicture}
    \vspace{-1mm}
    \caption{\textbf{Qualitative evaluation on the MultiPIE \cite{multipie} dataset}. Compared to the baselines that generate deformed faces with artifacts in extreme poses and illumination, \ourmethod{} is more robust and generate geometrically accurate and illumination-preserving faces.\vspace{-3mm}}
    \label{fig:multipie}
\end{figure*}

\noindent\textbf{Datasets:} We evaluate the proposed \ourmethod{} on the CelebA \cite{celeba} and CelebA-HQ \cite{CelebAMask-HQ} datasets. We use 80\% split for training and 20\% for evaluation. Further, to evaluate the robustness and generalization performance, we do a cross-dataset evaluation on the pose and illumination varying images from the MultiPIE \cite{multipie} dataset and $\sim$50 in-the-wild face images downloaded from the internet\footnote{Source: \url{https://unsplash.com/s/photos/face}}.

\vspace{5pt}
\noindent\textbf{Implementation Details:} We train both the 3D factorization and the completion modules independently using the Adam optimizer with a learning rate of $10^{-4}$. We first train the 3DMM module on the 300W-3D \cite{facealignzhu} and the CelebA \cite{celeba} datasets. Once the 3DMM encoder is trained, we freeze it and use it to train the completion module on the CelebA \cite{celeba} dataset for 30k iterations. We generate random rectangular masks of varying sizes and locations, and constrain them to lie in the segmented face region (Fig.~\ref{fig:overview}). We provide further details on implementation and computational analysis in the supplementary.

\vspace{5pt}
\noindent\textbf{Baselines:} To evaluate the efficacy of \ourmethod{}, we perform qualitative and quantitative comparison against baselines such as GFC \cite{genfacecompletion}, SymmFCNet \cite{li2020symmetry}, DeepFillv2 \cite{freeforminpainting, gencontextualattn} and PICNet\footnote{Following author guidelines, we sample top 10 completions ranked by its discriminator and chose the one closest to the groudtruth for evaluation.} \cite{zheng2019pluralistic}. We use the publicly available pretrained face models for DeepFillv2 \cite{freeforminpainting}, PICNet \cite{zheng2019pluralistic} and SymmFCNet \cite{li2020symmetry}. For GFC \cite{genfacecompletion}, the pretrained model was not trained on the same crop and alignment as ours, so we train it from scratch using their source code. Due to the absense of extensive results, we present additional evaluation against baselines that do not provide source codes or pre-trained models in the supplementary, using a small set of results obtained from the corresponding authors.


\subsection{Results}
\noindent\textbf{Quantitative Evaluation:} In addition to the typically used PSNR and SSIM metrics, we report LPIPS \cite{lpips}, which is more suitable for image completion. Table~\ref{tab:quantitative} reports the overall values of these metrics across all image-mask pairs for each dataset. Overall \ourmethod{} improves PSNR by 2dB-3dB and LPIPS by 5-10\% over the closest baselines. In addition, for all the methods, we report PSNR and LPIPS as a function of mask to face area ratio $(\frac{\# MaskPixels}{\#Face Pixels})$ in Fig.~\ref{fig:celeba_quant}, \ref{fig:celebahq_quant} and \ref{fig:multipie_quant} for the CelebA, CelebA-HQ and Multi-PIE datasets, respectively. We make the following observations: (1) Across all the datasets, \ourmethod{} achieves significantly better PSNR and LPIPS across all mask ratios. (2) Among the baselines, PIC \cite{zheng2019pluralistic} and DeepFillV2 \cite{freeforminpainting} perform comparably with the former being slightly better in terms of LPIPS. (3) The effectiveness of \ourmethod{} over the baselines is more apparent as larger parts of the face are to be completed i.e., as the mask ratio increases. (4) On the CelebA dataset \cite{celeba}, the improvement ranges from $\sim$2dB PSNR for 0-10\% mask ratio to $\sim$4dB PSNR for 60-80\% mask ratio. In terms of LPIPS, the improvement ranges from 5\% for 0-10\% mask ratio to ~25\% for 60-90\% mask ratio. Similar trends are seen across the CelebA-HQ \cite{CelebAMask-HQ} and MultiPIE \cite{multipie} datasets too. These results confirm our hypothesis that explicitly modeling the image formation process leads to significantly better face completion. We provide addtional quantitative comparisons against PConv \cite{partialconv}, DSA \cite{oracleattention} and UVGAN \cite{uvgan} in the supplementary since these results are based on a limited number of author-provided completions in the absense of source codes.



\noindent\textbf{Qualitative Evaluation:}
Fig.~\ref{fig:qualitative} qualitatively compares face completion between \ourmethod{} and the baselines over a wide variety of challenging conditions. Completions by the baselines are less photorealistic and often contain artifacts in scenarios with dark complexion, tend to deform facial components (\eg nose) and fail to preserve symmetry (\eg eye-gaze or eye-brow shape). In addition, the baselines tend to deform the shape of small faces (\eg children) since they are mostly trained on adult faces where the relative proportions of facial parts differs significantly. In contrast, \ourmethod{} generates more photorealistic completions in all these cases (diverse conditions and mask types) due to explicit 3D shape modeling, incorporating symmetry priors and disentanglement of pose and illumination. 

\vspace{5pt}
\noindent\textbf{Cross-Dataset Evaluation:} To further demonstrate the improved generalization performance and robustness afforded by our method, we perform a cross-dataset comparison on the pose and illumination varying images from the MultiPIE~\cite{multipie} dataset, \textit{using models that were trained on the CelebA dataset \cite{celeba}}. Note that most baselines \cite{gencontextualattn, genfacecompletion, zheng2019pluralistic, oracleattention} do not perform such an evaluation. Quantitative results are in the last rows of Table~\ref{tab:quantitative}, while Fig.~\ref{fig:multipie} shows the qualitative results. Fig~\ref{fig:multipie} (left) shows that the baselines generate fuzzy and deformed faces while \ourmethod{} generates consistently superior completion across all poses. Similarly, for the varying illumination case (Fig~\ref{fig:multipie}-right), \ourmethod{} not only generates superior completion but also preserves the illumination contrast across the face.

\begin{figure}
    \vspace{-0.3em}
    \centering
    \begin{tikzpicture}

    
    \node [] (b1) {\includegraphics[width=0.2\linewidth,trim={0 20 0 20},clip]{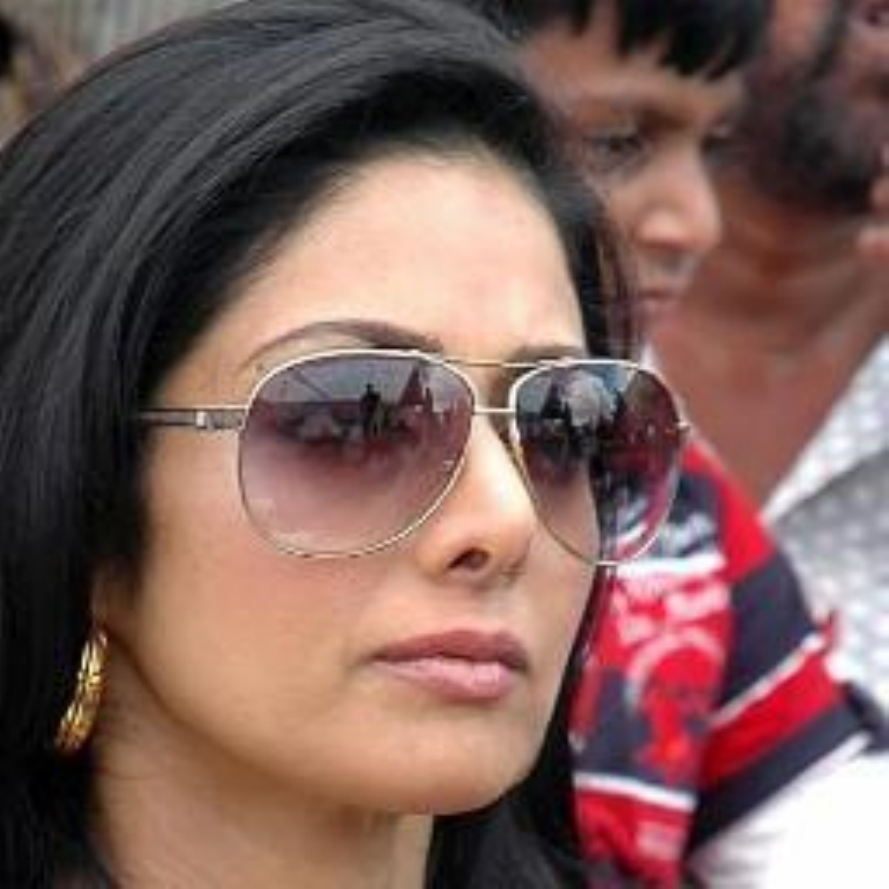}};
    \node[right of=b1, node distance=1.65cm] (b2) {\includegraphics[width=0.2\linewidth,trim={0 20 0 20},clip]{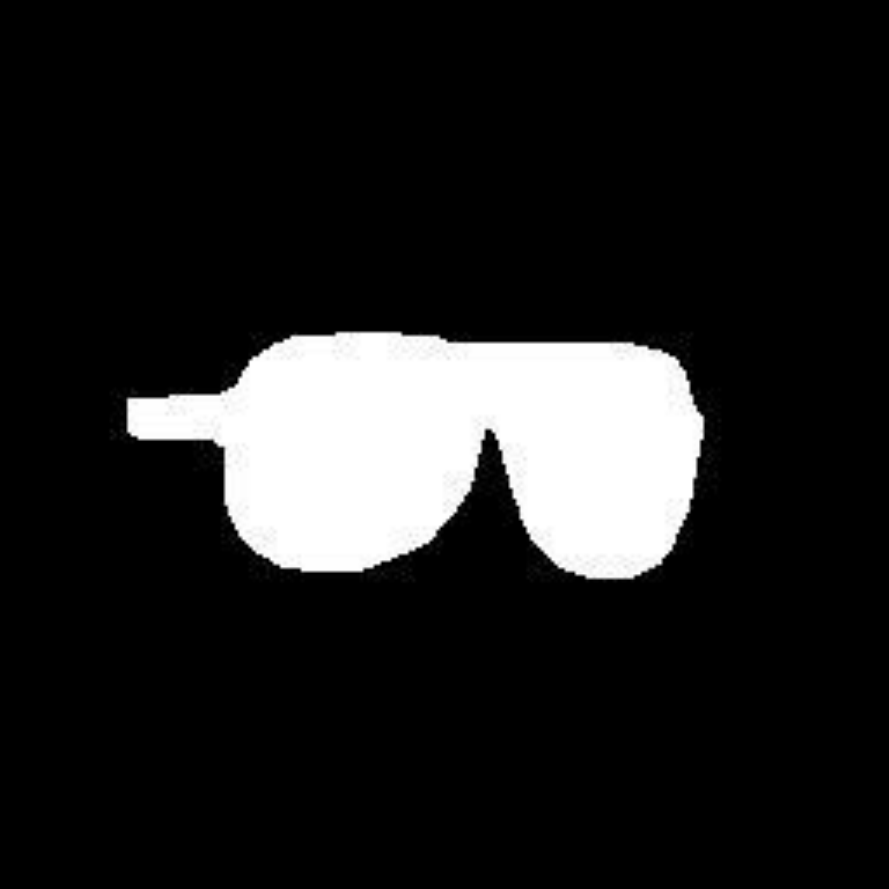}};
    \node[right of=b2, node distance=1.65cm] (b3) {\includegraphics[width=0.2\linewidth,trim={0 20 0 20},clip]{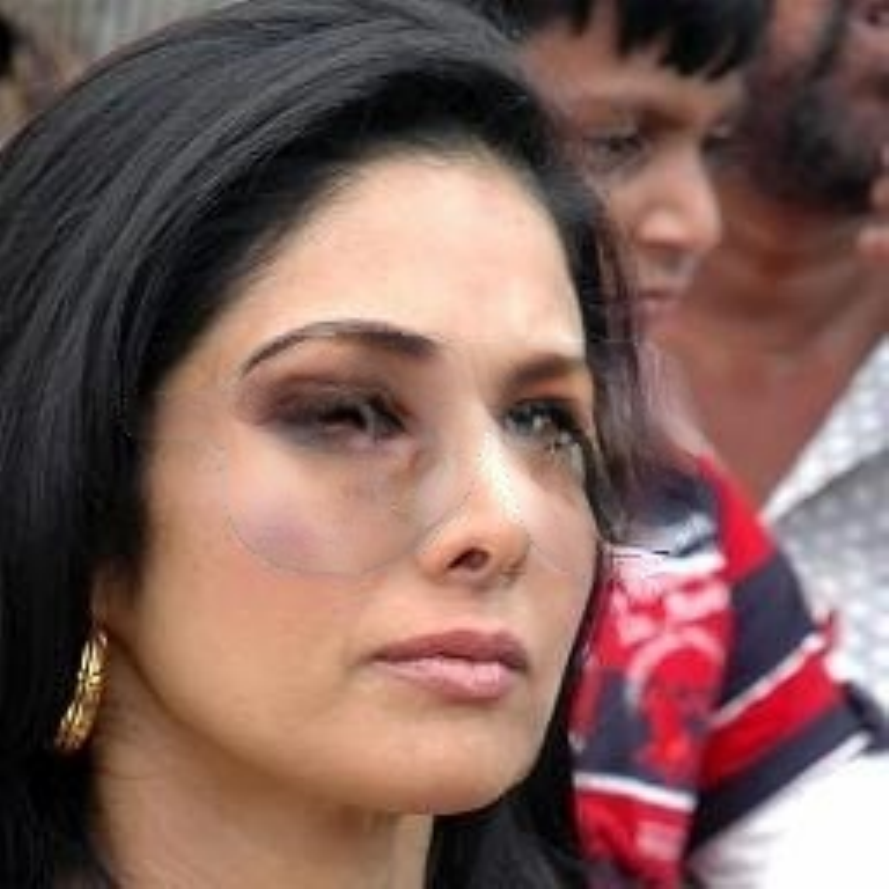}};
    \node[right of=b3, node distance=1.65cm] (b4) {\includegraphics[width=0.2\linewidth,trim={0 20 0 20},clip]{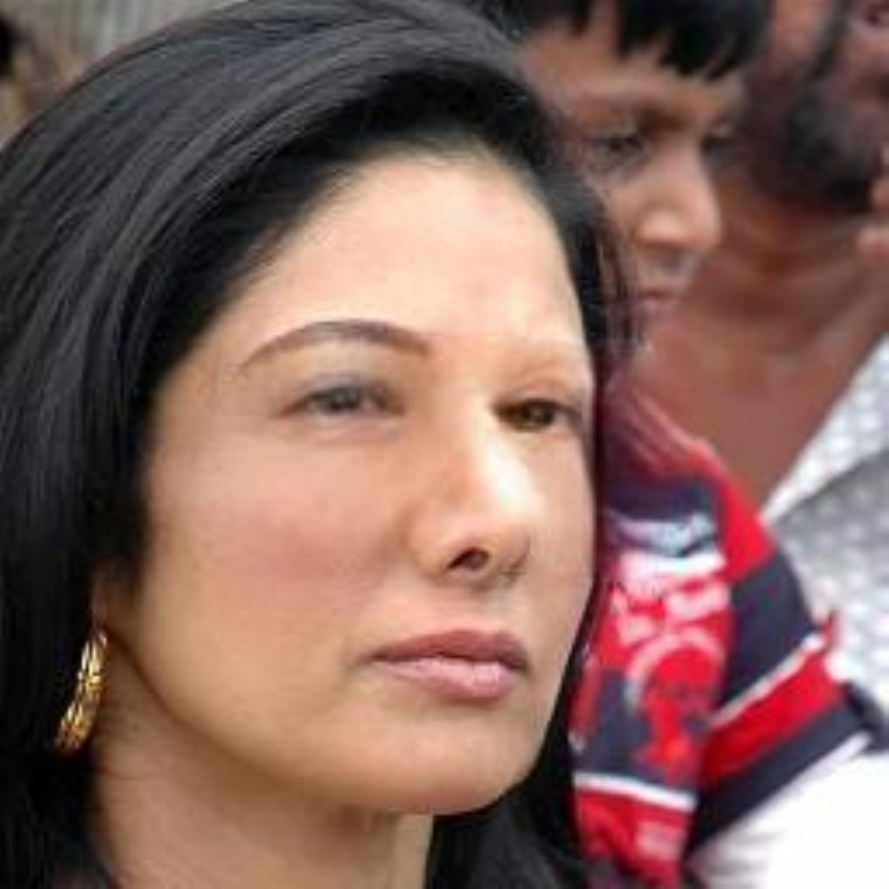}};
    \node[right of=b4, node distance=1.65cm] (b5) {\includegraphics[width=0.2\linewidth,trim={0 20 0 20},clip]{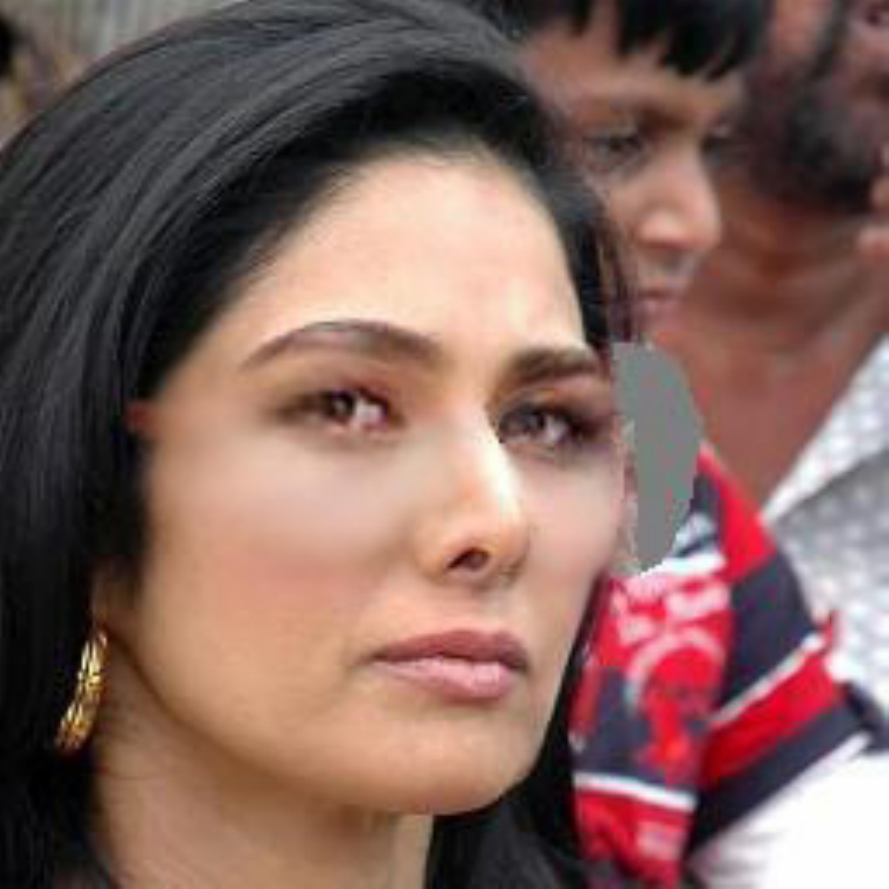}};

    
    
    
    \node [below of=b1, node distance=1.4cm] (d1) {\includegraphics[width=0.2\linewidth,trim={0 10 0 30},clip]{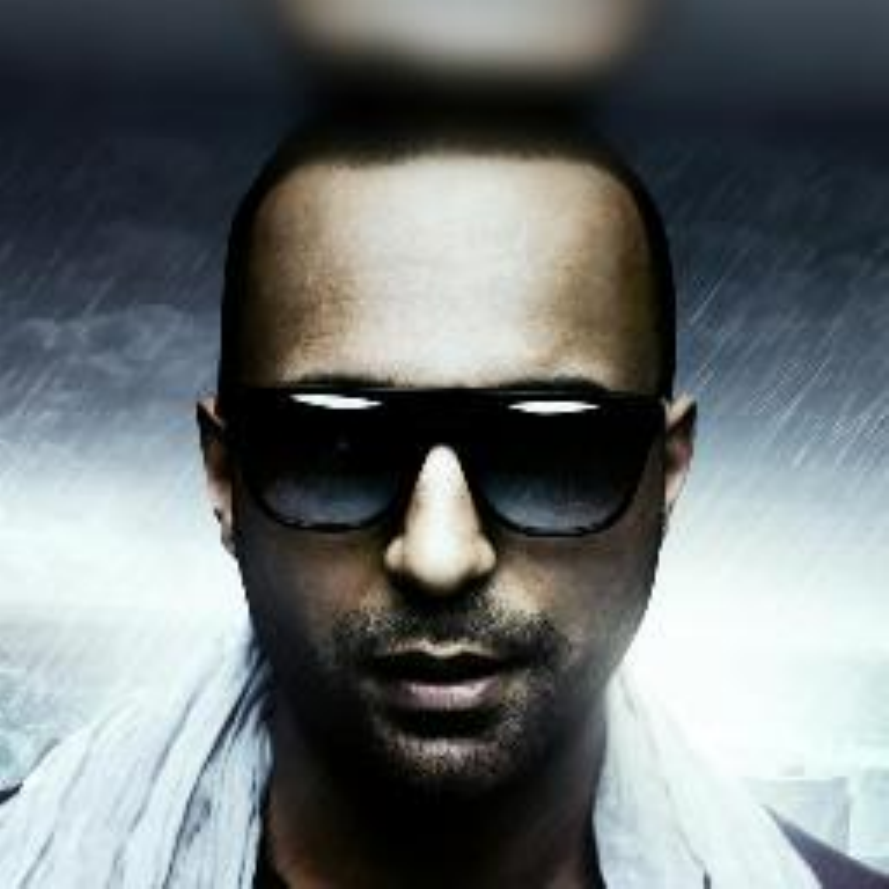}};
    \node[right of=d1, node distance=1.65cm] (d2) {\includegraphics[width=0.2\linewidth,trim={0 10 0 30},clip]{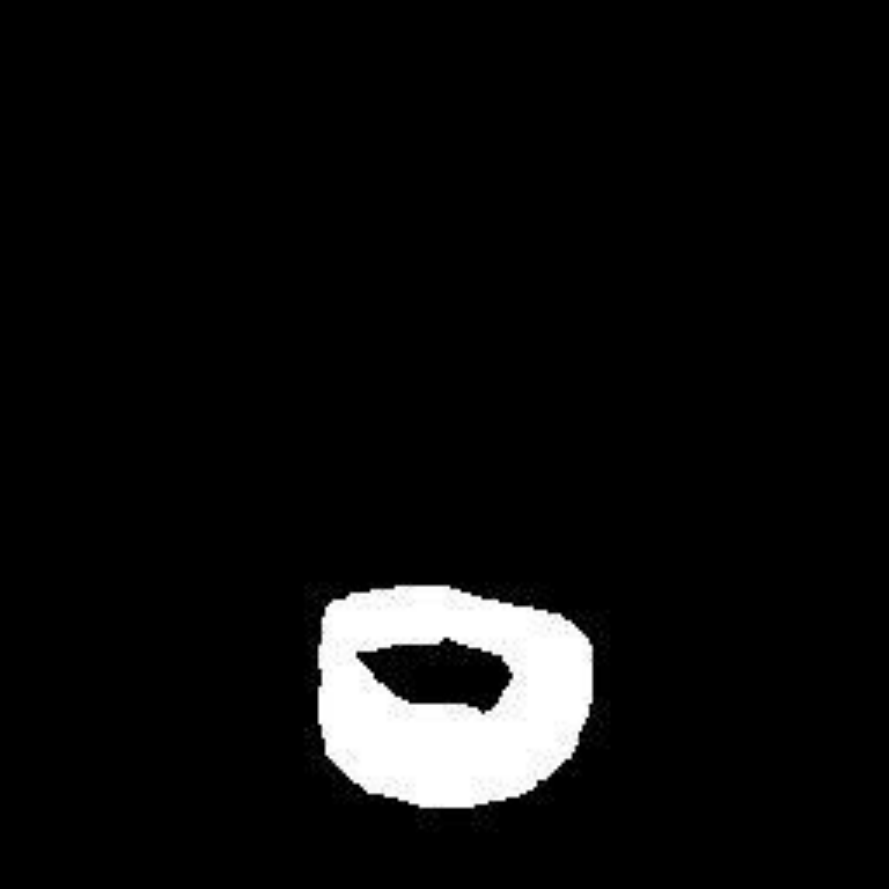}};
    \node[right of=d2, node distance=1.65cm] (d3) {\includegraphics[width=0.2\linewidth,trim={0 10 0 30},clip]{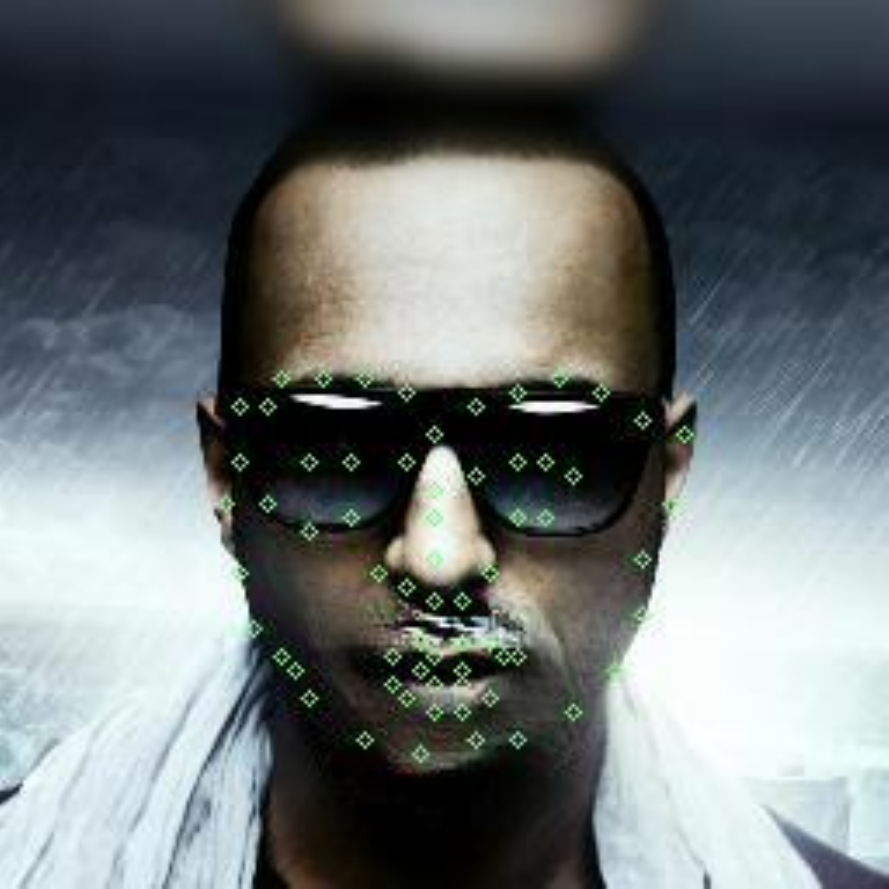}};
    \node[right of=d3, node distance=1.65cm] (d4) {\includegraphics[width=0.2\linewidth,trim={0 10 0 30},clip]{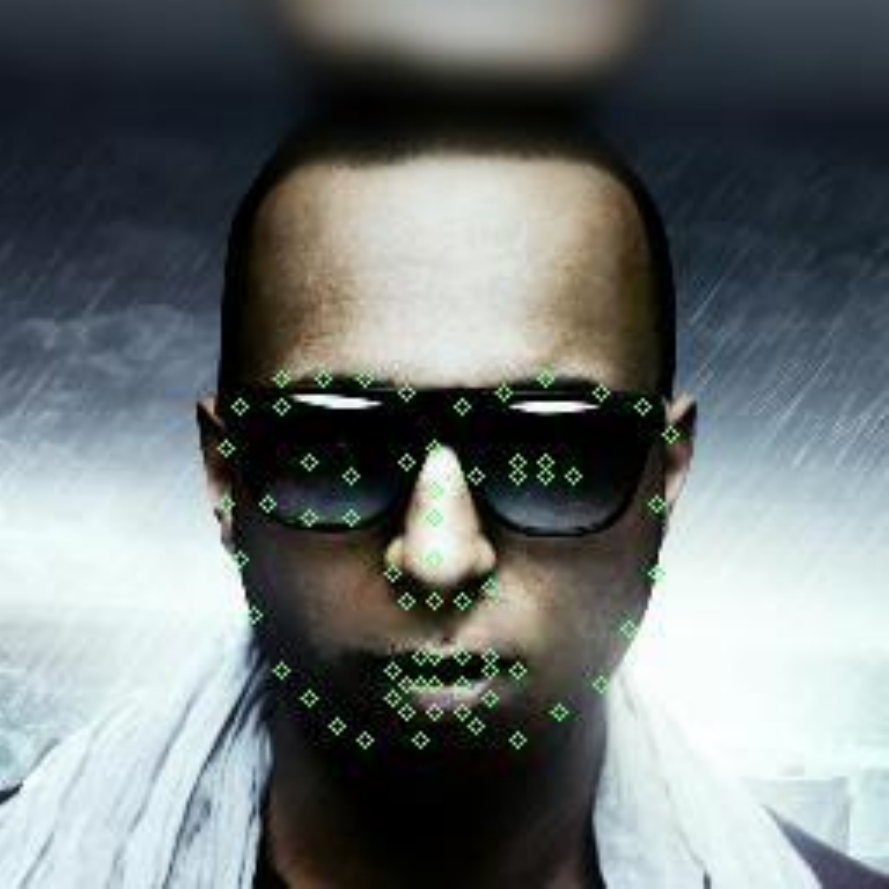}};
    \node[right of=d4, node distance=1.65cm] (d5) {\includegraphics[width=0.2\linewidth,trim={0 10 0 30},clip]{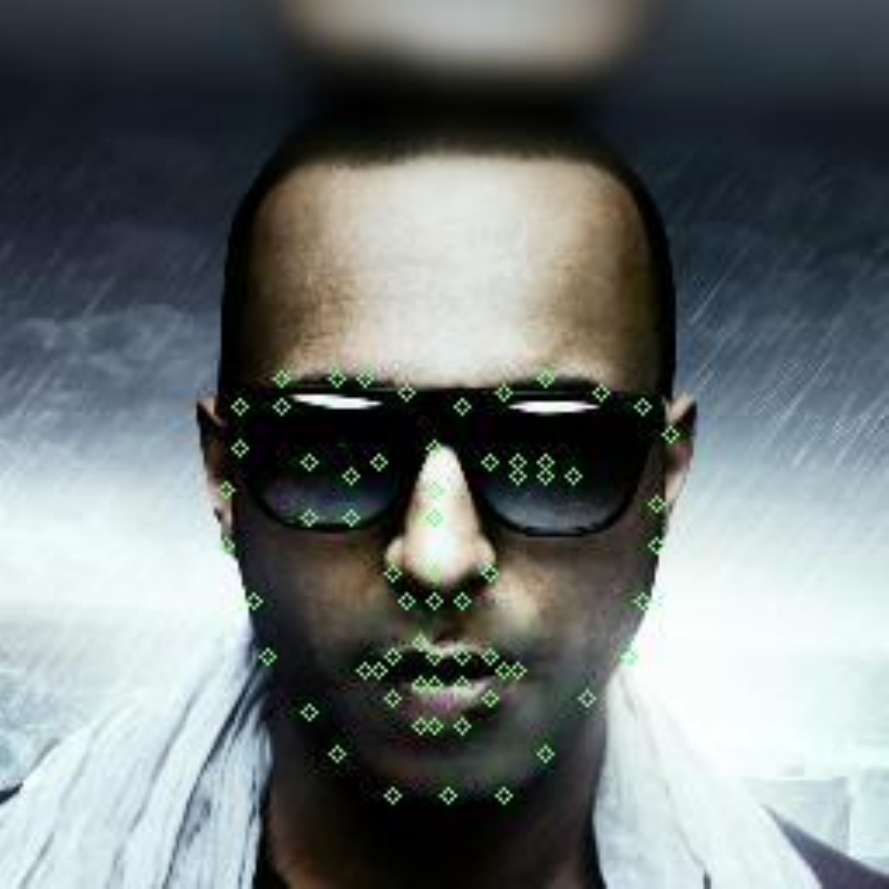}};
    
    
    \node[below of=d1, node distance=0.9cm] (e1) {\small Input};
    \node[below of=d2, node distance=0.86cm] (e2) {\small Mask};
    \node[below of=d3, node distance=0.9cm] (e3) {\small DeepFillv2 \cite{freeforminpainting}};
    \node[below of=d4, node distance=0.9cm] (e4) {\small PIC \cite{zheng2019pluralistic}};
    \node[below of=d5, node distance=0.9cm] (e5) {\small \ourmethod{}};
    
    \coordinate[below of=e1, node distance=0.9cm] (e1_below);
    \node[left of=e1_below, node distance=0.2cm] (f1) {\includegraphics[width=0.165\linewidth,trim={0 10 0 30},clip]{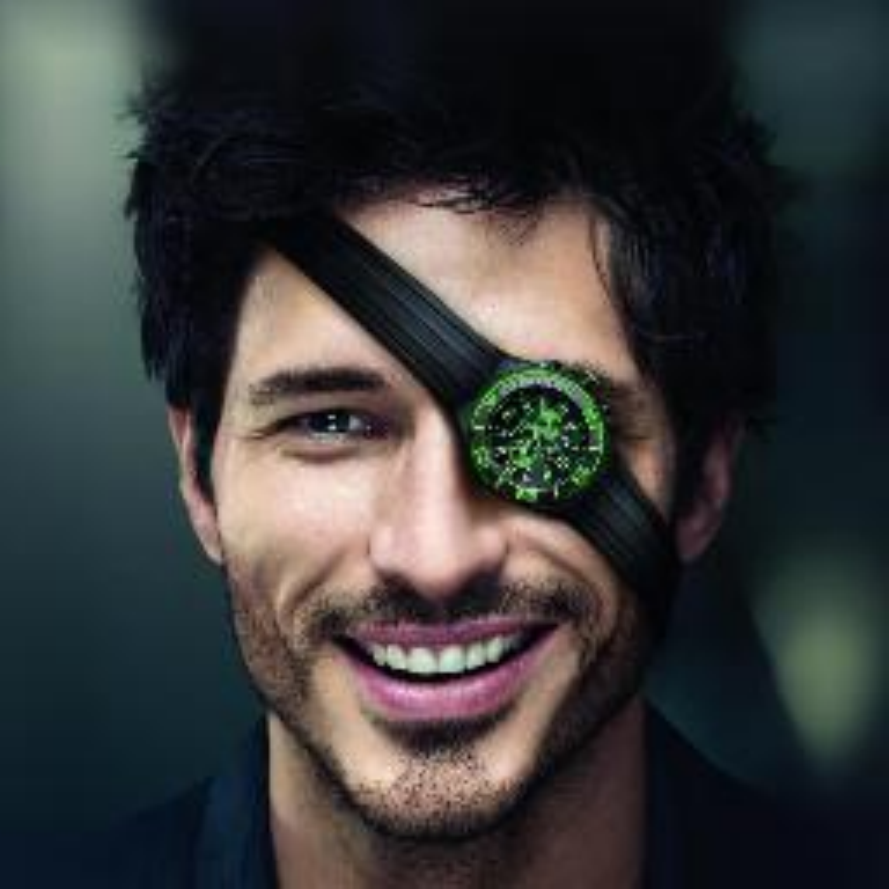}};
    \node[right of=f1, node distance=1.35cm] (f2) {\includegraphics[width=0.165\linewidth,trim={0 10 0 30},clip]{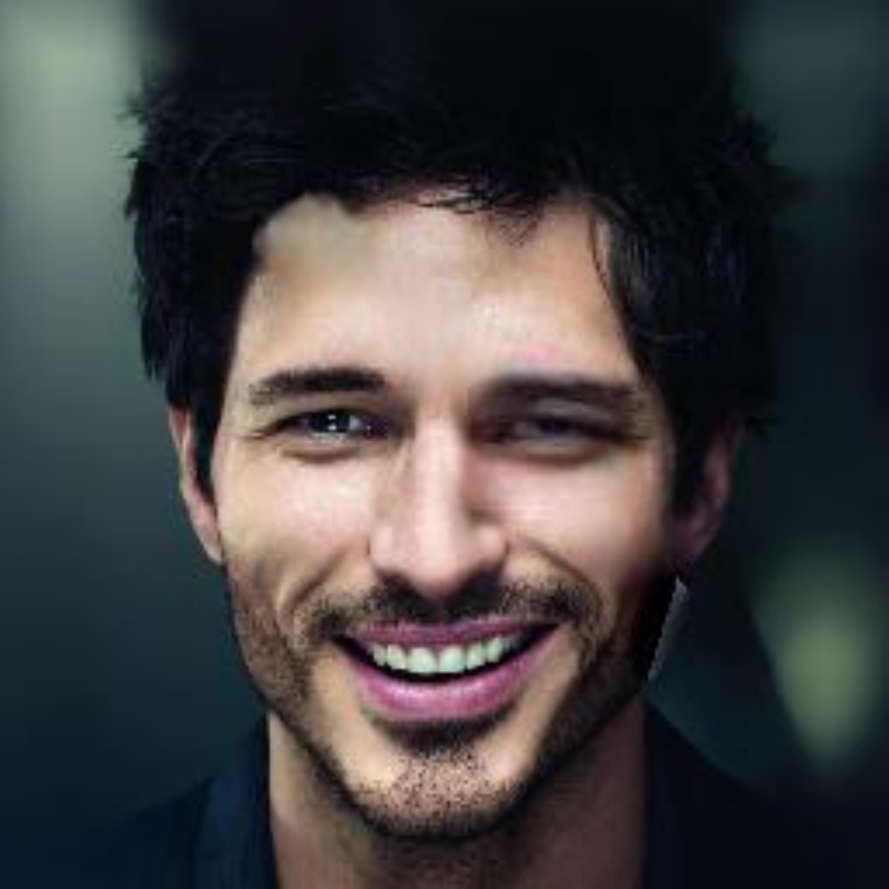}};
    
    \node[right of=f2, node distance=1.42cm] (f3) {\includegraphics[width=0.165\linewidth,trim={0 10 0 30},clip]{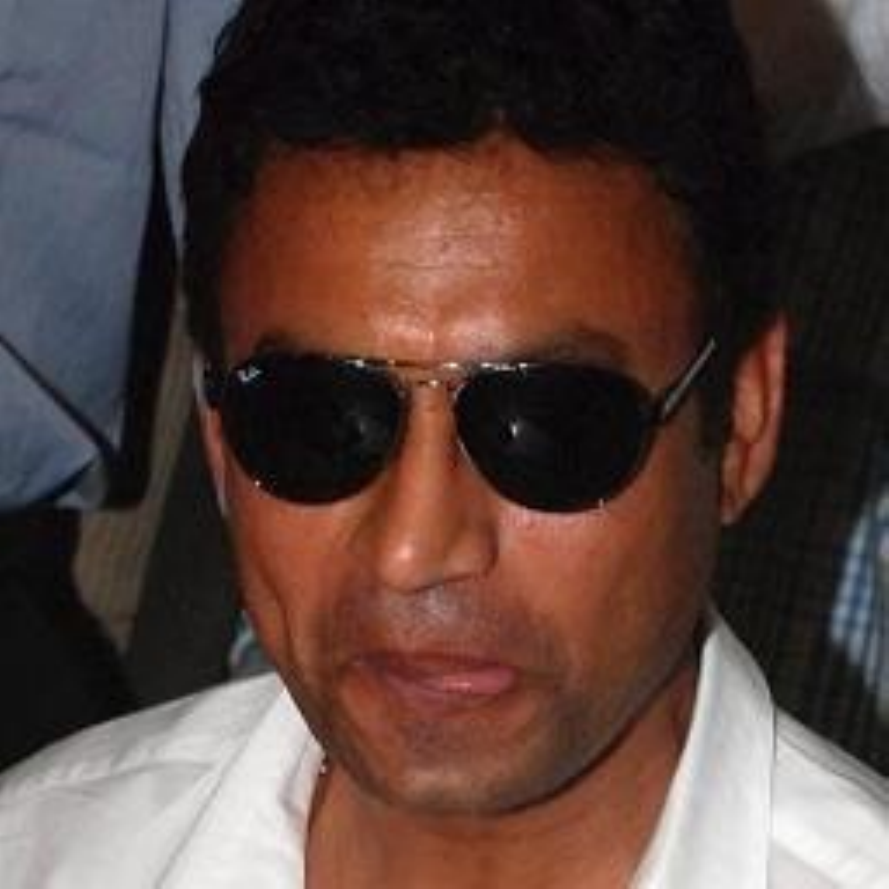}};
    \node[right of=f3, node distance=1.35cm] (f4) {\includegraphics[width=0.165\linewidth,trim={0 10 0 30},clip]{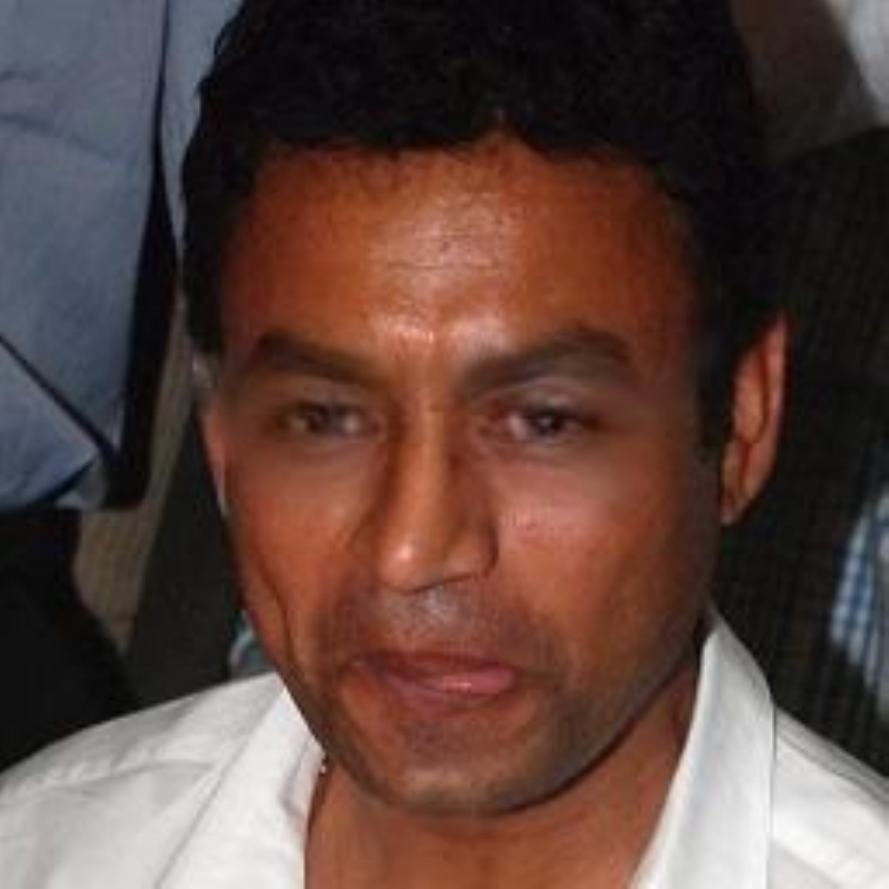}};
    
    \node[right of=f4, node distance=1.45cm] (f5) {\includegraphics[width=0.165\linewidth,trim={0 10 0 30},clip]{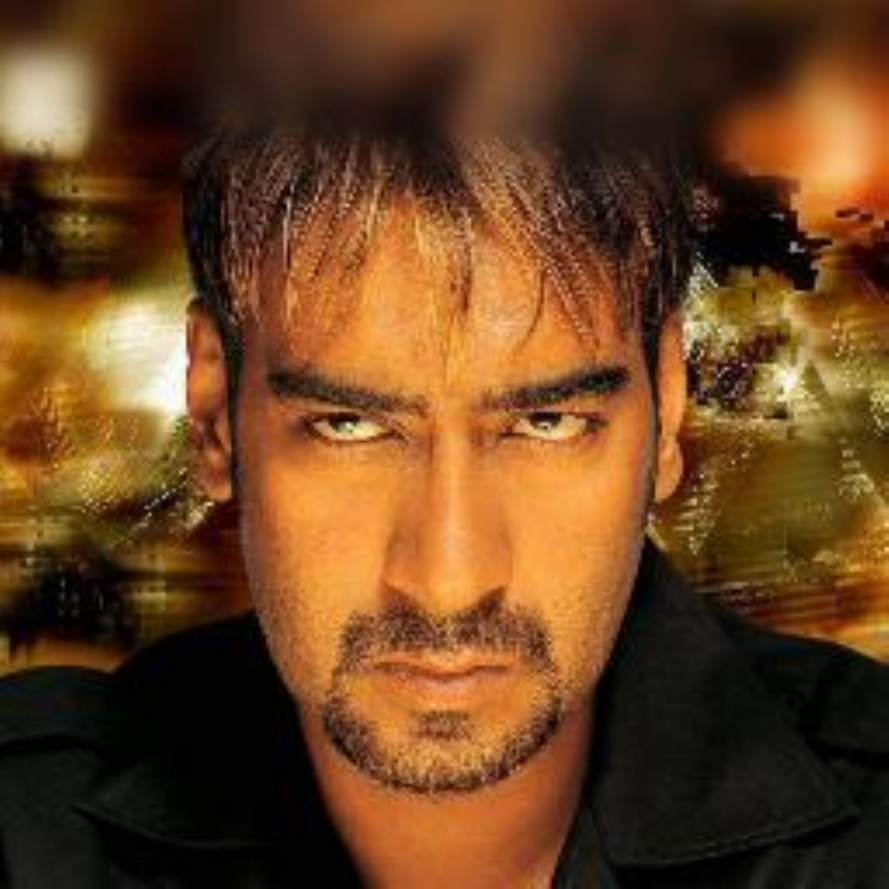}};
    \node[right of=f5, node distance=1.35cm] (f6) {\includegraphics[width=0.165\linewidth,trim={0 10 0 30},clip]{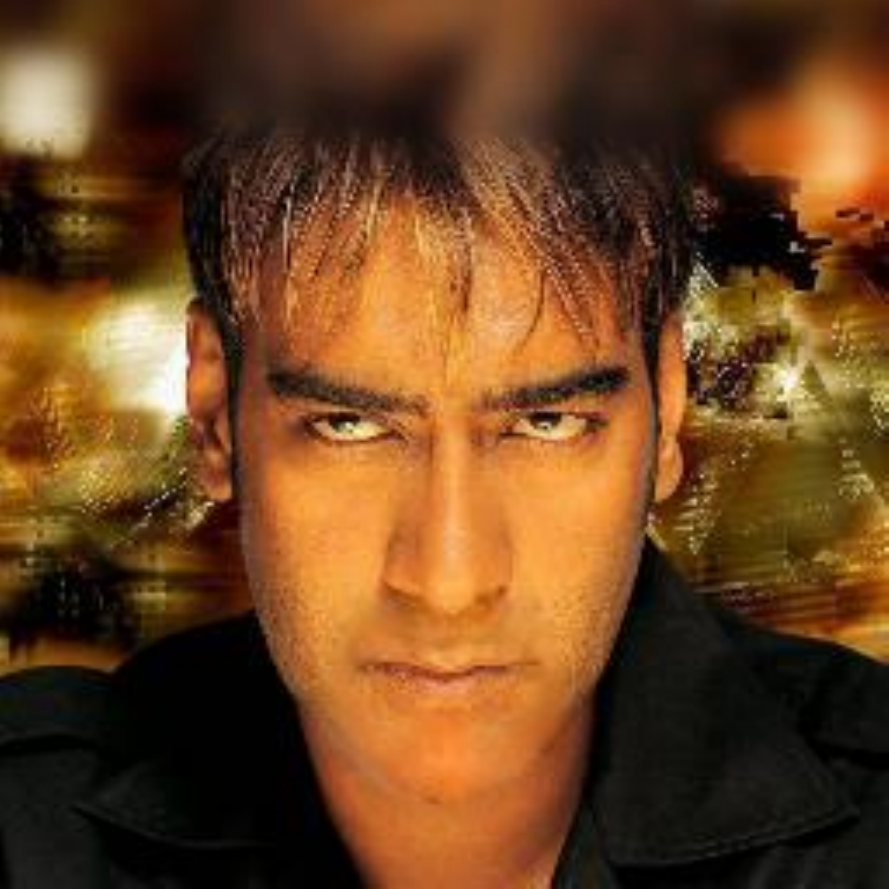}};
    
    \node[below of=f1, node distance=0.83cm] (g1) {\small Input};
    \node[below of=f2, node distance=0.8cm] (g2) {\small \ourmethod{}};
    \node[below of=f3, node distance=0.83cm] (g3) {\small Input};
    \node[below of=f4, node distance=0.8cm] (g4) {\small \ourmethod{}};
    \node[below of=f5, node distance=0.83cm] (g5) {\small Input};
    \node[below of=f6, node distance=0.8cm] (g6) {\small \ourmethod{}};
    
    \end{tikzpicture}
    \vspace{-5mm}
    \caption{\textbf{FC on real occlusions}. Note the asymmetric eye-gaze (row 1) and blurry shape (row 2) by the baselines.\vspace{-3mm}}
    \label{fig:real_world_compare}
\end{figure}
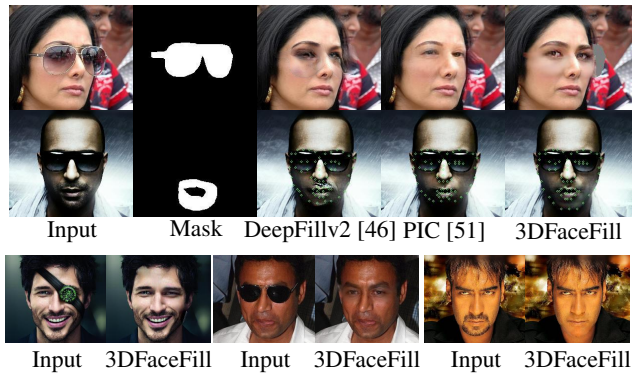

\vspace{5pt}
\noindent \textbf{Real Occlusions:} One of the potential applications of face completion is in de-occlusion. This is usually challenging when faces have large pose, illumination or shape variations. Fig.~\ref{fig:real_world_compare} shows a few real-world de-occlusion examples of faces in such conditions. Notice that, in cases of challenging pose, illumination, \etc, the baselines tend to generate blurry and asymmetric face completions, whereas \ourmethod{} does more realistic de-occlusion.

\subsection{Ablation Studies}

\begin{figure}
    \centering
    \captionsetup[sub]{font=footnotesize, labelformat = empty}
    \begin{subfigure}{.134\linewidth}
        \includegraphics[width=\textwidth,,trim={20 20 20 20},clip]{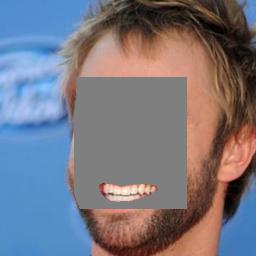}
        \caption{\footnotesize Input}
    \end{subfigure}
    \begin{subfigure}{.134\linewidth}
        \includegraphics[width=\textwidth,,trim={20 20 20 20},clip]{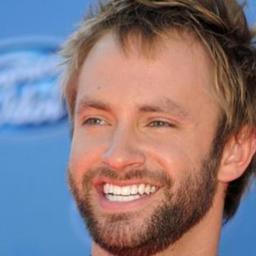}
        \caption{\footnotesize Original}
    \end{subfigure}
    \begin{subfigure}{.134\linewidth}
        \includegraphics[width=\textwidth,,trim={20 20 20 20},clip]{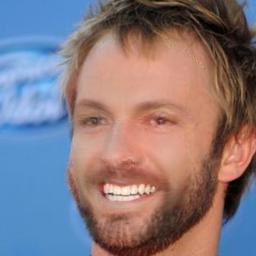}
        \caption{\footnotesize Iter1}
    \end{subfigure}
    \begin{subfigure}{.134\linewidth}
        \includegraphics[width=\textwidth,,trim={20 20 20 20},clip]{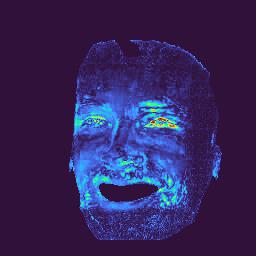}
        \caption{\footnotesize Iter1-Orig}
    \end{subfigure}
    \begin{subfigure}{.134\linewidth}
        \includegraphics[width=\textwidth,,trim={20 20 20 20},clip]{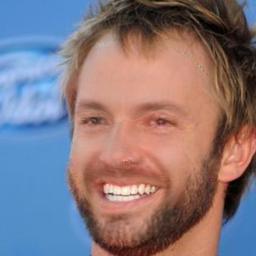}
        \caption{\footnotesize Iter2}
    \end{subfigure}
    \begin{subfigure}{.134\linewidth}
        \includegraphics[width=\textwidth,,trim={20 20 20 20},clip]{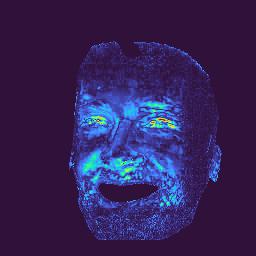}
        \caption{\footnotesize Iter2-Orig}
    \end{subfigure}
    \begin{subfigure}{.134\linewidth}
        \includegraphics[width=\textwidth,,trim={20 20 20 20},clip]{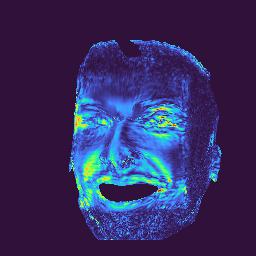}
        \caption{\footnotesize Iter2-Iter1}
    \end{subfigure}
    \vspace{-2mm}
    \caption{Visualization of raw completion (without blending) at iterations 1 and 2 along with the difference heatmaps.}
    \label{fig:ablation_iter_main}
\end{figure}

\noindent\textbf{Iterative Refinement:} To evaluate the effectiveness of iteratively refining face completion at inference, we compare the PSNR, SSIM and LPIPS \cite{lpips} metrics on raw output images (before blending with the visible image) at each iteration. As reported in Table~\ref{tab:ablation_iter_quant}, iteration 2 significantly improves upon iteration 1 over all the metrics. After iteration 2, the metrics become more or less stable, with a slight dip in performance. We hypothesize that it is a result of not training the model for iterative refinement and only performing it at inference. Further, we visualize the absolute difference heatmaps between the completed and the original image for both iterations 1 and 2 in Fig.~\ref{fig:ablation_iter_main} to understand which parts of the face benefit most from refinement. Observe that the largest differences are around the high-detail regions (eyes, beards, \etc), which we ascribe to more accurate 3D pose and shape estimation from the completed face after iteration 1 than from the partial face before.

\begin{table}
    \centering
    \resizebox{\columnwidth}{!}{\begin{tabular}{ccccccc}
         & Iter 1 & Iter 2 & Iter 3 & Iter 4 & Iter 5 & Iter 6 \\
         \hline
         \textbf{PSNR} $(\uparrow)$ & 33.7587 & \textbf{34.5347} & 34.5018 & 34.4943 & 34.4428 & 34.4018\\
         \textbf{SSIM} $(\uparrow)$ & 0.9510 & \textbf{0.9678} & 0.9675 & 0.9670 & 0.9666 & 0.9652\\
         \textbf{LPIPS} $(\downarrow)$ & 0.0192 & \textbf{0.0185} & 0.0186 & 0.0187 & 0.0188 & 0.0188\\
         \hline
    \end{tabular}}
    \caption{Quantitative evaluation of iterative refinement.}
    \label{tab:ablation_iter_quant}
\end{table}

\begin{table}
    \small
    \centering
    \resizebox{\columnwidth}{!}{\begin{tabular}{cccccc}
    \hline
    \textbf{Metric} & Full GAN & Patch GAN & NoSym & NoSym+Attn & Full Model\\
    \hline
    \textbf{PSNR} $(\uparrow)$ & 31.7125 & 31.7552 & 31.6110 & 31.7969 & \textbf{32.1950}\\
    \textbf{SSIM} $(\uparrow)$ & 0.9654 & 0.9658 & 0.9665 & 0.9667 & \textbf{0.9678}\\
    \textbf{LPIPS} $(\downarrow)$ & 0.0462 & 0.0454 & 0.0446 & 0.0442 & \textbf{0.0410}\\
    \hline
    \end{tabular}}
    \caption{Quantitative evaluation between the different ablation models and our full model on masks blocking one-half of the face.}
    \label{tab:ablation_quant}
\end{table}

\begin{figure}
    \centering
    \captionsetup[sub]{font=footnotesize, labelformat = empty}
    \begin{subfigure}{.175\linewidth}
        \includegraphics[width=\textwidth,,trim={20 20 20 20},clip]{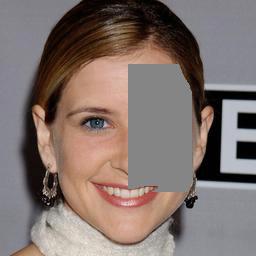}
        \caption{\footnotesize Input}
    \end{subfigure}
    \begin{subfigure}{.175\linewidth}
        \includegraphics[width=\textwidth,,trim={20 20 20 20},clip]{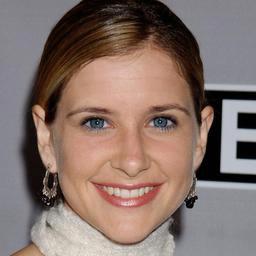}
        \caption{\footnotesize Original}
    \end{subfigure}
    \begin{subfigure}{.175\linewidth}
        \includegraphics[width=\textwidth,,trim={20 20 20 20},clip]{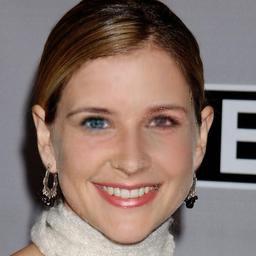}
        \caption{\footnotesize NoSym}
    \end{subfigure}
    \begin{subfigure}{.175\linewidth}
        \includegraphics[width=\textwidth,,trim={20 20 20 20},clip]{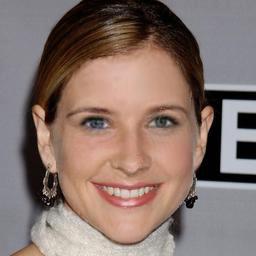}
        \caption{\footnotesize Full Model}
    \end{subfigure}
    \begin{subfigure}{.175\linewidth}
        \includegraphics[width=\textwidth,,trim={20 20 20 20},clip]{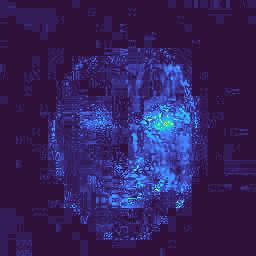}
        \caption{\footnotesize Full-NoSym}
    \end{subfigure}
    \vspace{-2mm}
    \caption{Visualizing the effect of symmetry for face inpainting. The full model includes Sym-UNet and symmetry loss (during training) and can copy symmetric features when available. The absolute difference heatmaps (Full-NoSym) shows that most difference is coming from components such as eyes, eye-brows, \etc \vspace{-25pt}}
    \label{fig:ablation_symmetry_main}
\end{figure}


\vspace{5pt}
\noindent\textbf{Symmetry Constraint:} To evaluate the effectiveness of Sym-UNet and the symmetry loss, we compare two variants of our full model. These include, (1) \textbf{NoSym:} Sym-UNet replaced by standard UNet and with no symmetry loss, and (2) \textbf{NoSym+Attn:} NoSym model plus a self-attention layer after the 3rd upsampling layer in the UNet decoder. Attention layers are commonly employed by many inpainting models \cite{gencontextualattn,freeforminpainting,zheng2019pluralistic} for capturing long-range spatial dependencies, so this variant seeks to compare the utility of attention in lieu of symmetry priors for face inpainting. To best evaluate the benefit of symmetry constraints for faces, the above model variations are evaluated on face images masked on one side of the face as shown in Fig.~\ref{fig:ablation_symmetry_main}.

The results in Table \ref{tab:ablation_quant} indicate that the full model outperforms all the variants, with NoSym being the worst among them. Also the NoSym+Attn variant does perform slightly better than NoSym but is still far behind the full model. This indicates that, (i) though attention helps in the absence of any prior constraints, explicitly enforcing geometric priors associated with structured objects like faces is significantly more effective than implicitly learning them through attention, and (ii) symmetry is a more useful feature for face inpainting and behaves like an attention on the visible symmetric parts. As shown in Fig.~\ref{fig:ablation_symmetry_main}, compared to the full model, the NoSym variant results in larger inpainting errors as indicated by the difference heatmaps. Therefore, unlike the full model the NoSym model tends to ignore the visible symmetric regions of the face leading to inconsistencies between the visible and inpainted regions.

\subsection{Discussions}
The above described experiments and ablation studies demonstrate the effectiveness of \ourmethod{}, along with the utility of each of its components in performing robust face completion in challenging cases of facial pose, shape, illumination, \etc. However, the formulation of our proposed approach do impose a dependency on the fidelity of the underlying 3D model. Essentially, our approach cannot inpaint on regions which are not included in the underlying 3D model and the resolution of inpainting depends on the density of the 3D mesh. \ourmethod{} currently uses the BFM model \cite{bfm}, thanks to its widespread support. However, BFM \cite{bfm} does not include the inner mouth, hairs and the upper head and has limited vertex density around the eyes, which restricts inpainting in these regions. However, these limitations of the underlying 3D model are not inherent to the proposed approach and do not invalidate the advantages of our model in improving the geometric and photometric consistency of completion. Furthermore, these limitations can potentially be mitigated by substituting BFM with a more detailed 3D face model, such as the Universal Head Model (UHM) \cite{uhm}, that includes the inner mouth and detailed eye-balls, along with other improvements.




\section{Conclusion} 
In this paper, we proposed \ourmethod{}, a 3D-aware face completion method. Our solution was driven by the hypothesis that performing face completion on the UV representation, as opposed to 2D pixel representation, will allow us to effectively leverage the power of 3D correspondence and ultimately lead to face completions that are geometrically and photometrically more accurate. Experimental evaluation across multiple datasets and against multiple baselines show that face completions from \ourmethod{} are significantly better, both qualitatively and quantitatively, under large variations in pose, illumination, shape and appearance. These results validate our primary hypothesis.

\appendix
\appendixpage
\section{Experimental Results}\label{sec:exp}

\begin{table}
    \centering
    \begin{tabular}{cccc}
        \hline
         & \textbf{PSNR} ($\uparrow$) & \textbf{SSIM} ($\uparrow$) & \textbf{LPIPS} \cite{lpips} ($\downarrow$) \\
         \hline
        DSA \cite{oracleattention} & 28.6205 & 0.9375 & 0.0436\\
        PConv \cite{partialconv} &  29.3067 & 0.9479 & 0.0379\\
        \ourmethod{} & \textbf{31.8823} & \textbf{0.9615} & \textbf{0.0335}\\
        \hline
    \end{tabular}
    \caption{\textbf{Quantitative comparison} of the proposed \ourmethod{} \versus{} PConv \cite{partialconv} and DSA \cite{oracleattention} on a small set of completed images obtained from the authors.}
    \label{tab:pconv}
\end{table}

\begin{figure*}
    \footnotesize
    \centering
    
    \tikzstyle{block} = [rectangle, draw, fill=blue!20, text centered]
    \begin{tikzpicture}

    \node[text width=3cm] (a0) {Missing eye-brows};
    \node[right of=a0, node distance=2.8cm] (a1) {\includegraphics[width=.16\textwidth]{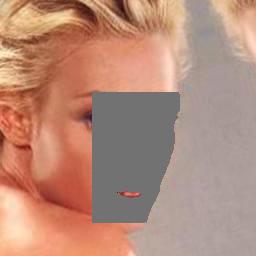}};
    \node[right of=a1, node distance=2.8cm] (a2) {\includegraphics[width=.16\textwidth]{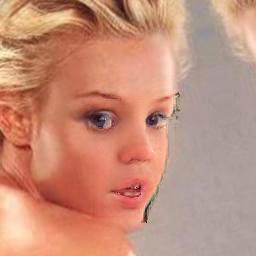}};
    \node[right of=a2, node distance=2.8cm] (a3) {\includegraphics[width=.16\textwidth]{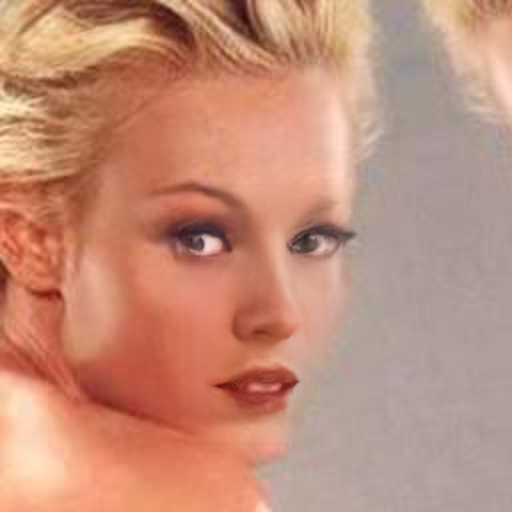}};
    \node[right of=a3, node distance=2.8cm] (a4) {\includegraphics[width=.16\textwidth]{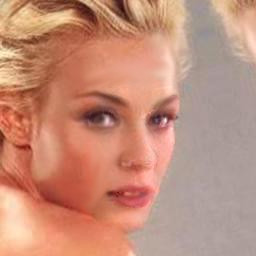}};
    \node[right of=a4, node distance=2.8cm] (a5) {\includegraphics[width=.16\textwidth]{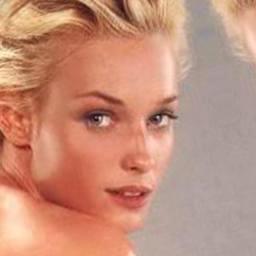}};
    
    \node[below of=a1, node distance=2.8cm] (b1) {\includegraphics[width=.16\textwidth]{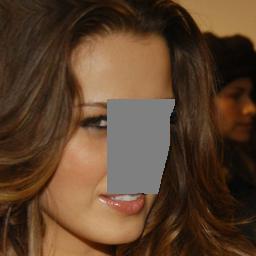}};
    \node[right of=b1, node distance=2.8cm] (b2) {\includegraphics[width=.16\textwidth]{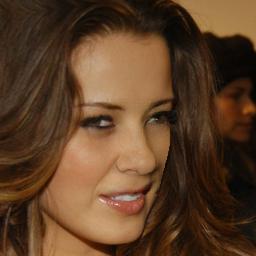}};
    \node[right of=b2, node distance=2.8cm] (b3) {\includegraphics[width=.16\textwidth]{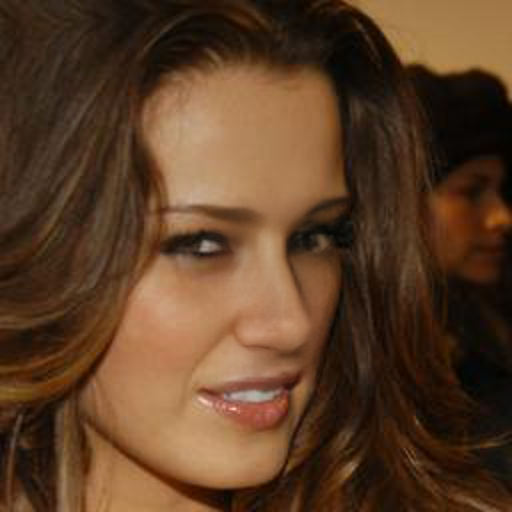}};
    \node[right of=b3, node distance=2.8cm] (b4) {\includegraphics[width=.16\textwidth]{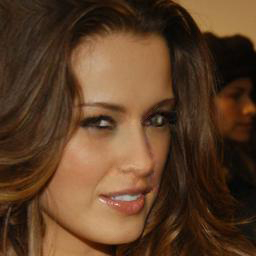}};
    \node[right of=b4, node distance=2.8cm] (b5) {\includegraphics[width=.16\textwidth]{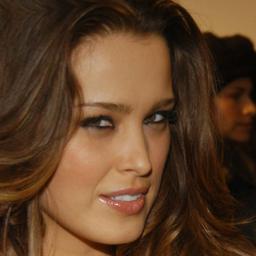}};
    \node[left of=b1, node distance=2.8cm, text width=3cm] (b0) {Blurred eyes and nose, Illumination contrast};
    
    \node[below of=b1, node distance=2.8cm] (c1) {\includegraphics[width=.16\textwidth]{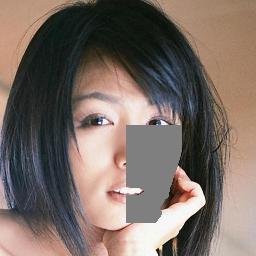}};
    \node[right of=c1, node distance=2.8cm] (c2) {\includegraphics[width=.16\textwidth]{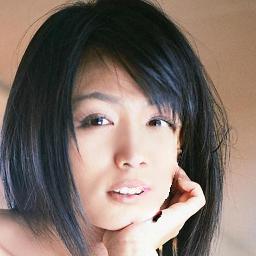}};
    \node[right of=c2, node distance=2.8cm] (c3) {\includegraphics[width=.16\textwidth]{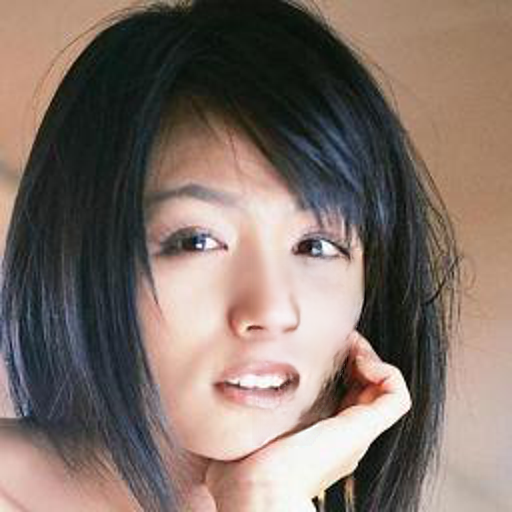}};
    \node[right of=c3, node distance=2.8cm] (c4) {\includegraphics[width=.16\textwidth]{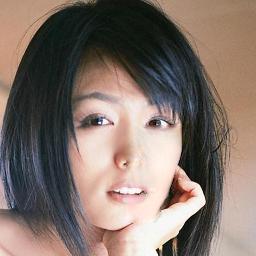}};
    \node[right of=c4, node distance=2.8cm] (c5) {\includegraphics[width=.16\textwidth]{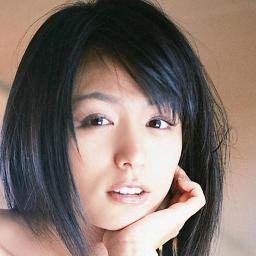}};
    \node[left of=c1, node distance=2.8cm, text width=3cm] (c0) {Blurry cheeks};
    
    \node[below of=c1, node distance=2.8cm] (d1) {\includegraphics[width=.16\textwidth]{figs/celeba/22011/input.jpg}};
    \node[right of=d1, node distance=2.8cm] (d2) {\includegraphics[width=.16\textwidth]{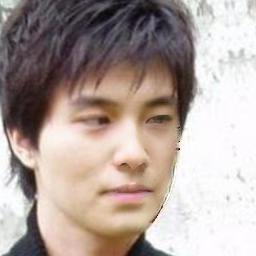}};
    \node[right of=d2, node distance=2.8cm] (d3) {\includegraphics[width=.16\textwidth]{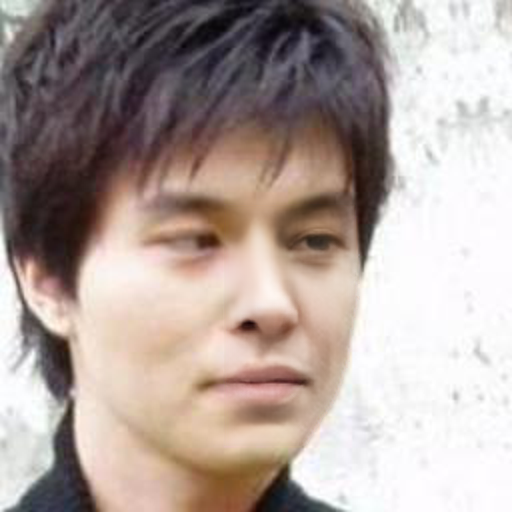}};
    \node[right of=d3, node distance=2.8cm] (d4) {\includegraphics[width=.16\textwidth]{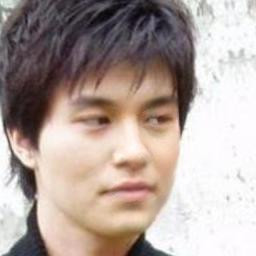}};
    \node[right of=d4, node distance=2.8cm] (d5) {\includegraphics[width=.16\textwidth]{figs/celeba/22011/original.jpg}};
    \node[left of=d1, node distance=2.8cm, text width=3cm] (d0) {Asymmetric eye-gaze};
    
    \node[below of=d1, node distance=2.8cm] (e1) {\includegraphics[width=.16\textwidth]{figs/celebahq/3489/input.jpg}};
    \node[right of=e1, node distance=2.8cm] (e2) {\includegraphics[width=.16\textwidth]{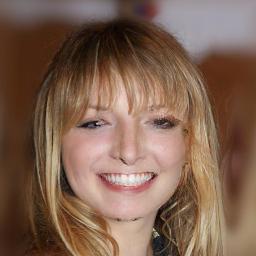}};
    \node[right of=e2, node distance=2.8cm] (e3) {\includegraphics[width=.16\textwidth]{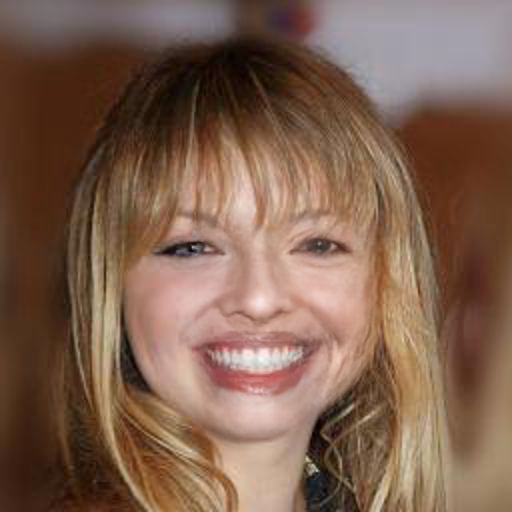}};
    \node[right of=e3, node distance=2.8cm] (e4) {\includegraphics[width=.16\textwidth]{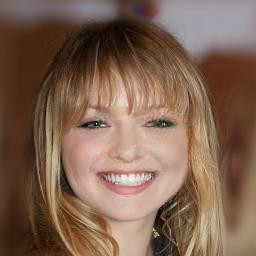}};
    \node[right of=e4, node distance=2.8cm] (e5) {\includegraphics[width=.16\textwidth]{figs/celebahq/3489/original.jpg}};
    \node[left of=e1, node distance=2.8cm, text width=3cm] (e0) {Blurry deformation near mouth and asymmetric eye-gaze};
    
    \node[below of=e1, node distance=1.8cm] (o1) {Input};
    \node[below of=e2, node distance=1.8cm] (o2) {DSA \cite{oracleattention}};
    \node[below of=e3, node distance=1.8cm] (o3) {PConv \cite{partialconv}};
    \node[below of=e4, node distance=1.8cm] (o5) {\ourmethod{}};
    \node[below of=e5, node distance=1.8cm] (o6) {Ground truth};
    \end{tikzpicture}
    \vspace{-1mm}
    \caption{\textbf{Qualitative evaluation} of \ourmethod{} \textit{vs.} PConv \cite{partialconv} and DSA \cite{oracleattention} on a subset of images received from the respective authors. The text on the left mention the specific deformities in the baselines (blurriness, artifacts, asymmetry and other geometric deformations), that is not present in the completions by \ourmethod{}.}
    \label{fig:with_pconv}
\end{figure*} 

\subsection{Comparison against PConv \cite{partialconv} and DSA \cite{oracleattention}}\label{subsec:pconv}
PConv \cite{partialconv} and DSA \cite{oracleattention} have not released publicly available source codes or pre-trained models. Hence, to compare against them, we obtained face completions for a small set of 14 partial images through correspondence with the respective authors\footnote{The images provided by PConv's authors were obtained from a model trained on 512x512 sized images, \versus{} 256x256 for the other baselines including \ourmethod{}.}. We show qualitative results in Fig.~\ref{fig:with_pconv}. One can observe that while PConv \cite{partialconv} and DSA \cite{oracleattention} tend to deform the facial components under certain conditions leading to geometric and photometric artifacts, \ourmethod{} is free of such artifacts and generates more realistic completions. In addition, we provide quantitative metrics on this small set in Table~\ref{tab:pconv}, where \ourmethod{} reports better PSNR, SSIM and LPIPS \cite{lpips} metrics over both the baselines.

\begin{figure*}
    \centering
    \resizebox{0.7\linewidth}{!}{\begin{tabular}{cccccc}
    \hline
    Method & IL & SYM & IR & PSNR $(\uparrow)$ & LPIPS $(\downarrow)$\\
    \hline
    UVGAN & \xmark & \xmark & \xmark & 28.719 & 0.0383\\
    UVGAN-\textit{Sym} & \xmark & \checkmark & \xmark & 28.621 & 0.0392\\
    \ourmethod-\textit{NoIR} & \checkmark & \checkmark & \xmark & 29.959 & 0.0334\\
    \textbf{\ourmethod{}} & \checkmark & \checkmark & \checkmark & \textbf{30.492} & \textbf{0.0326}\\
    \hline
    \end{tabular}}
    
    \begin{tikzpicture}
    \hspace{-2em}
    \node[] (a1) {\hspace{-0.15cm}\includegraphics[width=0.19\linewidth,trim={0 20 0 30},clip]{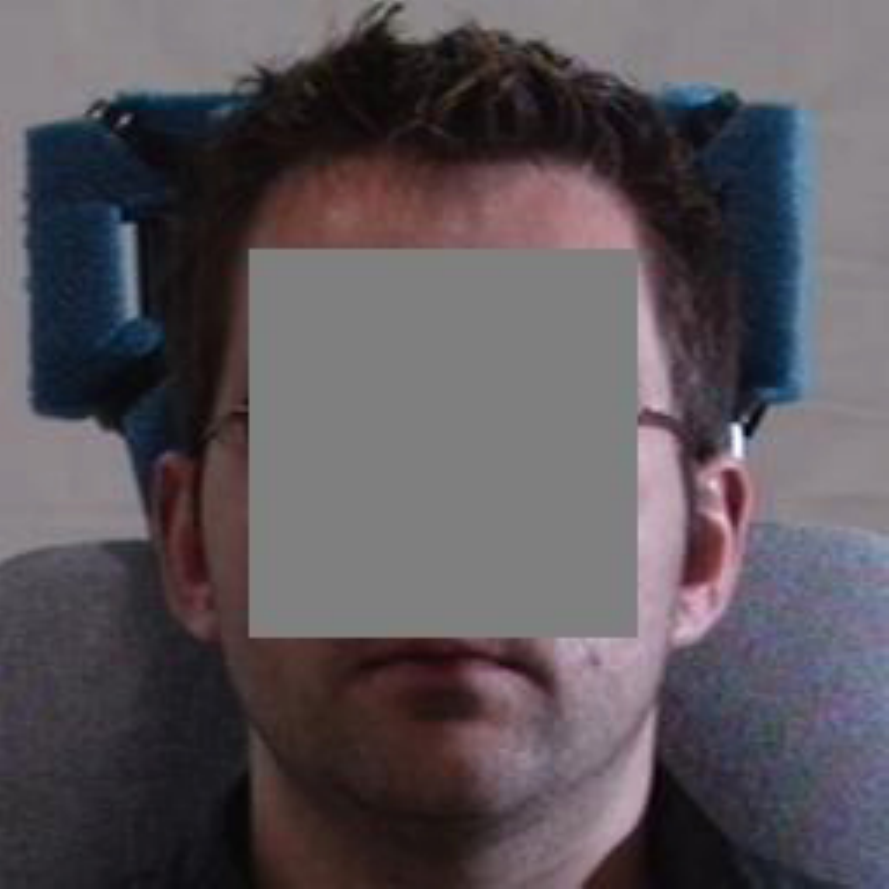}};
    \node[right of=a1, node distance=3.5cm] (a2) {\includegraphics[width=0.19\linewidth,trim={0 20 0 30},clip]{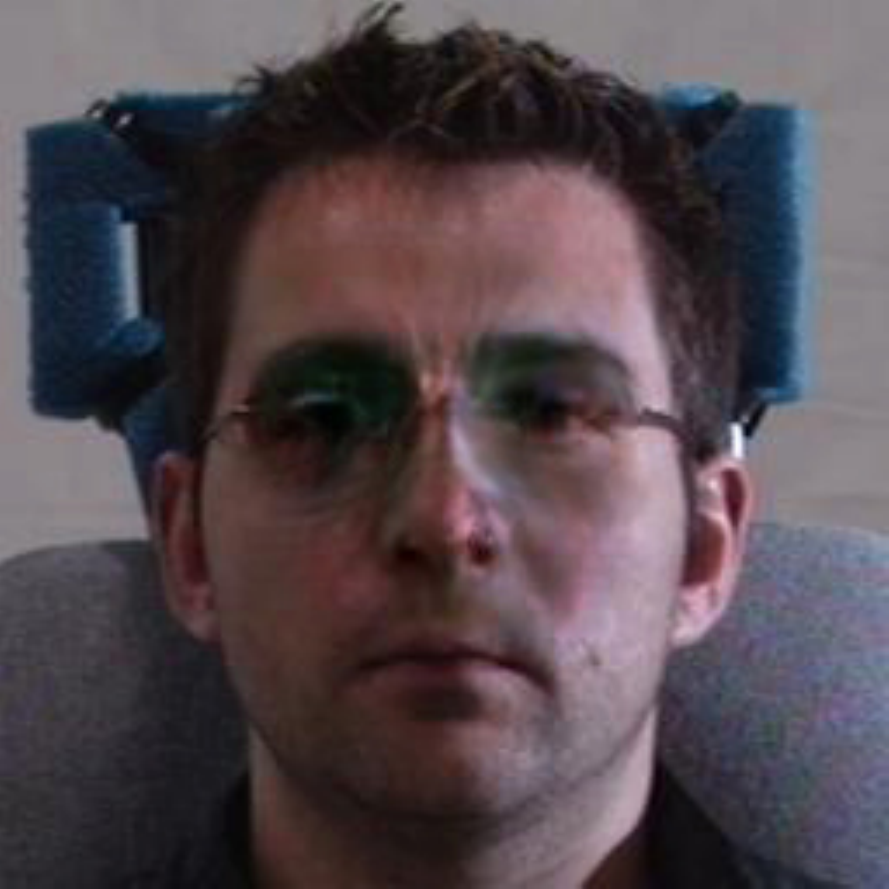}};
    \node[right of=a2, node distance=3.6cm] (a3) {\includegraphics[width=0.19\linewidth,trim={0 20 0 30},clip]{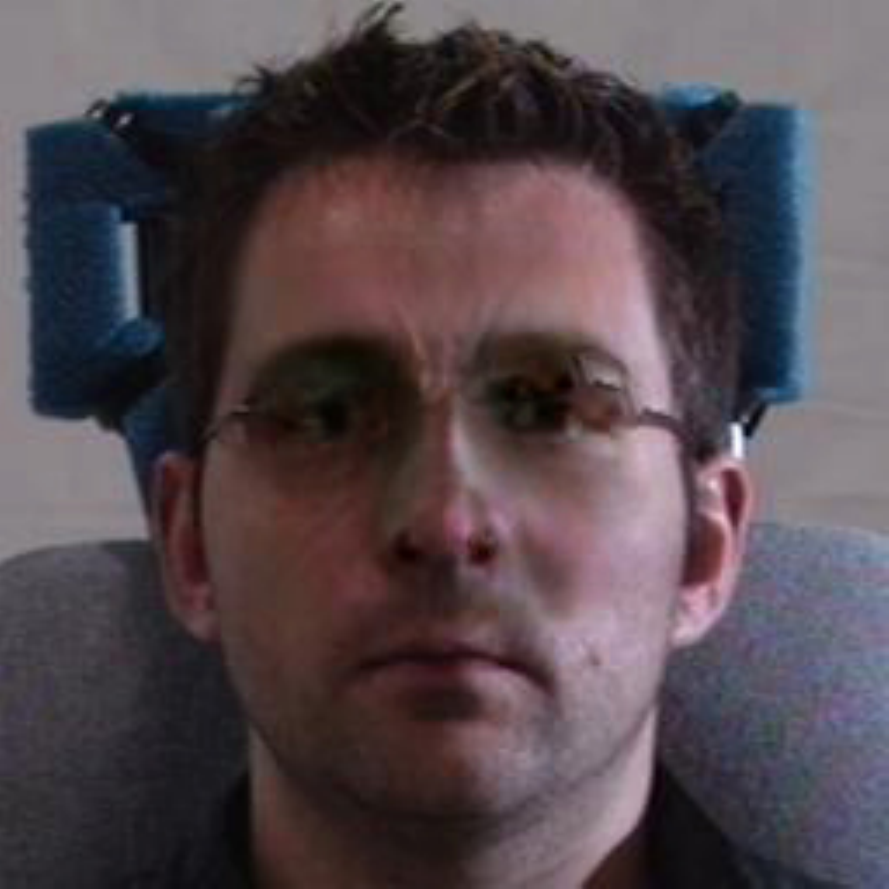}};
    \node[right of=a3, node distance=3.5cm] (a4) {\includegraphics[width=0.19\linewidth,trim={0 20 0 30},clip]{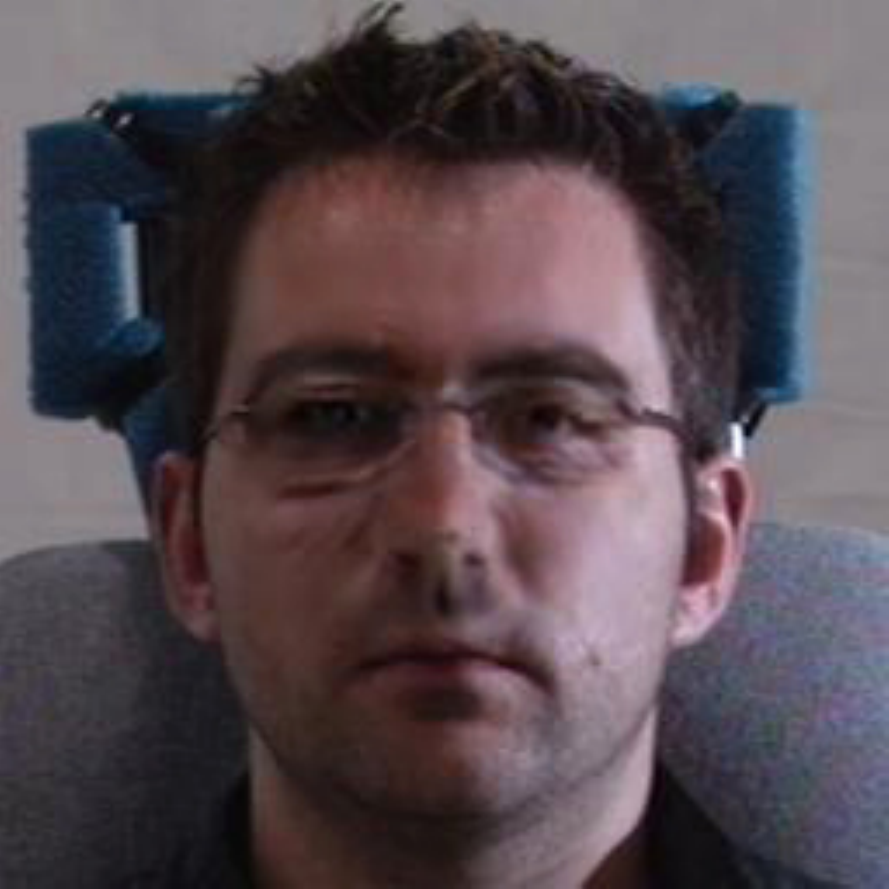}};
    \node[right of=a4, node distance=3.5cm] (a5) {\includegraphics[width=0.19\linewidth,trim={0 20 0 30},clip]{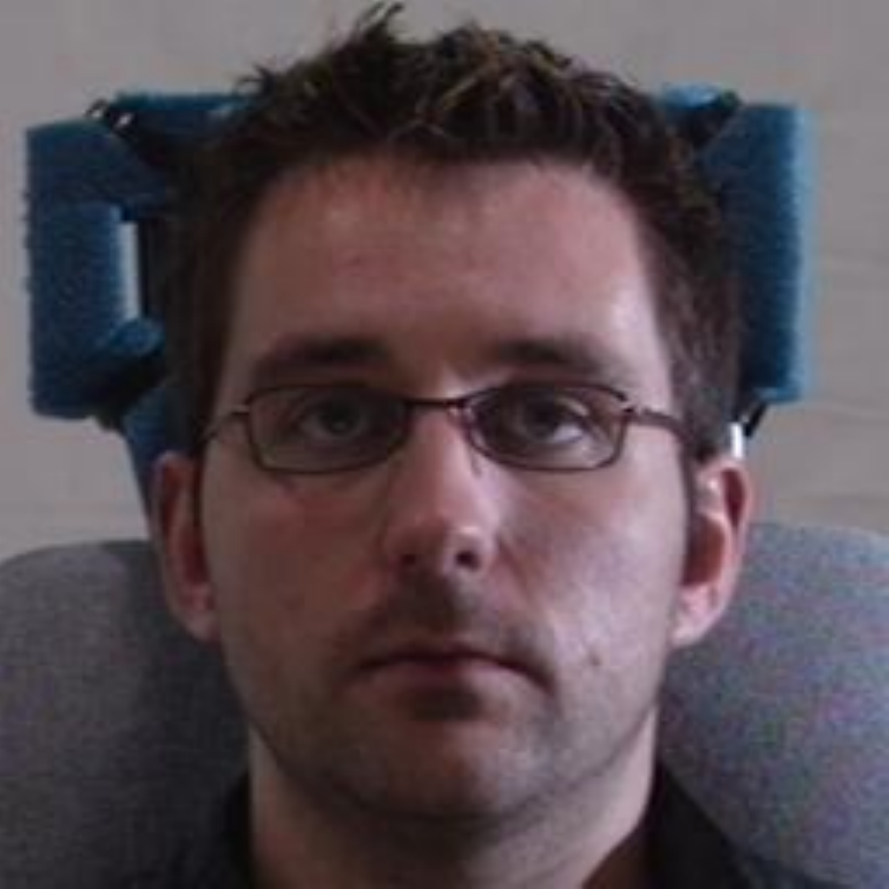}};
    
    \node[below of=a1, node distance=1.7cm] (b1) {Input};
    \node[below of=a2, node distance=1.7cm] (b2) {UVGAN};
    \node[below of=a3, node distance=1.7cm] (b3) {UVGAN-Sym};
    \node[below of=a4, node distance=1.7cm] (b4) {\ourmethod{}};
    \node[below of=a5, node distance=1.7cm] (b5) {Ground truth};
    \end{tikzpicture}
    \caption{Comparing UVGAN \cite{uvgan} reformulated for face completion \versus{} \ourmethod{}.}
    \label{fig:uvgan_reformulated}
\end{figure*}
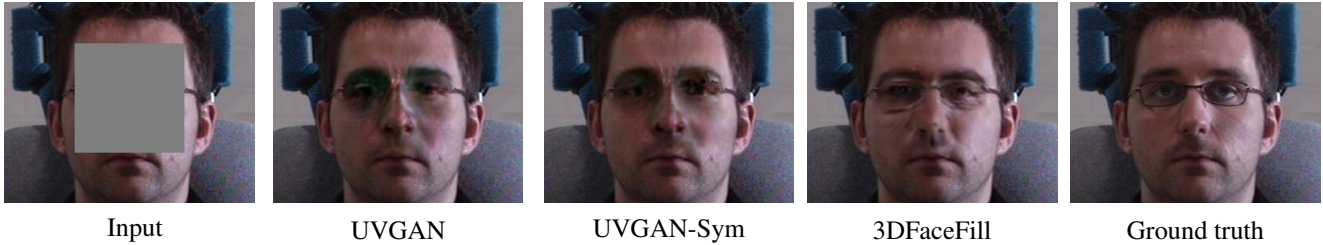

\begin{figure*}
    \centering
    \begin{subfigure}{\textwidth}
        \centering
        \includegraphics[width=0.24\linewidth]{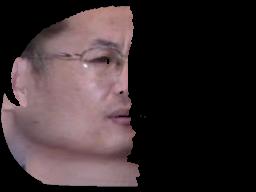}\hspace{8pt}
        \includegraphics[width=0.24\linewidth]{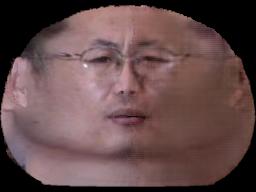}\hspace{8pt}
        \includegraphics[width=0.24\linewidth]{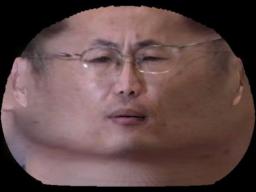}
    \end{subfigure}
    \vspace{8pt}
    \begin{subfigure}{\textwidth}
        \centering
        \includegraphics[width=0.24\linewidth]{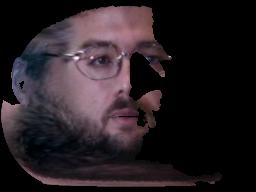}\hspace{8pt}
        \includegraphics[width=0.24\linewidth]{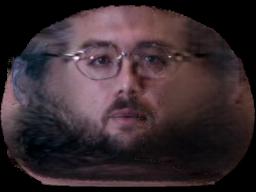}\hspace{8pt}
        \includegraphics[width=0.24\linewidth]{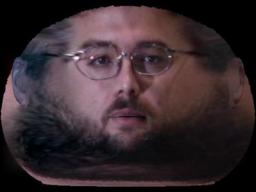}
    \end{subfigure}
    \vspace{8pt}
    \begin{subfigure}{\textwidth}
        \centering
        \includegraphics[width=0.24\linewidth]{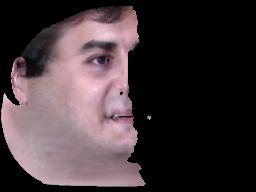}\hspace{8pt}
        \includegraphics[width=0.24\linewidth]{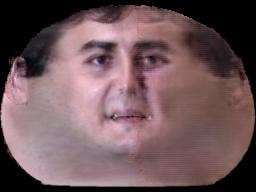}\hspace{8pt}
        \includegraphics[width=0.24\linewidth]{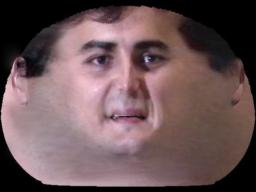}
    \end{subfigure}
    \vspace{8pt}
    \begin{subfigure}{.24\textwidth}
        \includegraphics[width=\textwidth]{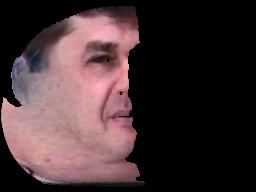}
        \caption{Input}
    \end{subfigure}\hspace{8pt}
    \begin{subfigure}{.24\textwidth}
        \includegraphics[width=\linewidth]{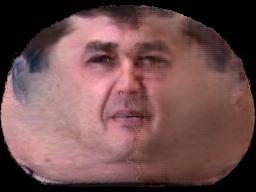}
        \caption{\ourmethod{}}
    \end{subfigure}\hspace{8pt}
    \begin{subfigure}{.24\textwidth}
        \includegraphics[width=\linewidth]{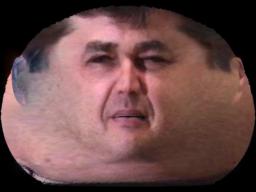}
        \caption{Groundtruth}
    \end{subfigure}
    \caption{Qualitative evaluation of texture completion by the proposed Sym-UNet on the UVDB-MPIE dataset \cite{uvgan}.\vspace{10pt}}
    \label{fig:uvdb}
\end{figure*}

\vspace{5pt}
\subsection{Comparison against UVGAN \cite{uvgan}}\label{subsec:uvgan}
The proposed face completion method, \ourmethod{}, has three parts, (i) disentangling 2D image into factors such as 3D pose, 3D shape, albedo and illumination (\textit{IL}), (ii) enforcing symmetry in UV albedo (\textit{SYM}), and (iii) iterative refinement of face completion through progressively more accurate 3D pose and shape estimation (\textit{IR}). UVGAN [7] on the other hand, (i) performs completion of the missing texture in the UV-representation due to self-occlusion instead of completing a partial face image itself, (ii) unlike \ourmethod{}, does not disentangle texture further into albedo and illumination, (iii) does not impose symmetry prior on the UV texture, and (iv) uses 3DMM on a fully visible face image rather than a partial image to obtain texture. Since no source code or pretrained model of UVGAN is available, we evaluate these differences in two ways: (A)  by reformulating UVGAN for face completion, and (B) comparing UVGAN with our Sym-UNet model on their publicly released texture dataset. We now present the two evaluations.

\vspace{5pt}
\noindent\textbf{A: Comparison with UVGAN \cite{uvgan} Reformulated for Face Completion}

To simulate UVGAN \cite{uvgan} for face completion, we remove the illumination disentanglement (\textit{IL}), symmetry loss (\textit{SYM}) and iterative refinement (\textit{IR}) from \ourmethod{} (refer to Fig.~\ref{fig:uvgan_reformulated}). We call the variant with \textit{SYM} as UVGAN-\textit{Sym}, and the variant with both \textit{IL} and \textit{SYM} as \ourmethod{}-\textit{NoIR}. Adding \textit{IR} makes for our full model \ourmethod{}. We compare the above-mentioned variants for face completion on the CelebA \cite{celeba} dataset and report the quantitative and qualitative results in Fig.~\ref{fig:uvgan_reformulated}. One can observe that \ourmethod{} \emph{significantly} outperforms UVGAN as well as the other variants both quantitatively as well as qualitatively. Further, we can see that introducing the symmetry loss (\textit{SYM}) in UVGAN-\textit{Sym} hurts performance since, unlike UV-albedo, UV-texture is not inherently symmetric in faces because of the entangled illumination. Completion on the disentangled albedo (\textit{IL}) instead improves performance in \ourmethod-\textit{NoIR}. Lastly, iterative refinement (\textit{IR}) further improves completion on top of \textit{IL} and \textit{SYM}. This demonstrates the effectiveness of the novelties that \ourmethod{} introduces over UVGAN \cite{uvgan}.

\vspace{5pt}
\noindent\textbf{B: Sym-UNet \versus{} UVGAN on Texture Completion}

In this evaluation, we trained our Sym-UNet model on the UVDB-MPIE texture dataset released by the authors of UVGAN \cite{uvgan}. We split the dataset into a 80:20 train-test split and resized the texture maps to $192 \times 256$ for training. Similar to UVGAN, we do not include the symmetry loss because of the presence of illumination variations and the availability of synthetically completed texture maps, which reduces the utility of symmetry-loss. The rest of the Sym-UNet is retained as such. On the test set, we report a PSNR of 30.1 (\versus{} UVGAN's 25.8) and SSIM of 0.937 (\versus{} UVGAN's 0.886). Further, we show qualitative results in Fig.~\ref{fig:uvdb}, where we see that our completed textures resemble the ground truth closely (we do not have the corresponding completions by UVGAN). Thus, our proposed Sym-UNet network is comparatively better suited for UV-completion than the network used in UVGAN \cite{uvgan}. 

\subsection{Further Qualitative Evaluation}\label{subsec:qualitative}

\vspace{5pt}
\noindent \textbf{A: Diverse Conditions}

\begin{figure*}
    \centering
    \tikzstyle{block} = [rectangle, draw, fill=blue!20, text centered]
    \begin{tikzpicture}
    \node[text width=5cm] (a0) {DARKER COMPLEXION};
    \node[below of=a0, node distance=1.8cm] (a1) {\includegraphics[width=.18\textwidth,trim={20 20 20 20},clip]{figs/internet/29/input.jpg}};
    \node[right of=a1, node distance=3.4cm] (a2) {\includegraphics[width=.18\textwidth,trim={20 20 20 20},clip]{figs/internet/29/gca.jpg}};
    \node[right of=a2, node distance=3.4cm] (a3) {\includegraphics[width=.18\textwidth,trim={20 20 20 20},clip]{figs/internet/29/pluralinpaint.jpg}};
    \node[right of=a3, node distance=3.4cm] (a4) {\includegraphics[width=.18\textwidth,trim={20 20 20 20},clip]{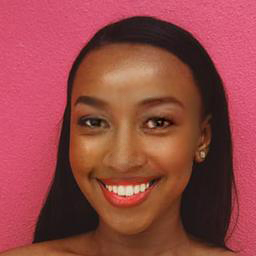}};
    \node[right of=a4, node distance=3.4cm] (a5) {\includegraphics[width=.18\textwidth,trim={20 20 20 20},clip]{figs/internet/29/original.jpg}};
    
    \node[below of=a1, node distance=3.4cm] (b1) {\includegraphics[width=.18\textwidth,trim={20 20 20 20},clip]{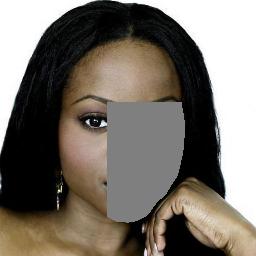}};
    \node[right of=b1, node distance=3.4cm] (b2) {\includegraphics[width=.18\textwidth,trim={20 20 20 20},clip]{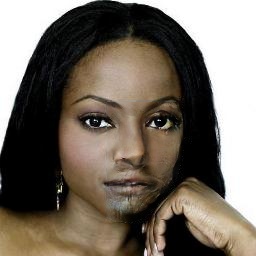}};
    \node[right of=b2, node distance=3.4cm] (b3) {\includegraphics[width=.18\textwidth,trim={20 20 20 20},clip]{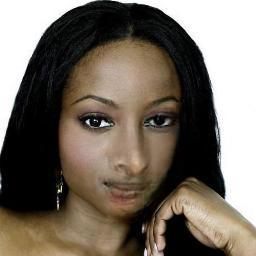}};
    \node[right of=b3, node distance=3.4cm] (b4) {\includegraphics[width=.18\textwidth,trim={20 20 20 20},clip]{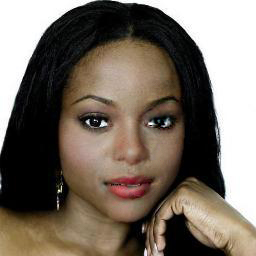}};
    \node[right of=b4, node distance=3.4cm] (b5) {\includegraphics[width=.18\textwidth,trim={20 20 20 20},clip]{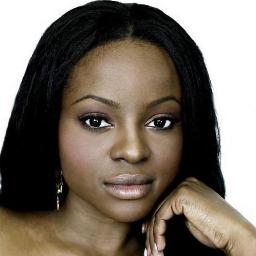}};
    
    \node[below of=b1, node distance=1.9cm, text width=5cm] (c0) {LARGE POSES};
    \node[below of=c0, node distance=1.8cm] (c1) {\includegraphics[width=.18\textwidth,trim={20 20 20 20},clip]{figs/celeba/48/input.jpg}};
    \node[right of=c1, node distance=3.4cm] (c2) {\includegraphics[width=.18\textwidth,trim={20 20 20 20},clip]{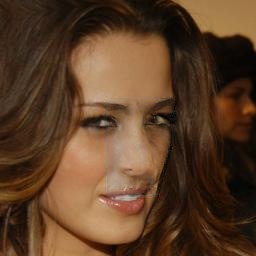}};
    \node[right of=c2, node distance=3.4cm] (c3) {\includegraphics[width=.18\textwidth,trim={20 20 20 20},clip]{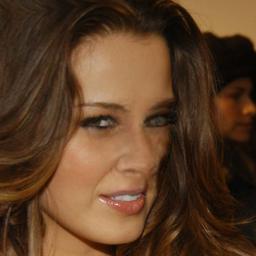}};
    \node[right of=c3, node distance=3.4cm] (c4) {\includegraphics[width=.18\textwidth,trim={20 20 20 20},clip]{figs/celeba/48/ours_nearup.jpg}};
    \node[right of=c4, node distance=3.4cm] (c5) {\includegraphics[width=.18\textwidth,trim={20 20 20 20},clip]{figs/celeba/48/original.jpg}};
    
    \node[below of=c1, node distance=3.4cm] (d1) {\includegraphics[width=.18\textwidth,trim={20 20 20 20},clip]{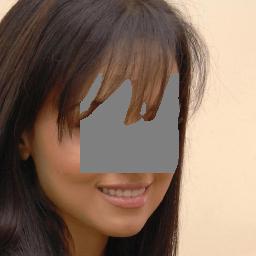}};
    \node[right of=d1, node distance=3.4cm] (d2) {\includegraphics[width=.18\textwidth,trim={20 20 20 20},clip]{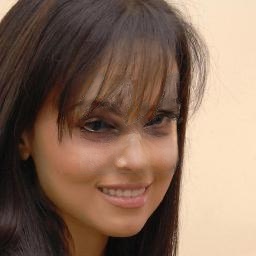}};
    \node[right of=d2, node distance=3.4cm] (d3) {\includegraphics[width=.18\textwidth,trim={20 20 20 20},clip]{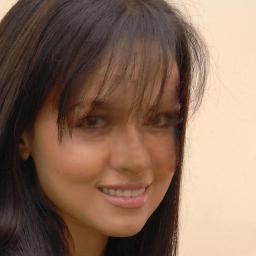}};
    \node[right of=d3, node distance=3.4cm] (d4) {\includegraphics[width=.18\textwidth,trim={20 20 20 20},clip]{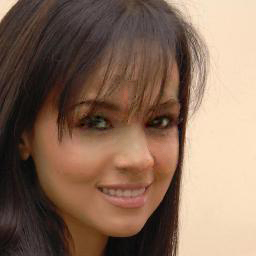}};
    \node[right of=d4, node distance=3.4cm] (d5) {\includegraphics[width=.18\textwidth,trim={20 20 20 20},clip]{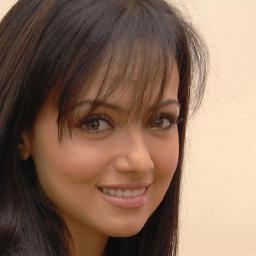}};
    
    \node[below of=d1, node distance=1.9cm, text width=5cm] (e0) {ILLUMINATION CONTRAST};
    \node[below of=e0, node distance=1.8cm] (e1) {\includegraphics[width=.18\textwidth,trim={20 20 20 20},clip]{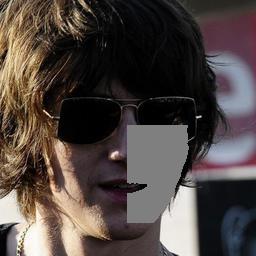}};
    \node[right of=e1, node distance=3.4cm] (e2) {\includegraphics[width=.18\textwidth,trim={20 20 20 20},clip]{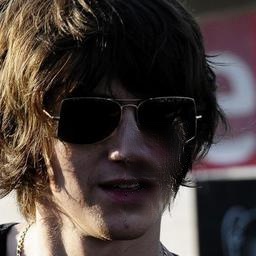}};
    \node[right of=e2, node distance=3.4cm] (e3) {\includegraphics[width=.18\textwidth,trim={20 20 20 20},clip]{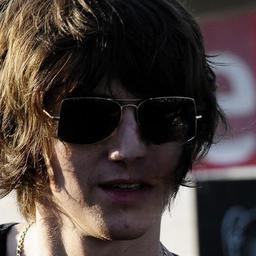}};
    \node[right of=e3, node distance=3.4cm] (e4) {\includegraphics[width=.18\textwidth,trim={20 20 20 20},clip]{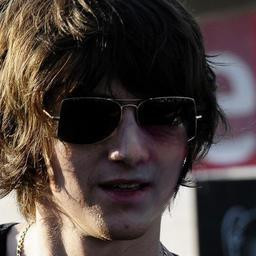}};
    \node[right of=e4, node distance=3.4cm] (e5) {\includegraphics[width=.18\textwidth,trim={20 20 20 20},clip]{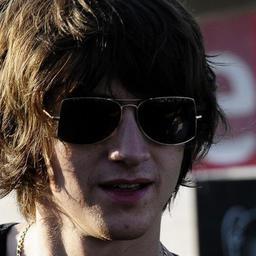}};
    
    \node[below of=e1, node distance=3.4cm] (f1) {\includegraphics[width=.18\textwidth,trim={20 20 20 20},clip]{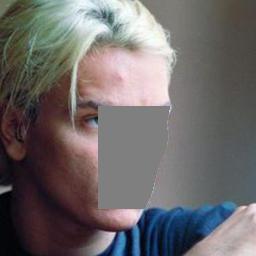}};
    \node[right of=f1, node distance=3.4cm] (f2) {\includegraphics[width=.18\textwidth,trim={20 20 20 20},clip]{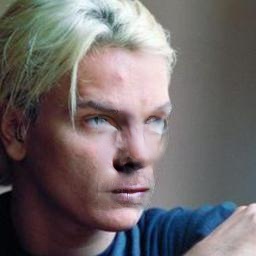}};
    \node[right of=f2, node distance=3.4cm] (f3) {\includegraphics[width=.18\textwidth,trim={20 20 20 20},clip]{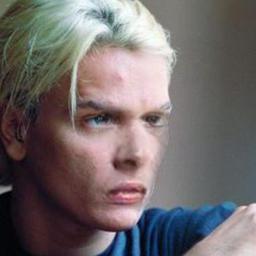}};
    \node[right of=f3, node distance=3.4cm] (f4) {\includegraphics[width=.18\textwidth,trim={20 20 20 20},clip]{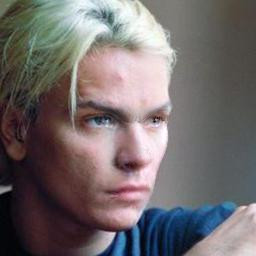}};
    \node[right of=f4, node distance=3.4cm] (f5) {\includegraphics[width=.18\textwidth,trim={20 20 20 20},clip]{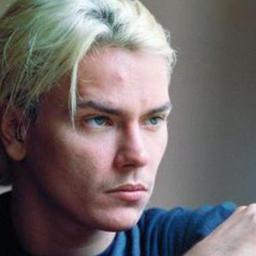}};
    
    \node[below of=f1, node distance=2.0cm] (o1) {Input};
    \node[below of=f2, node distance=2.0cm] (o5) {DeepFillv2 \cite{freeforminpainting}};
    \node[below of=f3, node distance=2.0cm] (o6) {PIC \cite{zheng2019pluralistic}};
    \node[below of=f4, node distance=2.0cm] (o8) {\ourmethod{} (Ours)};
    \node[below of=f5, node distance=2.0cm] (o9) {Ground Truth};
    \end{tikzpicture}
    \caption{\vspace{-1mm}Qualitative evaluation under diverse conditions (complexion, pose, illumination).}
    \label{fig:qual1}
\end{figure*}

\begin{figure*}
    \centering
    \tikzstyle{block} = [rectangle, draw, fill=blue!20, text centered]
    \begin{tikzpicture}
    \node[text width=5cm] (a0) {ASYMMETRY IN EYE-GAZE};
    \node[below of=a0, node distance=2.0cm] (a1) {\includegraphics[width=.18\textwidth,trim={20 20 20 20},clip]{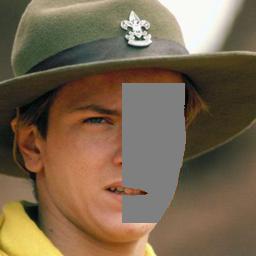}};
    \node[right of=a1, node distance=3.4cm] (a2) {\includegraphics[width=.18\textwidth,trim={20 20 20 20},clip]{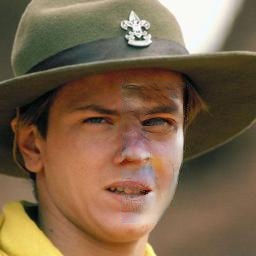}};
    \node[right of=a2, node distance=3.4cm] (a3) {\includegraphics[width=.18\textwidth,trim={20 20 20 20},clip]{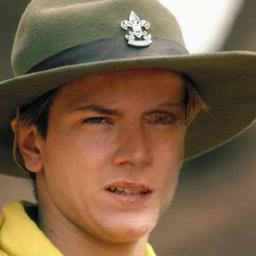}};
    \node[right of=a3, node distance=3.4cm] (a4) {\includegraphics[width=.18\textwidth,trim={20 20 20 20},clip]{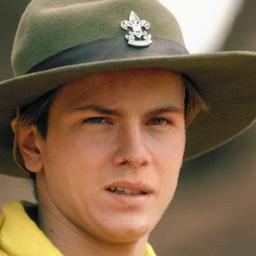}};
    \node[right of=a4, node distance=3.4cm] (a5) {\includegraphics[width=.18\textwidth,trim={20 20 20 20},clip]{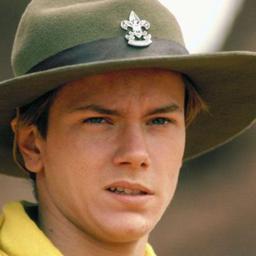}};
    
    \node[below of=a1, node distance=3.4cm] (b1) {\includegraphics[width=.18\textwidth,trim={20 20 20 20},clip]{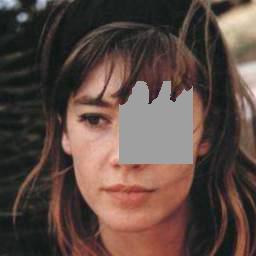}};
    \node[right of=b1, node distance=3.4cm] (b2) {\includegraphics[width=.18\textwidth,trim={20 20 20 20},clip]{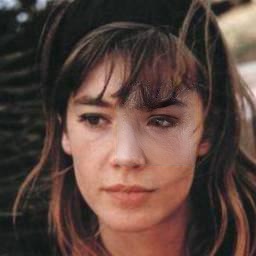}};
    \node[right of=b2, node distance=3.4cm] (b3) {\includegraphics[width=.18\textwidth,trim={20 20 20 20},clip]{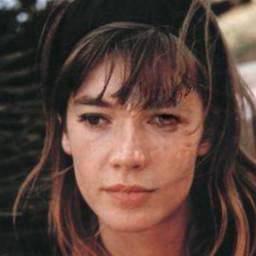}};
    \node[right of=b3, node distance=3.4cm] (b4) {\includegraphics[width=.18\textwidth,trim={20 20 20 20},clip]{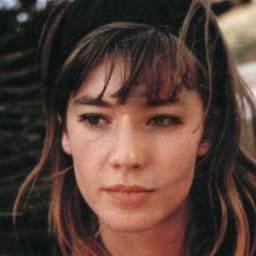}};
    \node[right of=b4, node distance=3.4cm] (b5) {\includegraphics[width=.18\textwidth,trim={20 20 20 20},clip]{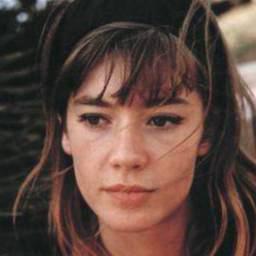}};
    
    \node[below of=b1, node distance=2cm, text width=5cm] (c0) {SHAPE DEFORMATIONS};
    \node[below of=c0, node distance=2.0cm] (c1) {\includegraphics[width=.18\textwidth,trim={20 20 20 20},clip]{figs/celebahq/531/input.jpg}};
    \node[right of=c1, node distance=3.4cm] (c2) {\includegraphics[width=.18\textwidth,trim={20 20 20 20},clip]{figs/celebahq/531/gca.jpg}};
    \node[right of=c2, node distance=3.4cm] (c3) {\includegraphics[width=.18\textwidth,trim={20 20 20 20},clip]{figs/celebahq/531/pluralinpaint.jpg}};
    \node[right of=c3, node distance=3.4cm] (c4) {\includegraphics[width=.18\textwidth,trim={20 20 20 20},clip]{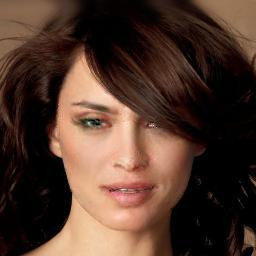}};
    \node[right of=c4, node distance=3.4cm] (c5) {\includegraphics[width=.18\textwidth,trim={20 20 20 20},clip]{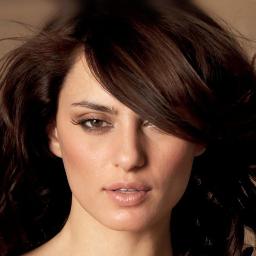}};
    
    \node[below of=c1, node distance=3.4cm] (d1) {\includegraphics[width=.18\textwidth,trim={20 20 20 20},clip]{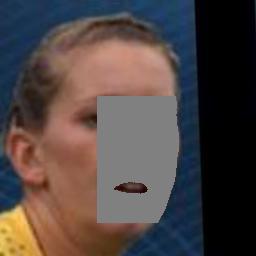}};
    \node[right of=d1, node distance=3.4cm] (d2) {\includegraphics[width=.18\textwidth,trim={20 20 20 20},clip]{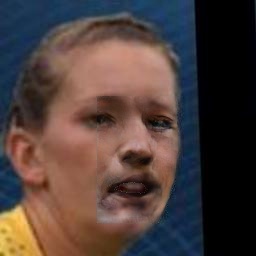}};
    \node[right of=d2, node distance=3.4cm] (d3) {\includegraphics[width=.18\textwidth,trim={20 20 20 20},clip]{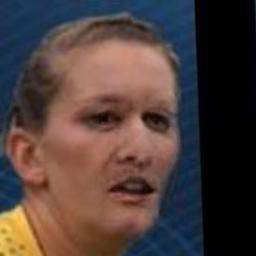}};
    \node[right of=d3, node distance=3.4cm] (d4) {\includegraphics[width=.18\textwidth,trim={20 20 20 20},clip]{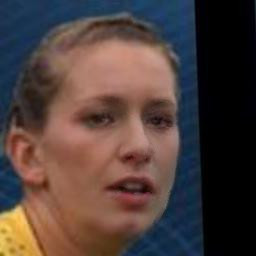}};
    \node[right of=d4, node distance=3.4cm] (d5) {\includegraphics[width=.18\textwidth,trim={20 20 20 20},clip]{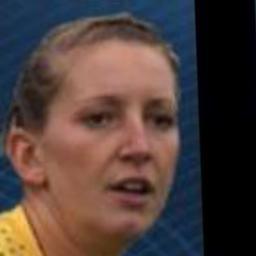}};
    
    \node[below of=d1, node distance=3.4cm] (e1) {\includegraphics[width=.18\textwidth,trim={20 20 20 20},clip]{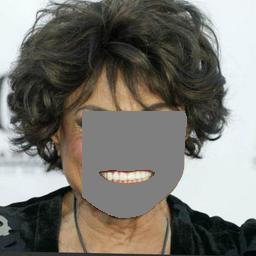}};
    \node[right of=e1, node distance=3.4cm] (e2) {\includegraphics[width=.18\textwidth,trim={20 20 20 20},clip]{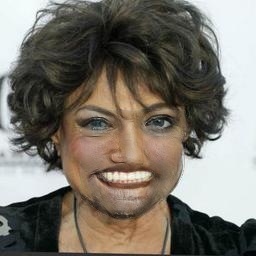}};
    \node[right of=e2, node distance=3.4cm] (e3) {\includegraphics[width=.18\textwidth,trim={20 20 20 20},clip]{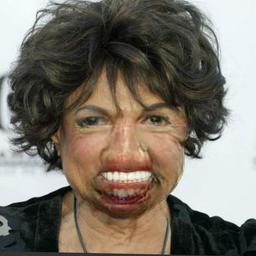}};
    \node[right of=e3, node distance=3.4cm] (e4) {\includegraphics[width=.18\textwidth,trim={20 20 20 20},clip]{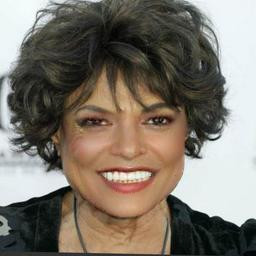}};
    \node[right of=e4, node distance=3.4cm] (e5) {\includegraphics[width=.18\textwidth,trim={20 20 20 20},clip]{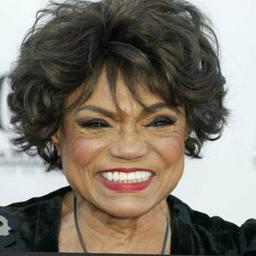}};
    
    \node[below of=e1, node distance=3.4cm] (f1) {\includegraphics[width=.18\textwidth,trim={20 20 20 20},clip]{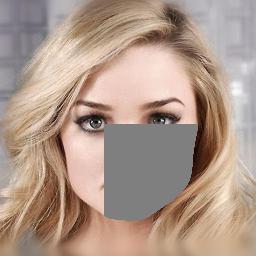}};
    \node[right of=f1, node distance=3.4cm] (f2) {\includegraphics[width=.18\textwidth,trim={20 20 20 20},clip]{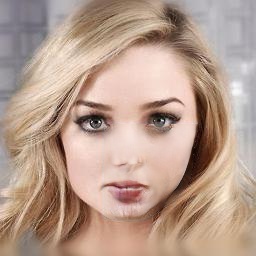}};
    \node[right of=f2, node distance=3.4cm] (f3) {\includegraphics[width=.18\textwidth,trim={20 20 20 20},clip]{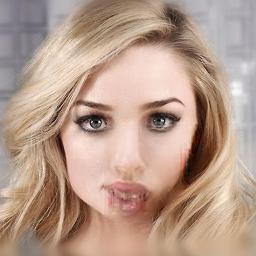}};
    \node[right of=f3, node distance=3.4cm] (f4) {\includegraphics[width=.18\textwidth,trim={20 20 20 20},clip]{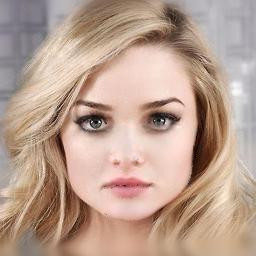}};
    \node[right of=f4, node distance=3.4cm] (f5) {\includegraphics[width=.18\textwidth,trim={20 20 20 20},clip]{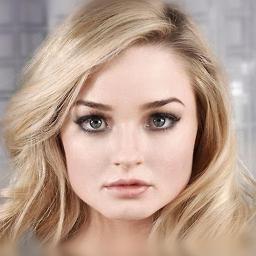}};
    
    \node[below of=f1, node distance=2.0cm] (o1) {Input};
    \node[below of=f2, node distance=2.0cm] (o5) {DeepFillv2 \cite{freeforminpainting}};
    \node[below of=f3, node distance=2.0cm] (o6) {PIC \cite{zheng2019pluralistic}};
    \node[below of=f4, node distance=2.0cm] (o8) {\ourmethod{} (Ours)};
    \node[below of=f5, node distance=2.0cm] (o9) {Ground Truth};
    \end{tikzpicture}
    \caption{\vspace{-1mm}Qualitative evaluation under diverse conditions (eye-gaze, shape).}
    \label{fig:qual2}
\end{figure*}

We present further qualitative comparison of face completion by the proposed \ourmethod{} \versus{} DeepFillv2 \cite{freeforminpainting} and PIC \cite{zheng2019pluralistic} under diverse conditions. Fig.~\ref{fig:qual1} show the qualitative comparison on faces with dark complexion, challenging poses and illumination contrast. Fig.~\ref{fig:qual2} shows examples where baselines tend to introduce asymmetry in eye-gaze or deform various face components, such as, nose, mouth, \etc. In all these cases, \ourmethod{} generates more realistic completions while preserving the visible illumination contrast and bilateral symmetry, because of the disentangled completion of albedo and explicit enforcement of 3D shape, pose, illumination and symmetric priors.

\begin{table*}[t]
    \centering
    \newcolumntype{M}[1]{>{\raggedright\arraybackslash}m{#1}}
    \begin{tabular}{M{25mm}l@{\hspace{1\tabcolsep}}c@{\hspace{1\tabcolsep}}c@{\hspace{1\tabcolsep}}c@{\hspace{1\tabcolsep}}c@{\hspace{1\tabcolsep}}c}
        \hline
        \textbf{Dataset} & \textbf{Metric} & GFC \cite{genfacecompletion} & SymmFC \cite{li2020symmetry} & DeepFillv2 \cite{freeforminpainting} & PIC \cite{zheng2019pluralistic} & \ourmethod{}\\
        \hline
        \multirow{3}{*}{\textbf{MultiPIE:Pose}} & \textbf{PSNR} $(\uparrow)$ & 24.7557 & 24.7177 & 26.3385 & 26.4301 & \textbf{27.8226}\\
        & \textbf{SSIM} $(\uparrow)$ & 0.9187 & 0.9289 & 0.9383 & 0.9451 & \textbf{0.9482}\\
        & \textbf{LPIPS} $(\downarrow)$ & 0.0822 & 0.0692 & 0.0527 & 0.0471 & \textbf{0.0409}\\
        \hline
        \multirow{3}{*}{\textbf{MultiPIE:Illu}} & \textbf{PSNR} $(\uparrow)$ & 23.5749 & 24.4813 & 26.4981 & 26.2938 & \textbf{27.8865}\\
        & \textbf{SSIM} $(\uparrow)$ & 0.8676 & 0.8618 & 00.8718 & 0.8825 & \textbf{0.8935}\\
        & \textbf{LPIPS} $(\downarrow)$ & 0.1232 & 0.0747 & 0.0640 & 0.0540 & \textbf{0.0484}\\
        \hline
        \multirow{3}{*}{\textbf{Internet}} & \textbf{PSNR} $(\uparrow)$ & 24.1775 & 24.2829 & 26.4957 & 25.6326 & \textbf{28.8463}\\
        & \textbf{SSIM} $(\uparrow)$ & 0.9042 & 0.9168 & 0.9293 & 0.9317 & \textbf{0.9526}\\
        & \textbf{LPIPS} $(\downarrow)$ & 0.0913 & 0.0625 & 0.0493 & 0.0466 & \textbf{0.0390}\\
        \hline
    \end{tabular}
    \caption{Further quantitative evaluation of \ourmethod{} \textit{vs.} the baselines on the pose-varying (MultiPIE:Pose) and illumination varying (MultiPIE:Illu) subsets of the  MultiPIE dataset \cite{multipie} and in-the-wild images downloaded from the Internet.\vspace{20pt}}
    \label{tab:quantitative}
\end{table*}

\begin{figure*}
    \scriptsize
    \centering
    \tikzstyle{block} = [rectangle, draw, fill=blue!20, text centered]
    \begin{tikzpicture}
        \node[rotate=90, text centered] (a00) {Input};
        \node[right of=a00, node distance=1.35cm] (a01) {\includegraphics[width=.12\textwidth,trim={30 30 30 30},clip]{figs/multipie_pose/012/010/input.jpg}};
        \node[right of=a01, node distance=2.10cm] (a02) {\includegraphics[width=.12\textwidth,trim={30 30 30 30},clip]{figs/multipie_pose/012/200/input.jpg}};
        \node[right of=a02, node distance=2.10cm] (a03) {\includegraphics[width=.12\textwidth,trim={30 30 30 30},clip]{figs/multipie_pose/012/190/input.jpg}};
        \node[right of=a03, node distance=2.10cm] (a04) {\includegraphics[width=.12\textwidth,trim={30 30 30 30},clip]{figs/multipie_pose/012/041/input.jpg}};
        \node[right of=a04, node distance=2.10cm] (a05) {\includegraphics[width=.12\textwidth,trim={30 30 30 30},clip]{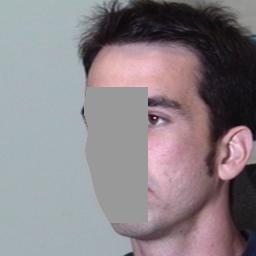}};
        \node[right of=a05, node distance=2.10cm] (a06) {\includegraphics[width=.12\textwidth,trim={30 30 30 30},clip]{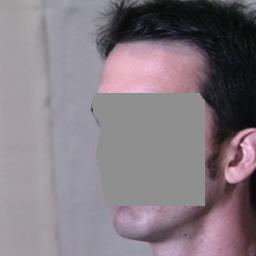}};
        \node[right of=a06, node distance=2.10cm] (a07) {\includegraphics[width=.12\textwidth,trim={30 30 30 30},clip]{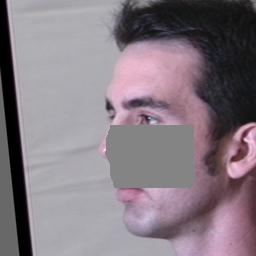}};
        \node[right of=a07, node distance=2.10cm] (a08) {\includegraphics[width=.12\textwidth,trim={30 30 30 30},clip]{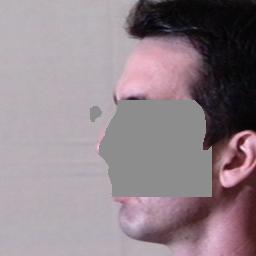}};

        \node[below of=a01, node distance=2.10cm] (b01) {\includegraphics[width=.12\textwidth,trim={30 30 30 30},clip]{figs/multipie_pose/012/010/gca.jpg}};
        \node[right of=b01, node distance=2.10cm] (b02) {\includegraphics[width=.12\textwidth,trim={30 30 30 30},clip]{figs/multipie_pose/012/200/gca.jpg}};
        \node[right of=b02, node distance=2.10cm] (b03) {\includegraphics[width=.12\textwidth,trim={30 30 30 30},clip]{figs/multipie_pose/012/190/gca.jpg}};
        \node[right of=b03, node distance=2.10cm] (b04) {\includegraphics[width=.12\textwidth,trim={30 30 30 30},clip]{figs/multipie_pose/012/041/gca.jpg}};
        \node[right of=b04, node distance=2.10cm] (b05) {\includegraphics[width=.12\textwidth,trim={30 30 30 30},clip]{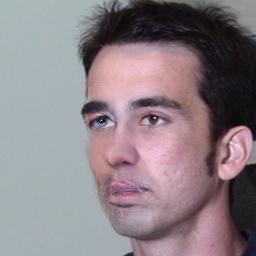}};
        \node[right of=b05, node distance=2.10cm] (b06) {\includegraphics[width=.12\textwidth,trim={30 30 30 30},clip]{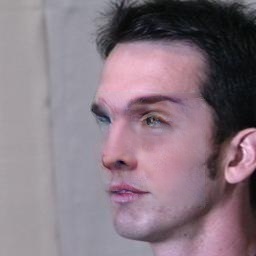}};
        \node[right of=b06, node distance=2.10cm] (b07) {\includegraphics[width=.12\textwidth,trim={30 30 30 30},clip]{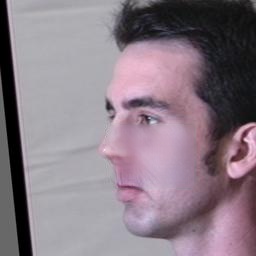}};
        \node[right of=b07, node distance=2.10cm] (b08) {\includegraphics[width=.12\textwidth,trim={30 30 30 30},clip]{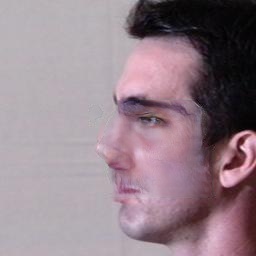}};
        \node[left of=b01, node distance=1.35cm, rotate=90, text centered] (b00) {DeepFillv2 \cite{freeforminpainting}};
        
        \node[below of=b01, node distance=2.10cm] (c01) {\includegraphics[width=.12\textwidth,trim={30 30 30 30},clip]{figs/multipie_pose/012/010/pluralinpaint.jpg}};
        \node[right of=c01, node distance=2.10cm] (c02) {\includegraphics[width=.12\textwidth,trim={30 30 30 30},clip]{figs/multipie_pose/012/200/pluralinpaint.jpg}};
        \node[right of=c02, node distance=2.10cm] (c03) {\includegraphics[width=.12\textwidth,trim={30 30 30 30},clip]{figs/multipie_pose/012/190/pluralinpaint.jpg}};
        \node[right of=c03, node distance=2.10cm] (c04) {\includegraphics[width=.12\textwidth,trim={30 30 30 30},clip]{figs/multipie_pose/012/041/pluralinpaint.jpg}};
        \node[right of=c04, node distance=2.10cm] (c05) {\includegraphics[width=.12\textwidth,trim={30 30 30 30},clip]{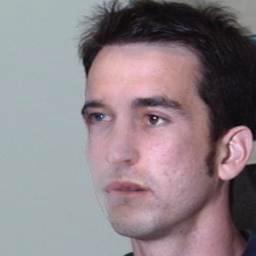}};
        \node[right of=c05, node distance=2.10cm] (c06) {\includegraphics[width=.12\textwidth,trim={30 30 30 30},clip]{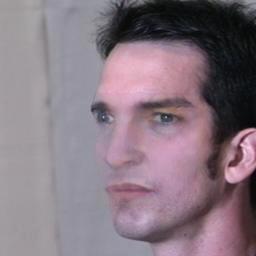}};
        \node[right of=c06, node distance=2.10cm] (c07) {\includegraphics[width=.12\textwidth,trim={30 30 30 30},clip]{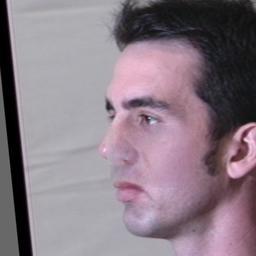}};
        \node[right of=c07, node distance=2.10cm] (c08) {\includegraphics[width=.12\textwidth,trim={30 30 30 30},clip]{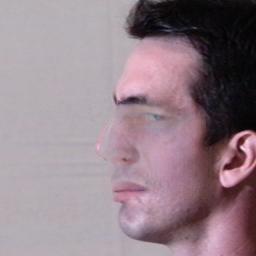}};
        \node[left of=c01, node distance=1.35cm, rotate=90, text centered] (c00) {PICNet \cite{zheng2019pluralistic}};
        
        \node[below of=c01, node distance=2.10cm] (d01) {\includegraphics[width=.12\textwidth,trim={30 30 30 30},clip]{figs/multipie_pose/012/010/output_poisson.jpg}};
        \node[right of=d01, node distance=2.10cm] (d02) {\includegraphics[width=.12\textwidth,trim={30 30 30 30},clip]{figs/multipie_pose/012/200/output_poisson.jpg}};
        \node[right of=d02, node distance=2.10cm] (d03) {\includegraphics[width=.12\textwidth,trim={30 30 30 30},clip]{figs/multipie_pose/012/190/output_poisson.jpg}};
        \node[right of=d03, node distance=2.10cm] (d04) {\includegraphics[width=.12\textwidth,trim={30 30 30 30},clip]{figs/multipie_pose/012/041/output_poisson.jpg}};
        \node[right of=d04, node distance=2.10cm] (d05) {\includegraphics[width=.12\textwidth,trim={30 30 30 30},clip]{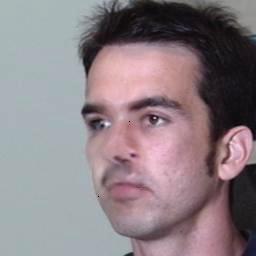}};
        \node[right of=d05, node distance=2.10cm] (d06) {\includegraphics[width=.12\textwidth,trim={30 30 30 30},clip]{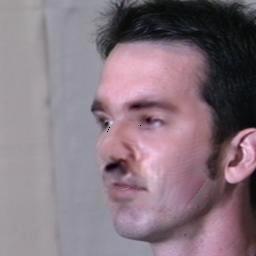}};
        \node[right of=d06, node distance=2.10cm] (d07) {\includegraphics[width=.12\textwidth,trim={30 30 30 30},clip]{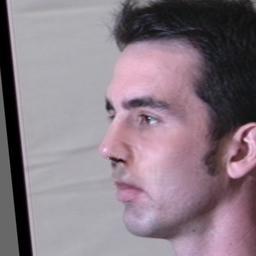}};
        \node[right of=d07, node distance=2.10cm] (d08) {\includegraphics[width=.12\textwidth,trim={30 30 30 30},clip]{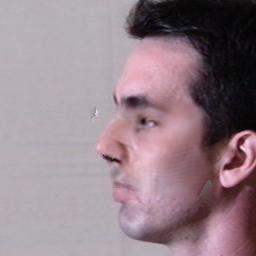}};
        \node[left of=d01, node distance=1.35cm, rotate=90, text centered] (d00) {\ourmethod{}};
        
        \node[below of=d01, node distance=2.10cm] (e01) {\includegraphics[width=.12\textwidth,trim={30 30 30 30},clip]{figs/multipie_pose/012/010/original.jpg}};
        \node[right of=e01, node distance=2.10cm] (e02) {\includegraphics[width=.12\textwidth,trim={30 30 30 30},clip]{figs/multipie_pose/012/200/original.jpg}};
        \node[right of=e02, node distance=2.10cm] (e03) {\includegraphics[width=.12\textwidth,trim={30 30 30 30},clip]{figs/multipie_pose/012/190/original.jpg}};
        \node[right of=e03, node distance=2.10cm] (e04) {\includegraphics[width=.12\textwidth,trim={30 30 30 30},clip]{figs/multipie_pose/012/041/original.jpg}};
        \node[right of=e04, node distance=2.10cm] (e05) {\includegraphics[width=.12\textwidth,trim={30 30 30 30},clip]{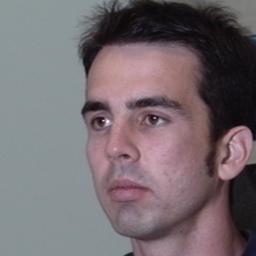}};
        \node[right of=e05, node distance=2.10cm] (e06) {\includegraphics[width=.12\textwidth,trim={30 30 30 30},clip]{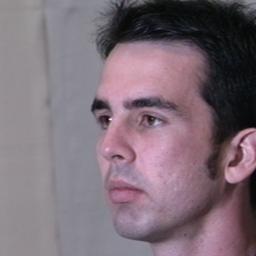}};
        \node[right of=e06, node distance=2.10cm] (e07) {\includegraphics[width=.12\textwidth,trim={30 30 30 30},clip]{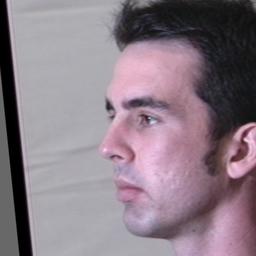}};
        \node[right of=e07, node distance=2.10cm] (e08) {\includegraphics[width=.12\textwidth,trim={30 30 30 30},clip]{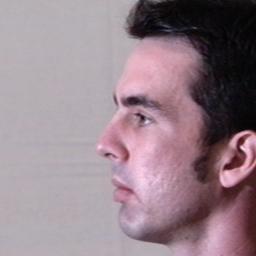}};
        \node[left of=e01, node distance=1.35cm, rotate=90, text centered] (e00) {Ground Truth};
        
        \node[below of=e01, node distance=2.30cm] (a11) {\includegraphics[width=.12\textwidth,trim={30 30 30 30},clip]{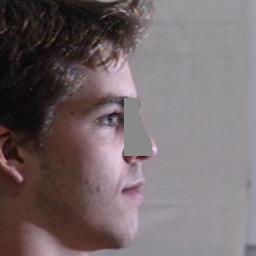}};
        \node[right of=a11, node distance=2.10cm] (a12) {\includegraphics[width=.12\textwidth,trim={30 30 30 30},clip]{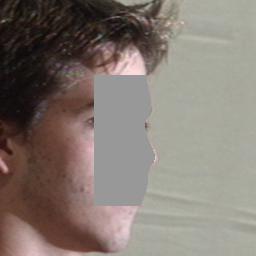}};
        \node[right of=a12, node distance=2.10cm] (a13) {\includegraphics[width=.12\textwidth,trim={30 30 30 30},clip]{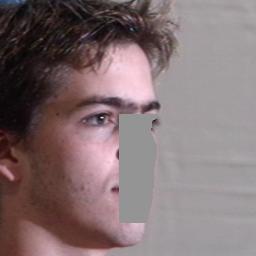}};
        \node[right of=a13, node distance=2.10cm] (a14) {\includegraphics[width=.12\textwidth,trim={30 30 30 30},clip]{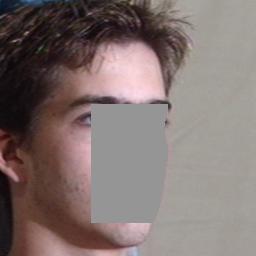}};
        \node[right of=a14, node distance=2.10cm] (a15) {\includegraphics[width=.12\textwidth,trim={30 30 30 30},clip]{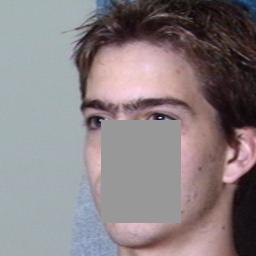}};
        \node[right of=a15, node distance=2.10cm] (a16) {\includegraphics[width=.12\textwidth,trim={30 30 30 30},clip]{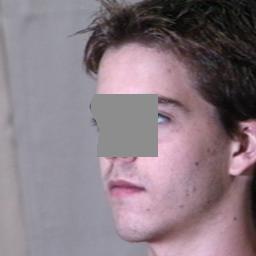}};
        \node[right of=a16, node distance=2.10cm] (a17) {\includegraphics[width=.12\textwidth,trim={30 30 30 30},clip]{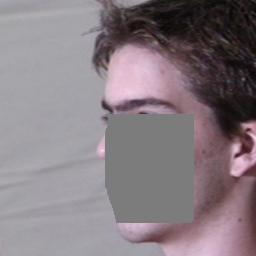}};
        \node[right of=a17, node distance=2.10cm] (a18) {\includegraphics[width=.12\textwidth,trim={30 30 30 30},clip]{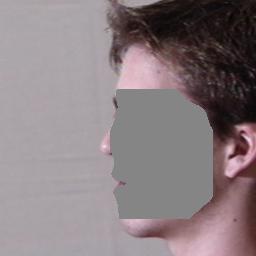}};
        \node[left of=a11, node distance=1.35cm, rotate=90, text centered] (a10) {Input};

        \node[below of=a11, node distance=2.10cm] (b11) {\includegraphics[width=.12\textwidth,trim={30 30 30 30},clip]{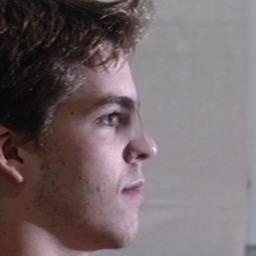}};
        \node[right of=b11, node distance=2.10cm] (b12) {\includegraphics[width=.12\textwidth,trim={30 30 30 30},clip]{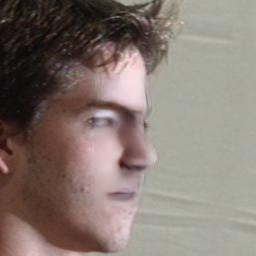}};
        \node[right of=b12, node distance=2.10cm] (b13) {\includegraphics[width=.12\textwidth,trim={30 30 30 30},clip]{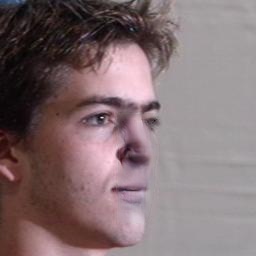}};
        \node[right of=b13, node distance=2.10cm] (b14) {\includegraphics[width=.12\textwidth,trim={30 30 30 30},clip]{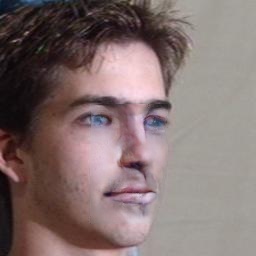}};
        \node[right of=b14, node distance=2.10cm] (b15) {\includegraphics[width=.12\textwidth,trim={30 30 30 30},clip]{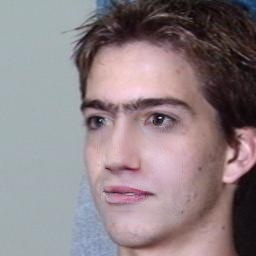}};
        \node[right of=b15, node distance=2.10cm] (b16) {\includegraphics[width=.12\textwidth,trim={30 30 30 30},clip]{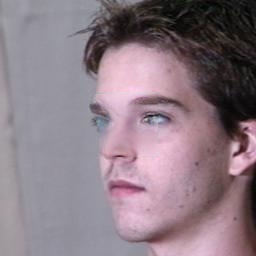}};
        \node[right of=b16, node distance=2.10cm] (b17) {\includegraphics[width=.12\textwidth,trim={30 30 30 30},clip]{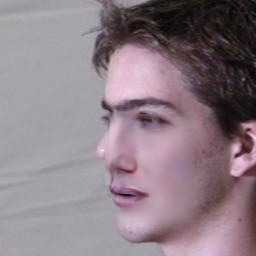}};
        \node[right of=b17, node distance=2.10cm] (b18) {\includegraphics[width=.12\textwidth,trim={30 30 30 30},clip]{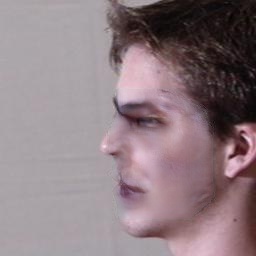}};
        \node[left of=b11, node distance=1.35cm, rotate=90, text centered] (b10) {DeepFillv2 \cite{freeforminpainting}};
        
        \node[below of=b11, node distance=2.10cm] (c11) {\includegraphics[width=.12\textwidth,trim={30 30 30 30},clip]{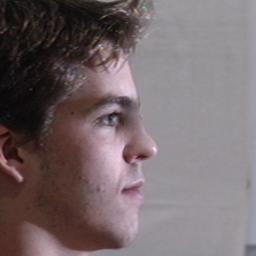}};
        \node[right of=c11, node distance=2.10cm] (c12) {\includegraphics[width=.12\textwidth,trim={30 30 30 30},clip]{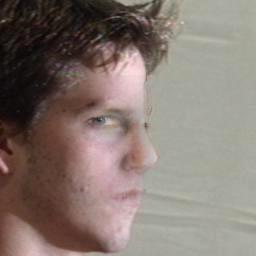}};
        \node[right of=c12, node distance=2.10cm] (c13) {\includegraphics[width=.12\textwidth,trim={30 30 30 30},clip]{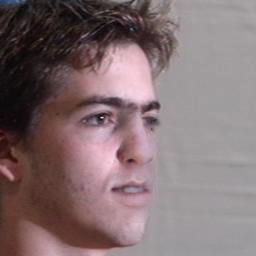}};
        \node[right of=c13, node distance=2.10cm] (c14) {\includegraphics[width=.12\textwidth,trim={30 30 30 30},clip]{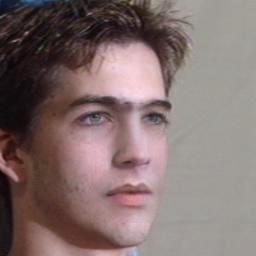}};
        \node[right of=c14, node distance=2.10cm] (c15) {\includegraphics[width=.12\textwidth,trim={30 30 30 30},clip]{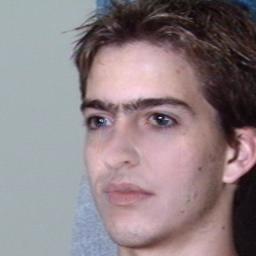}};
        \node[right of=c15, node distance=2.10cm] (c16) {\includegraphics[width=.12\textwidth,trim={30 30 30 30},clip]{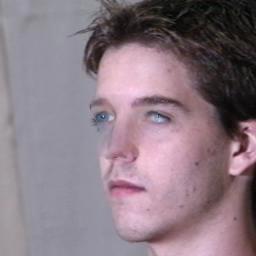}};
        \node[right of=c16, node distance=2.10cm] (c17) {\includegraphics[width=.12\textwidth,trim={30 30 30 30},clip]{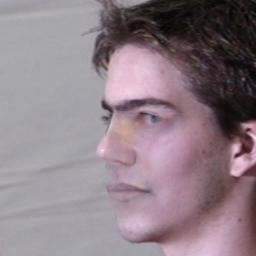}};
        \node[right of=c17, node distance=2.10cm] (c18) {\includegraphics[width=.12\textwidth,trim={30 30 30 30},clip]{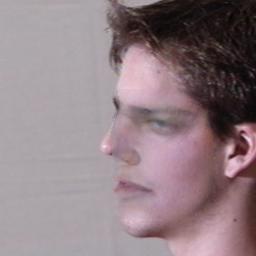}};
        \node[left of=c11, node distance=1.35cm, rotate=90, text centered] (c10) {PICNet \cite{zheng2019pluralistic}};
        
        \node[below of=c11, node distance=2.10cm] (d11) {\includegraphics[width=.12\textwidth,trim={30 30 30 30},clip]{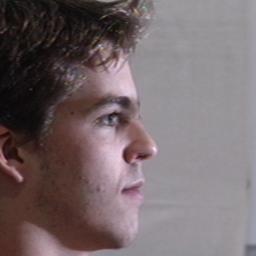}};
        \node[right of=d11, node distance=2.10cm] (d12) {\includegraphics[width=.12\textwidth,trim={30 30 30 30},clip]{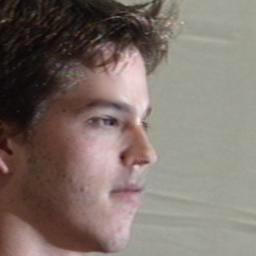}};
        \node[right of=d12, node distance=2.10cm] (d13) {\includegraphics[width=.12\textwidth,trim={30 30 30 30},clip]{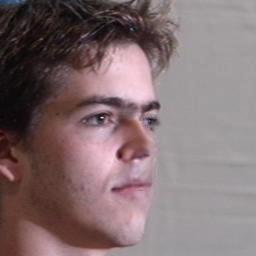}};
        \node[right of=d13, node distance=2.10cm] (d14) {\includegraphics[width=.12\textwidth,trim={30 30 30 30},clip]{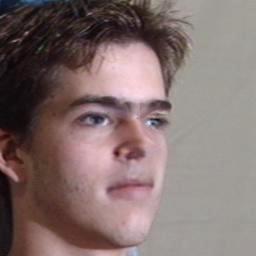}};
        \node[right of=d14, node distance=2.10cm] (d15) {\includegraphics[width=.12\textwidth,trim={30 30 30 30},clip]{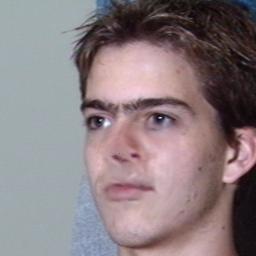}};
        \node[right of=d15, node distance=2.10cm] (d16) {\includegraphics[width=.12\textwidth,trim={30 30 30 30},clip]{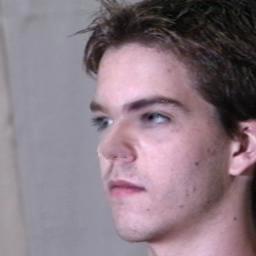}};
        \node[right of=d16, node distance=2.10cm] (d17) {\includegraphics[width=.12\textwidth,trim={30 30 30 30},clip]{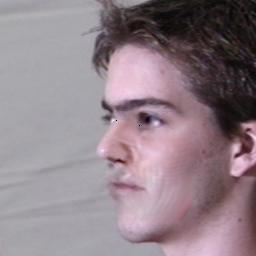}};
        \node[right of=d17, node distance=2.10cm] (d18) {\includegraphics[width=.12\textwidth,trim={30 30 30 30},clip]{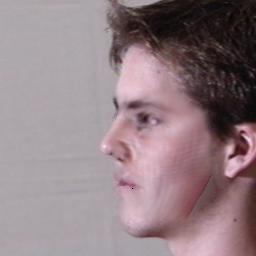}};
        \node[left of=d11, node distance=1.35cm, rotate=90, text centered] (d10) {\ourmethod{}};
        
        \node[below of=d11, node distance=2.10cm] (e11) {\includegraphics[width=.12\textwidth,trim={30 30 30 30},clip]{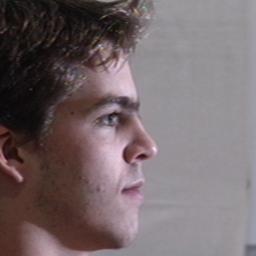}};
        \node[right of=e11, node distance=2.10cm] (e12) {\includegraphics[width=.12\textwidth,trim={30 30 30 30},clip]{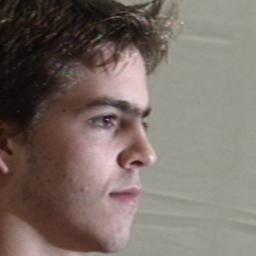}};
        \node[right of=e12, node distance=2.10cm] (e13) {\includegraphics[width=.12\textwidth,trim={30 30 30 30},clip]{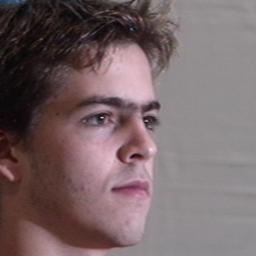}};
        \node[right of=e13, node distance=2.10cm] (e14) {\includegraphics[width=.12\textwidth,trim={30 30 30 30},clip]{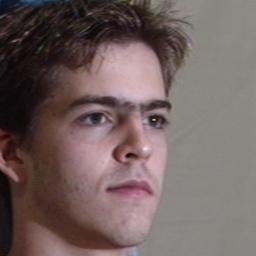}};
        \node[right of=e14, node distance=2.10cm] (e15) {\includegraphics[width=.12\textwidth,trim={30 30 30 30},clip]{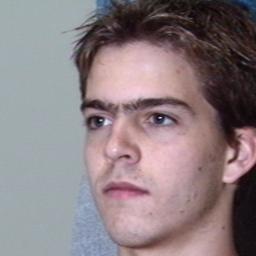}};
        \node[right of=e15, node distance=2.10cm] (e16) {\includegraphics[width=.12\textwidth,trim={30 30 30 30},clip]{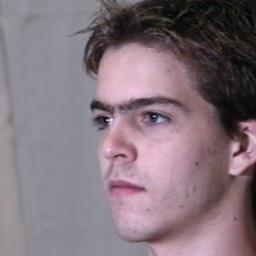}};
        \node[right of=e16, node distance=2.10cm] (e17) {\includegraphics[width=.12\textwidth,trim={30 30 30 30},clip]{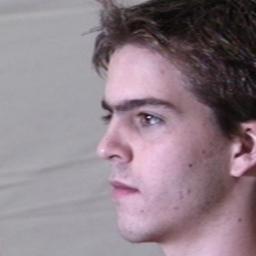}};
        \node[right of=e17, node distance=2.10cm] (e18) {\includegraphics[width=.12\textwidth,trim={30 30 30 30},clip]{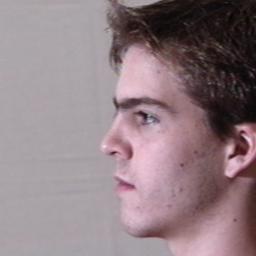}};
        \node[left of=e11, node distance=1.35cm, rotate=90, text centered] (e10) {Ground Truth};
    \end{tikzpicture}
    \caption{Qualitative evaluation of \ourmethod{} \versus{} baselines DeepFillv2 \cite{freeforminpainting} and PIC \cite{zheng2019pluralistic} on the pose-varying MultiPIE:Pose split \cite{multipie}. While the baselines tend to generate blurred and deformed faces in extreme poses, \ourmethod{} is pose-robust and generates more accurate completions across a range of pose.}
    \label{fig:multipie_pose}
\end{figure*}

\begin{figure*}
    \scriptsize
    \centering
    \tikzstyle{block} = [rectangle, draw, fill=blue!20, text centered]
    \begin{tikzpicture}
        \node[rotate=90, text centered] (a00) {Input};
        \node[right of=a00, node distance=1.35cm] (a01) {\includegraphics[width=.12\textwidth,trim={30 30 30 30},clip]{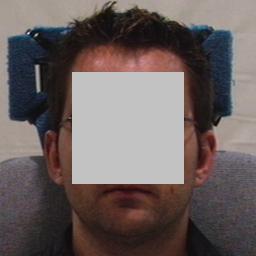}};
        \node[right of=a01, node distance=2.10cm] (a02) {\includegraphics[width=.12\textwidth,trim={30 30 30 30},clip]{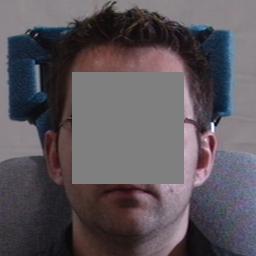}};
        \node[right of=a02, node distance=2.10cm] (a03) {\includegraphics[width=.12\textwidth,trim={30 30 30 30},clip]{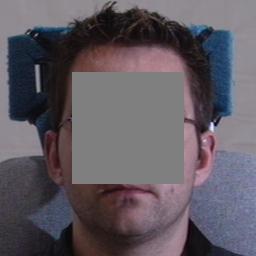}};
        \node[right of=a03, node distance=2.10cm] (a04) {\includegraphics[width=.12\textwidth,trim={30 30 30 30},clip]{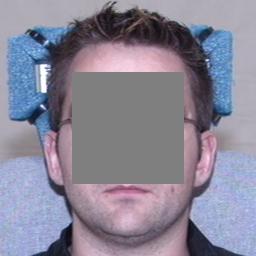}};
        \node[right of=a04, node distance=2.10cm] (a05) {\includegraphics[width=.12\textwidth,trim={30 30 30 30},clip]{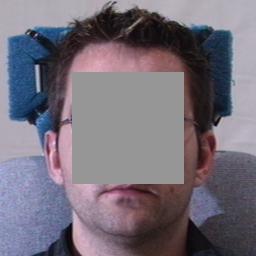}};
        \node[right of=a05, node distance=2.10cm] (a06) {\includegraphics[width=.12\textwidth,trim={30 30 30 30},clip]{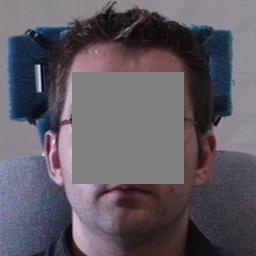}};
        \node[right of=a06, node distance=2.10cm] (a07) {\includegraphics[width=.12\textwidth,trim={30 30 30 30},clip]{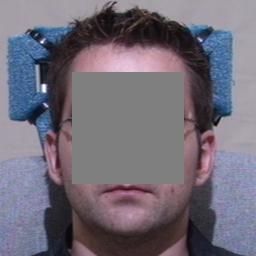}};
        \node[right of=a07, node distance=2.10cm] (a08) {\includegraphics[width=.12\textwidth,trim={30 30 30 30},clip]{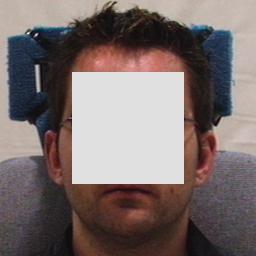}};

        \node[below of=a01, node distance=2.10cm] (b01) {\includegraphics[width=.12\textwidth,trim={30 30 30 30},clip]{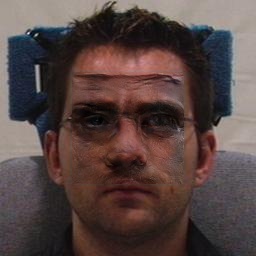}};
        \node[right of=b01, node distance=2.10cm] (b02) {\includegraphics[width=.12\textwidth,trim={30 30 30 30},clip]{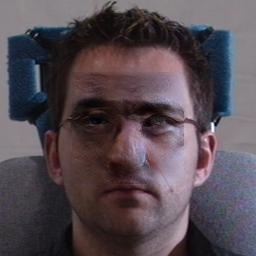}};
        \node[right of=b02, node distance=2.10cm] (b03) {\includegraphics[width=.12\textwidth,trim={30 30 30 30},clip]{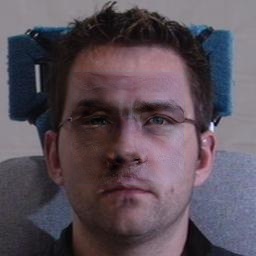}};
        \node[right of=b03, node distance=2.10cm] (b04) {\includegraphics[width=.12\textwidth,trim={30 30 30 30},clip]{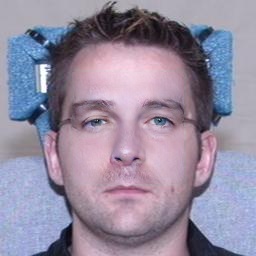}};
        \node[right of=b04, node distance=2.10cm] (b05) {\includegraphics[width=.12\textwidth,trim={30 30 30 30},clip]{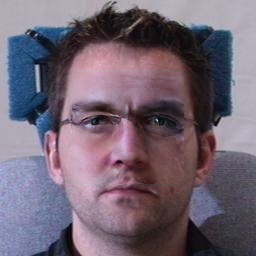}};
        \node[right of=b05, node distance=2.10cm] (b06) {\includegraphics[width=.12\textwidth,trim={30 30 30 30},clip]{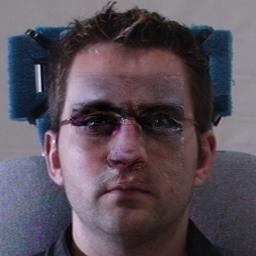}};
        \node[right of=b06, node distance=2.10cm] (b07) {\includegraphics[width=.12\textwidth,trim={30 30 30 30},clip]{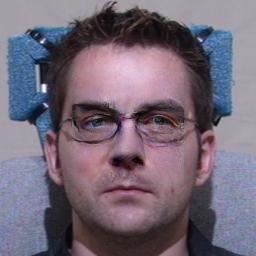}};
        \node[right of=b07, node distance=2.10cm] (b08) {\includegraphics[width=.12\textwidth,trim={30 30 30 30},clip]{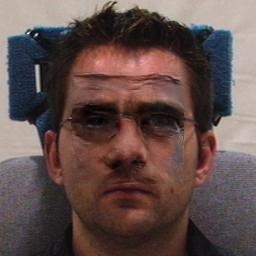}};
        \node[left of=b01, node distance=1.35cm, rotate=90, text centered] (b00) {DeepFillv2 \cite{freeforminpainting}};
        
        \node[below of=b01, node distance=2.10cm] (c01) {\includegraphics[width=.12\textwidth,trim={30 30 30 30},clip]{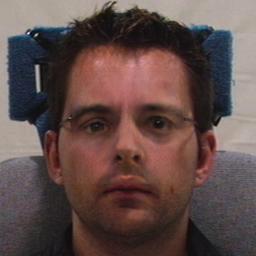}};
        \node[right of=c01, node distance=2.10cm] (c02) {\includegraphics[width=.12\textwidth,trim={30 30 30 30},clip]{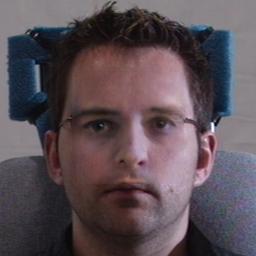}};
        \node[right of=c02, node distance=2.10cm] (c03) {\includegraphics[width=.12\textwidth,trim={30 30 30 30},clip]{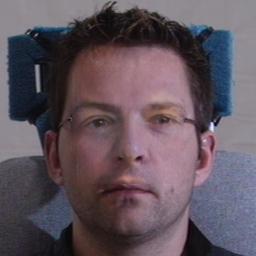}};
        \node[right of=c03, node distance=2.10cm] (c04) {\includegraphics[width=.12\textwidth,trim={30 30 30 30},clip]{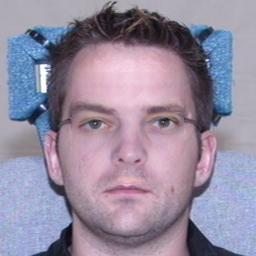}};
        \node[right of=c04, node distance=2.10cm] (c05) {\includegraphics[width=.12\textwidth,trim={30 30 30 30},clip]{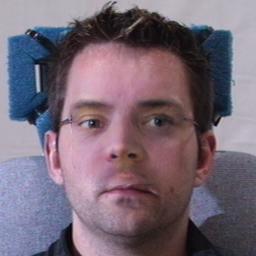}};
        \node[right of=c05, node distance=2.10cm] (c06) {\includegraphics[width=.12\textwidth,trim={30 30 30 30},clip]{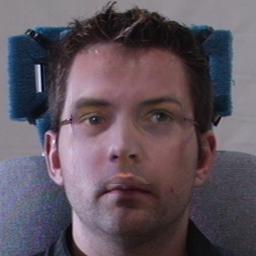}};
        \node[right of=c06, node distance=2.10cm] (c07) {\includegraphics[width=.12\textwidth,trim={30 30 30 30},clip]{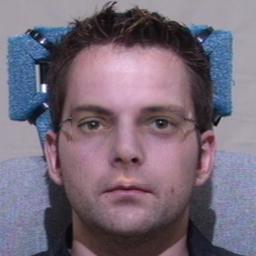}};
        \node[right of=c07, node distance=2.10cm] (c08) {\includegraphics[width=.12\textwidth,trim={30 30 30 30},clip]{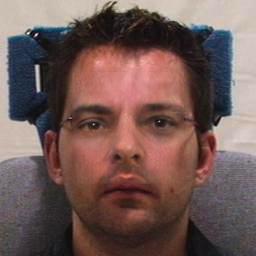}};
        \node[left of=c01, node distance=1.35cm, rotate=90, text centered] (c00) {PICNet \cite{zheng2019pluralistic}};
        
        \node[below of=c01, node distance=2.10cm] (d01) {\includegraphics[width=.12\textwidth,trim={30 30 30 30},clip]{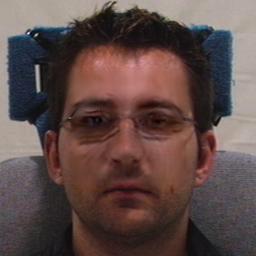}};
        \node[right of=d01, node distance=2.10cm] (d02) {\includegraphics[width=.12\textwidth,trim={30 30 30 30},clip]{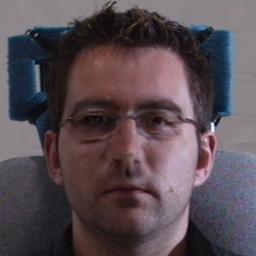}};
        \node[right of=d02, node distance=2.10cm] (d03) {\includegraphics[width=.12\textwidth,trim={30 30 30 30},clip]{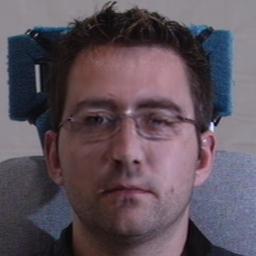}};
        \node[right of=d03, node distance=2.10cm] (d04) {\includegraphics[width=.12\textwidth,trim={30 30 30 30},clip]{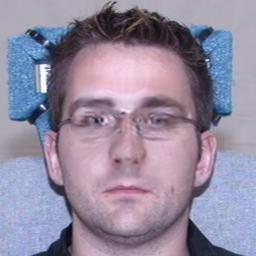}};
        \node[right of=d04, node distance=2.10cm] (d05) {\includegraphics[width=.12\textwidth,trim={30 30 30 30},clip]{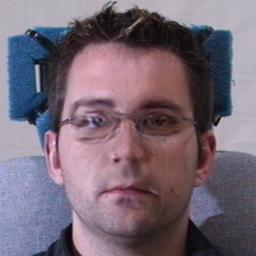}};
        \node[right of=d05, node distance=2.10cm] (d06) {\includegraphics[width=.12\textwidth,trim={30 30 30 30},clip]{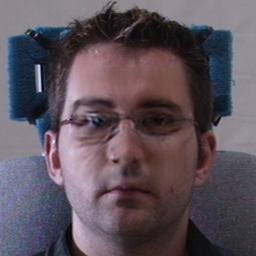}};
        \node[right of=d06, node distance=2.10cm] (d07) {\includegraphics[width=.12\textwidth,trim={30 30 30 30},clip]{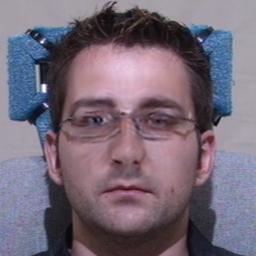}};
        \node[right of=d07, node distance=2.10cm] (d08) {\includegraphics[width=.12\textwidth,trim={30 30 30 30},clip]{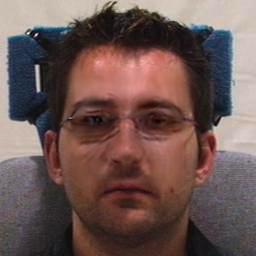}};
        \node[left of=d01, node distance=1.35cm, rotate=90, text centered] (d00) {\ourmethod{}};
        
        \node[below of=d01, node distance=2.10cm] (e01) {\includegraphics[width=.12\textwidth,trim={30 30 30 30},clip]{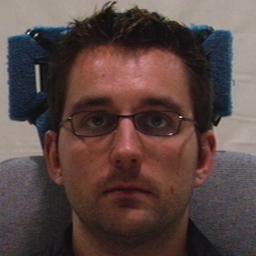}};
        \node[right of=e01, node distance=2.10cm] (e02) {\includegraphics[width=.12\textwidth,trim={30 30 30 30},clip]{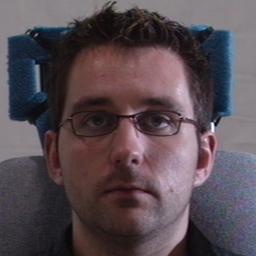}};
        \node[right of=e02, node distance=2.10cm] (e03) {\includegraphics[width=.12\textwidth,trim={30 30 30 30},clip]{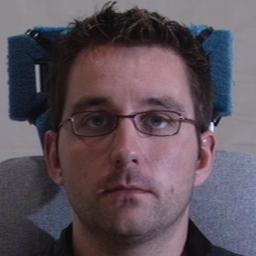}};
        \node[right of=e03, node distance=2.10cm] (e04) {\includegraphics[width=.12\textwidth,trim={30 30 30 30},clip]{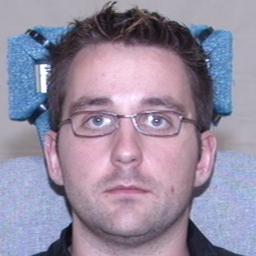}};
        \node[right of=e04, node distance=2.10cm] (e05) {\includegraphics[width=.12\textwidth,trim={30 30 30 30},clip]{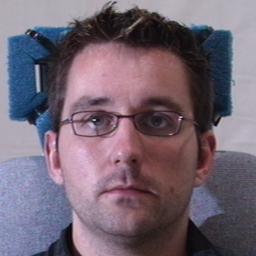}};
        \node[right of=e05, node distance=2.10cm] (e06) {\includegraphics[width=.12\textwidth,trim={30 30 30 30},clip]{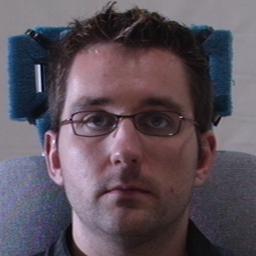}};
        \node[right of=e06, node distance=2.10cm] (e07) {\includegraphics[width=.12\textwidth,trim={30 30 30 30},clip]{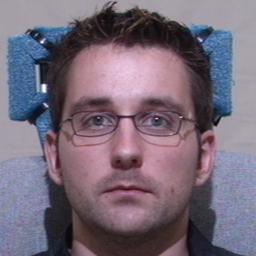}};
        \node[right of=e07, node distance=2.10cm] (e08) {\includegraphics[width=.12\textwidth,trim={30 30 30 30},clip]{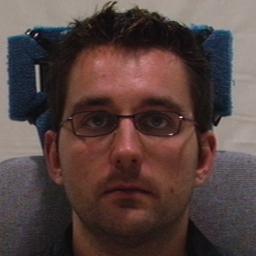}};
        \node[left of=e01, node distance=1.35cm, rotate=90, text centered] (e00) {Ground Truth};
        
        \node[below of=e01, node distance=2.30cm] (a11) {\includegraphics[width=.12\textwidth,trim={30 30 30 30},clip]{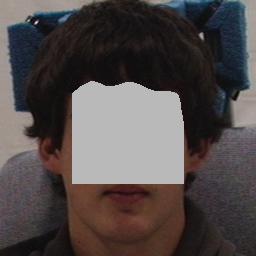}};
        \node[right of=a11, node distance=2.10cm] (a12) {\includegraphics[width=.12\textwidth,trim={30 30 30 30},clip]{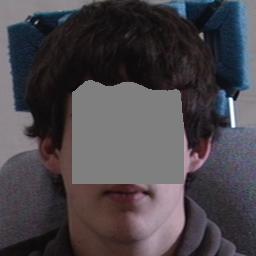}};
        \node[right of=a12, node distance=2.10cm] (a13) {\includegraphics[width=.12\textwidth,trim={30 30 30 30},clip]{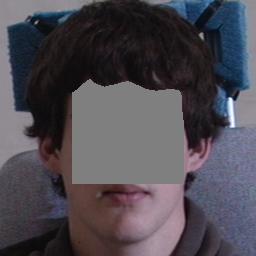}};
        \node[right of=a13, node distance=2.10cm] (a14) {\includegraphics[width=.12\textwidth,trim={30 30 30 30},clip]{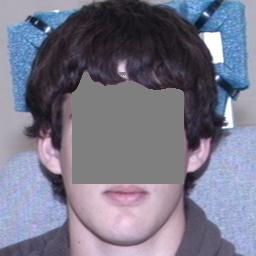}};
        \node[right of=a14, node distance=2.10cm] (a15) {\includegraphics[width=.12\textwidth,trim={30 30 30 30},clip]{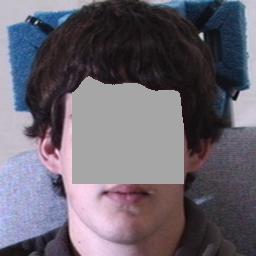}};
        \node[right of=a15, node distance=2.10cm] (a16) {\includegraphics[width=.12\textwidth,trim={30 30 30 30},clip]{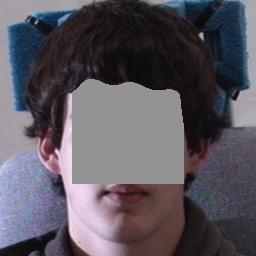}};
        \node[right of=a16, node distance=2.10cm] (a17) {\includegraphics[width=.12\textwidth,trim={30 30 30 30},clip]{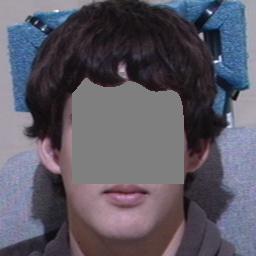}};
        \node[right of=a17, node distance=2.10cm] (a18) {\includegraphics[width=.12\textwidth,trim={30 30 30 30},clip]{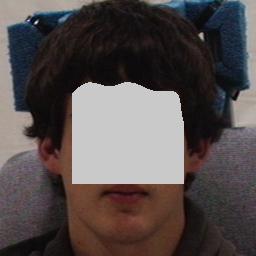}};
        \node[left of=a11, node distance=1.35cm, rotate=90, text centered] (a10) {Input};

        \node[below of=a11, node distance=2.10cm] (b11) {\includegraphics[width=.12\textwidth,trim={30 30 30 30},clip]{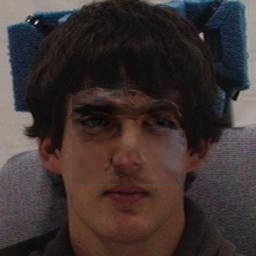}};
        \node[right of=b11, node distance=2.10cm] (b12) {\includegraphics[width=.12\textwidth,trim={30 30 30 30},clip]{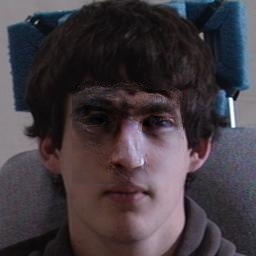}};
        \node[right of=b12, node distance=2.10cm] (b13) {\includegraphics[width=.12\textwidth,trim={30 30 30 30},clip]{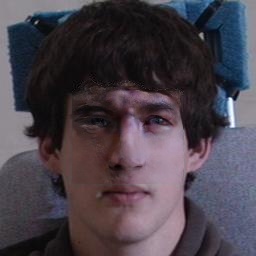}};
        \node[right of=b13, node distance=2.10cm] (b14) {\includegraphics[width=.12\textwidth,trim={30 30 30 30},clip]{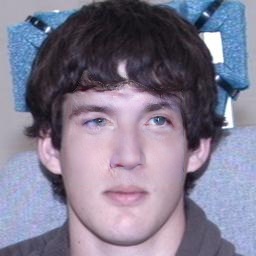}};
        \node[right of=b14, node distance=2.10cm] (b15) {\includegraphics[width=.12\textwidth,trim={30 30 30 30},clip]{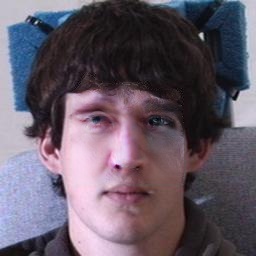}};
        \node[right of=b15, node distance=2.10cm] (b16) {\includegraphics[width=.12\textwidth,trim={30 30 30 30},clip]{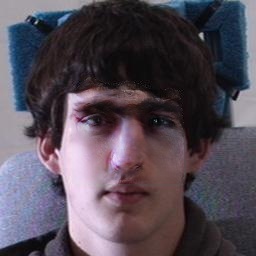}};
        \node[right of=b16, node distance=2.10cm] (b17) {\includegraphics[width=.12\textwidth,trim={30 30 30 30},clip]{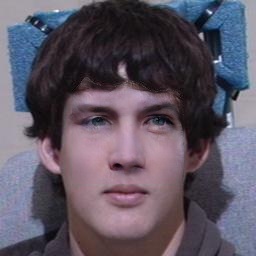}};
        \node[right of=b17, node distance=2.10cm] (b18) {\includegraphics[width=.12\textwidth,trim={30 30 30 30},clip]{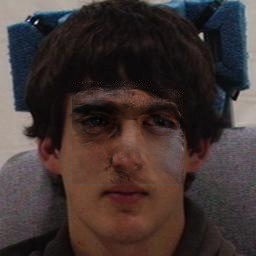}};
        \node[left of=b11, node distance=1.35cm, rotate=90, text centered] (b10) {DeepFillv2 \cite{freeforminpainting}};
        
        \node[below of=b11, node distance=2.10cm] (c11) {\includegraphics[width=.12\textwidth,trim={30 30 30 30},clip]{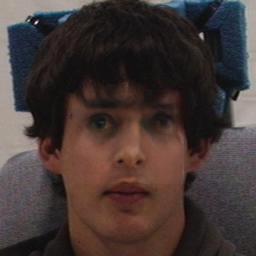}};
        \node[right of=c11, node distance=2.10cm] (c12) {\includegraphics[width=.12\textwidth,trim={30 30 30 30},clip]{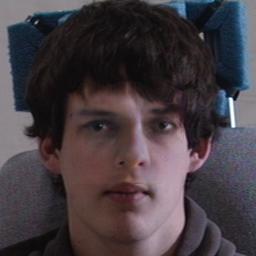}};
        \node[right of=c12, node distance=2.10cm] (c13) {\includegraphics[width=.12\textwidth,trim={30 30 30 30},clip]{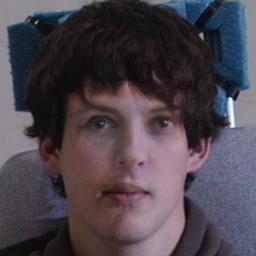}};
        \node[right of=c13, node distance=2.10cm] (c14) {\includegraphics[width=.12\textwidth,trim={30 30 30 30},clip]{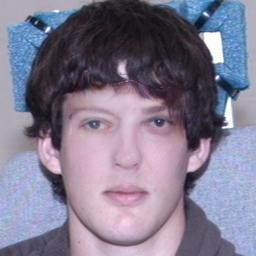}};
        \node[right of=c14, node distance=2.10cm] (c15) {\includegraphics[width=.12\textwidth,trim={30 30 30 30},clip]{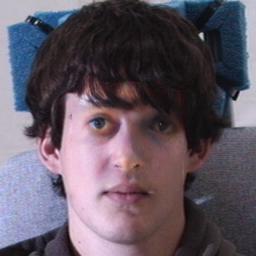}};
        \node[right of=c15, node distance=2.10cm] (c16) {\includegraphics[width=.12\textwidth,trim={30 30 30 30},clip]{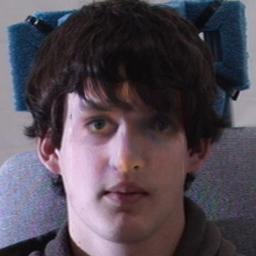}};
        \node[right of=c16, node distance=2.10cm] (c17) {\includegraphics[width=.12\textwidth,trim={30 30 30 30},clip]{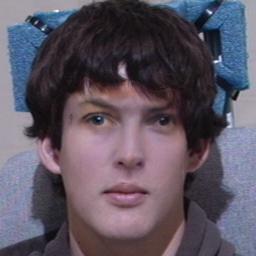}};
        \node[right of=c17, node distance=2.10cm] (c18) {\includegraphics[width=.12\textwidth,trim={30 30 30 30},clip]{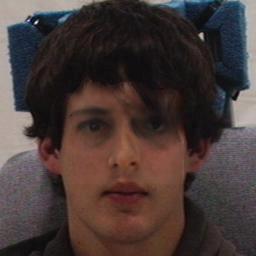}};
        \node[left of=c11, node distance=1.35cm, rotate=90, text centered] (c10) {PICNet \cite{zheng2019pluralistic}};
        
        \node[below of=c11, node distance=2.10cm] (d11) {\includegraphics[width=.12\textwidth,trim={30 30 30 30},clip]{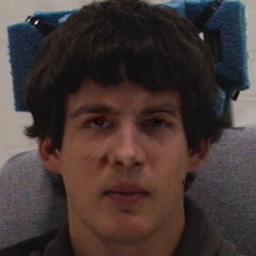}};
        \node[right of=d11, node distance=2.10cm] (d12) {\includegraphics[width=.12\textwidth,trim={30 30 30 30},clip]{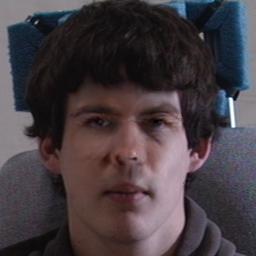}};
        \node[right of=d12, node distance=2.10cm] (d13) {\includegraphics[width=.12\textwidth,trim={30 30 30 30},clip]{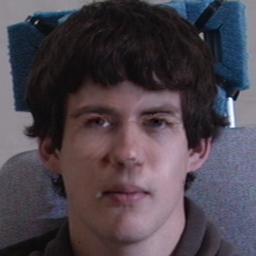}};
        \node[right of=d13, node distance=2.10cm] (d14) {\includegraphics[width=.12\textwidth,trim={30 30 30 30},clip]{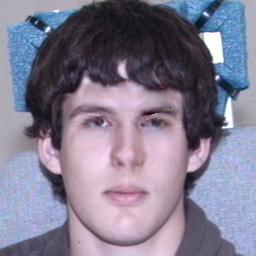}};
        \node[right of=d14, node distance=2.10cm] (d15) {\includegraphics[width=.12\textwidth,trim={30 30 30 30},clip]{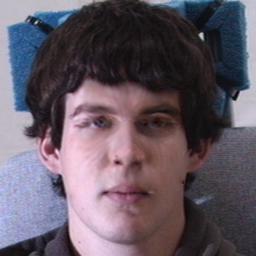}};
        \node[right of=d15, node distance=2.10cm] (d16) {\includegraphics[width=.12\textwidth,trim={30 30 30 30},clip]{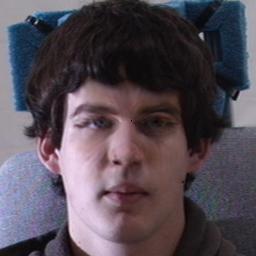}};
        \node[right of=d16, node distance=2.10cm] (d17) {\includegraphics[width=.12\textwidth,trim={30 30 30 30},clip]{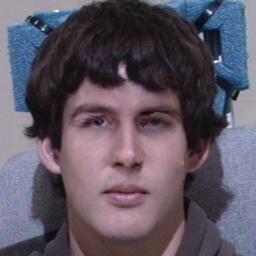}};
        \node[right of=d17, node distance=2.10cm] (d18) {\includegraphics[width=.12\textwidth,trim={30 30 30 30},clip]{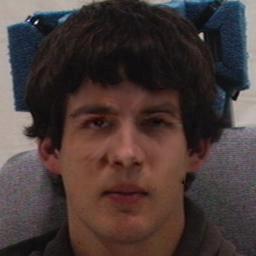}};
        \node[left of=d11, node distance=1.35cm, rotate=90, text centered] (d10) {\ourmethod{}};
        
        \node[below of=d11, node distance=2.10cm] (e11) {\includegraphics[width=.12\textwidth,trim={30 30 30 30},clip]{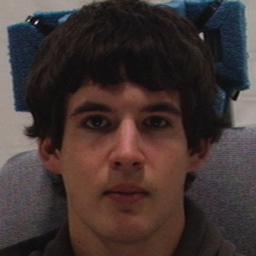}};
        \node[right of=e11, node distance=2.10cm] (e12) {\includegraphics[width=.12\textwidth,trim={30 30 30 30},clip]{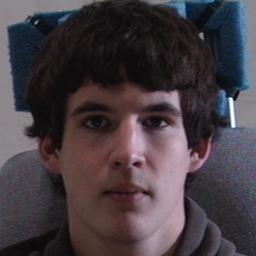}};
        \node[right of=e12, node distance=2.10cm] (e13) {\includegraphics[width=.12\textwidth,trim={30 30 30 30},clip]{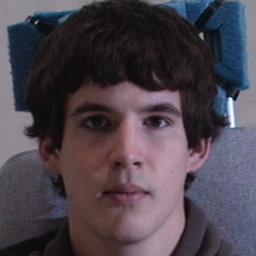}};
        \node[right of=e13, node distance=2.10cm] (e14) {\includegraphics[width=.12\textwidth,trim={30 30 30 30},clip]{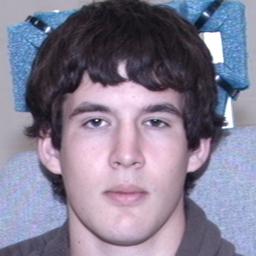}};
        \node[right of=e14, node distance=2.10cm] (e15) {\includegraphics[width=.12\textwidth,trim={30 30 30 30},clip]{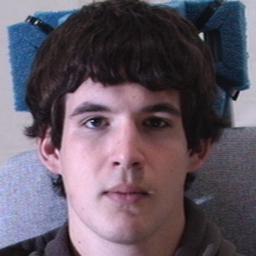}};
        \node[right of=e15, node distance=2.10cm] (e16) {\includegraphics[width=.12\textwidth,trim={30 30 30 30},clip]{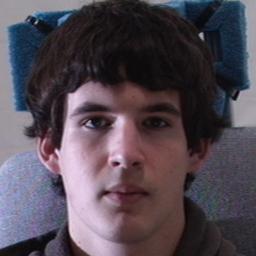}};
        \node[right of=e16, node distance=2.10cm] (e17) {\includegraphics[width=.12\textwidth,trim={30 30 30 30},clip]{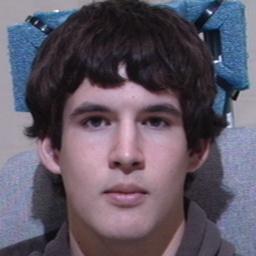}};
        \node[right of=e17, node distance=2.10cm] (e18) {\includegraphics[width=.12\textwidth,trim={30 30 30 30},clip]{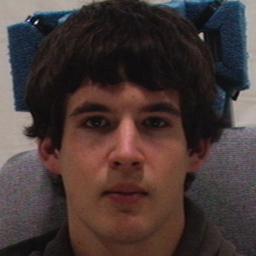}};
        \node[left of=e11, node distance=1.35cm, rotate=90, text centered] (e10) {Ground Truth};
    \end{tikzpicture}
    \caption{Qualitative evaluation of \ourmethod{} \versus{}the baselines DeepFillv2 \cite{freeforminpainting} and PIC \cite{zheng2019pluralistic} on the illumination varying MultiPIE:Illu split \cite{multipie}. While the baselines tend to generate artifacts in extreme illuminations, \ourmethod{} generates completions that look geometrically accurate and preserve the illumination contrast (notice (i) the illumination contrast in cols. 2,3,5,6 (b), and (ii) assymetric eye-brows in cols. 1,2,3,5,6 (b) by the baselines.)}
    \label{fig:multipie_illu}
\end{figure*}

\vspace{8pt}
\noindent\textbf{B: Robustness Across Pose and Illumination Variations}

We present further cross-dataset evaluation on the pose and illumination varying MultiPIE dataset \cite{multipie} by splitting the dataset into two subsets: (1) a pose varying subset with constant frontal illumination and expression, referred to as MultiPIE:Pose and (2) an illumination varying subset with constant frontal pose and expression, referred to as MultiPIE:Illu. Table \ref{tab:quantitative} reports the PSNR, SSIM and LPIPS \cite{lpips} metrics for all the methods on these two splits. It can be seen that \ourmethod{} significantly outperforms the baselines in both the splits. Further, we show more example completions by \ourmethod{} \versus{} the baselines DeepFillv2 \cite{freeforminpainting} and PIC \cite{zheng2019pluralistic} in Fig.~\ref{fig:multipie_pose} (for Pose) and Fig.~\ref{fig:multipie_illu} (for Illumination), respectively. From Fig.~\ref{fig:multipie_pose}, one can observe that the baselines tend to generate fuzzy and deformed faces for extreme poses while \ourmethod{} generates sharper and geometry-preserving completions. And, in the illumination-varying case, DeepFillv2 \cite{freeforminpainting} tends to generate artifacts and PIC \cite{zheng2019pluralistic} tends to generate asymmetric completions for extreme illumination, whereas the completions by \ourmethod{} are free of such artifacts and preserve illumination contrast and symmetry.

\begin{figure*}
    \centering
    \tikzstyle{block} = [rectangle, draw, fill=blue!20, text centered]
    \begin{tikzpicture}
    \node[] (a1) {\includegraphics[width=.18\textwidth,trim={20 20 20 20},clip]{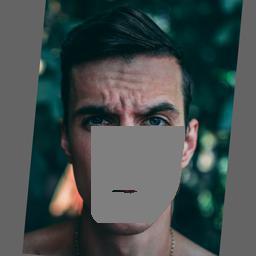}};
    \node[right of=a1, node distance=3.4cm] (a2) {\includegraphics[width=.18\textwidth,trim={20 20 20 20},clip]{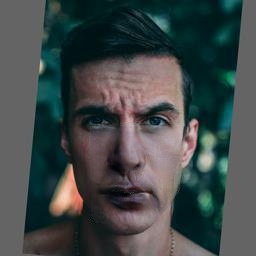}};
    \node[right of=a2, node distance=3.4cm] (a3) {\includegraphics[width=.18\textwidth,trim={20 20 20 20},clip]{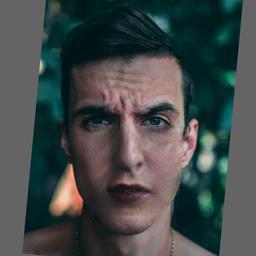}};
    \node[right of=a3, node distance=3.4cm] (a4) {\includegraphics[width=.18\textwidth,trim={20 20 20 20},clip]{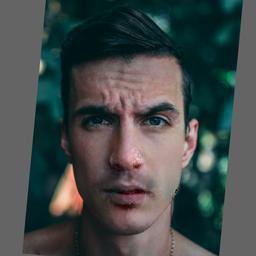}};
    \node[right of=a4, node distance=3.4cm] (a5) {\includegraphics[width=.18\textwidth,trim={20 20 20 20},clip]{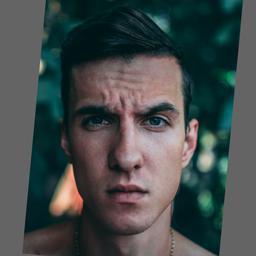}};
    
    \node[below of=a1, node distance=3.4cm] (b1) {\includegraphics[width=.18\textwidth,trim={20 20 20 20},clip]{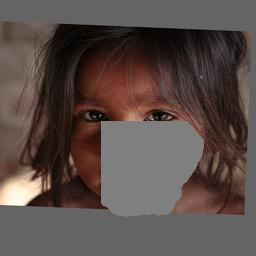}};
    \node[right of=b1, node distance=3.4cm] (b2) {\includegraphics[width=.18\textwidth,trim={20 20 20 20},clip]{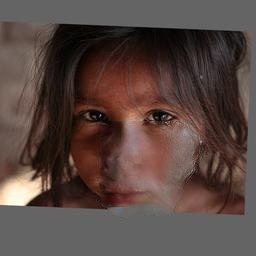}};
    \node[right of=b2, node distance=3.4cm] (b3) {\includegraphics[width=.18\textwidth,trim={20 20 20 20},clip]{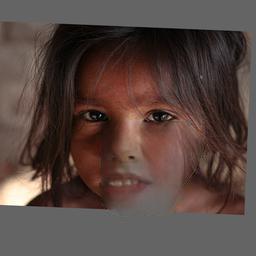}};
    \node[right of=b3, node distance=3.4cm] (b4) {\includegraphics[width=.18\textwidth,trim={20 20 20 20},clip]{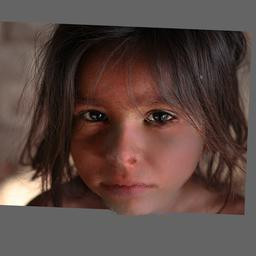}};
    \node[right of=b4, node distance=3.4cm] (b5) {\includegraphics[width=.18\textwidth,trim={20 20 20 20},clip]{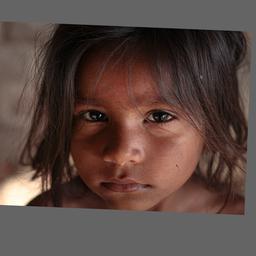}};
    
    \node[below of=b1, node distance=3.4cm] (c1) {\includegraphics[width=.18\textwidth,trim={20 20 20 20},clip]{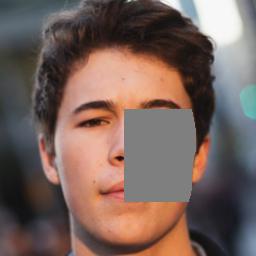}};
    \node[right of=c1, node distance=3.4cm] (c2) {\includegraphics[width=.18\textwidth,trim={20 20 20 20},clip]{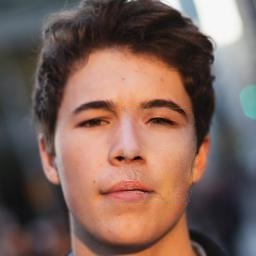}};
    \node[right of=c2, node distance=3.4cm] (c3) {\includegraphics[width=.18\textwidth,trim={20 20 20 20},clip]{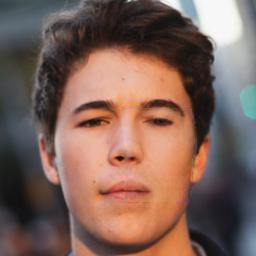}};
    \node[right of=c3, node distance=3.4cm] (c4) {\includegraphics[width=.18\textwidth,trim={20 20 20 20},clip]{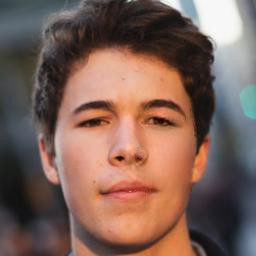}};
    \node[right of=c4, node distance=3.4cm] (c5) {\includegraphics[width=.18\textwidth,trim={20 20 20 20},clip]{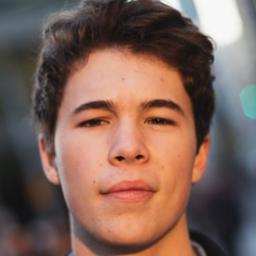}};
    
    \node[below of=c1, node distance=3.4cm] (d1) {\includegraphics[width=.18\textwidth,trim={20 20 20 20},clip]{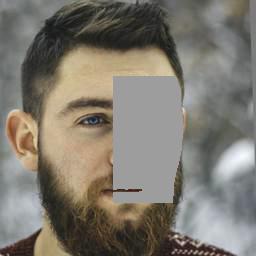}};
    \node[right of=d1, node distance=3.4cm] (d2) {\includegraphics[width=.18\textwidth,trim={20 20 20 20},clip]{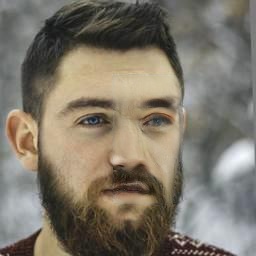}};
    \node[right of=d2, node distance=3.4cm] (d3) {\includegraphics[width=.18\textwidth,trim={20 20 20 20},clip]{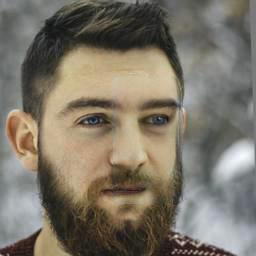}};
    \node[right of=d3, node distance=3.4cm] (d4) {\includegraphics[width=.18\textwidth,trim={20 20 20 20},clip]{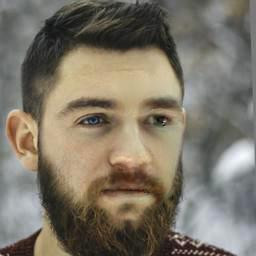}};
    \node[right of=d4, node distance=3.4cm] (d5) {\includegraphics[width=.18\textwidth,trim={20 20 20 20},clip]{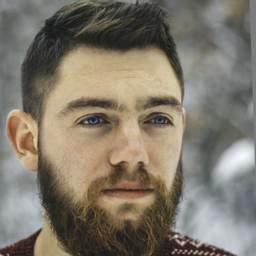}};
    
    \node[below of=d1, node distance=2.0cm] (o1) {Input};
    \node[below of=d2, node distance=2.0cm] (o5) {DeepFillv2 \cite{freeforminpainting}};
    \node[below of=d3, node distance=2.0cm] (o6) {PIC \cite{zheng2019pluralistic}};
    \node[below of=d4, node distance=2.0cm] (o8) {\ourmethod{} (Ours)};
    \node[below of=d5, node distance=2.0cm] (o9) {Ground Truth};
    \end{tikzpicture}
    \vspace{-1mm}
    \caption{Qualitative evaluation (of generalization performance) on the Internet downloaded images.}
    \label{fig:internet}
\end{figure*}
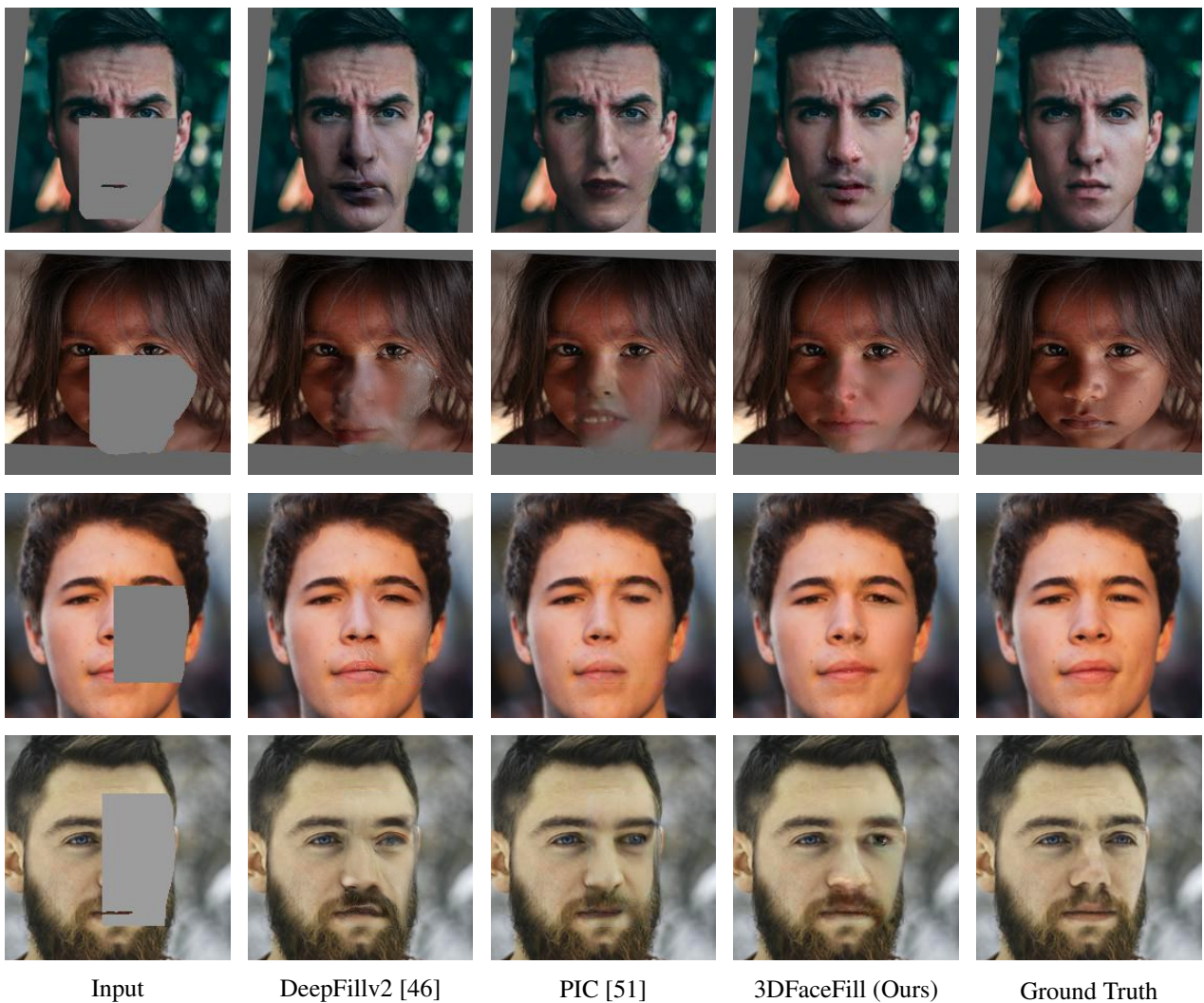

\vspace{8pt}
\noindent\textbf{C: Generalization Performance on In-the-Wild Images downloaded from the Internet} 

To compare the generalization performance of different methods, we evaluate face completion on a small dataset of $\sim$ 50 in-the-wild face images downloaded from the internet\footnote{Source: \url{https://unsplash.com/s/photos/face}} (referred to as Internet). We report the quantitative metrics in Table \ref{tab:quantitative}, where one can see significant margins between \ourmethod{} and the closest baselines across all the three metrics, demonstrating the better generalization performance of our proposed method. Fig.~\ref{fig:internet} shows qualitative comparison on a small sample where \ourmethod{} generates more realistic completions, thanks to the explicit imposition of 3D face priors. This shows that the principles behind \ourmethod{} can improve the generalization performance of image completion approaches on structured objects such as faces. 

\subsection{3D View Synthesis of Masked Faces}\label{subsec:multiview}
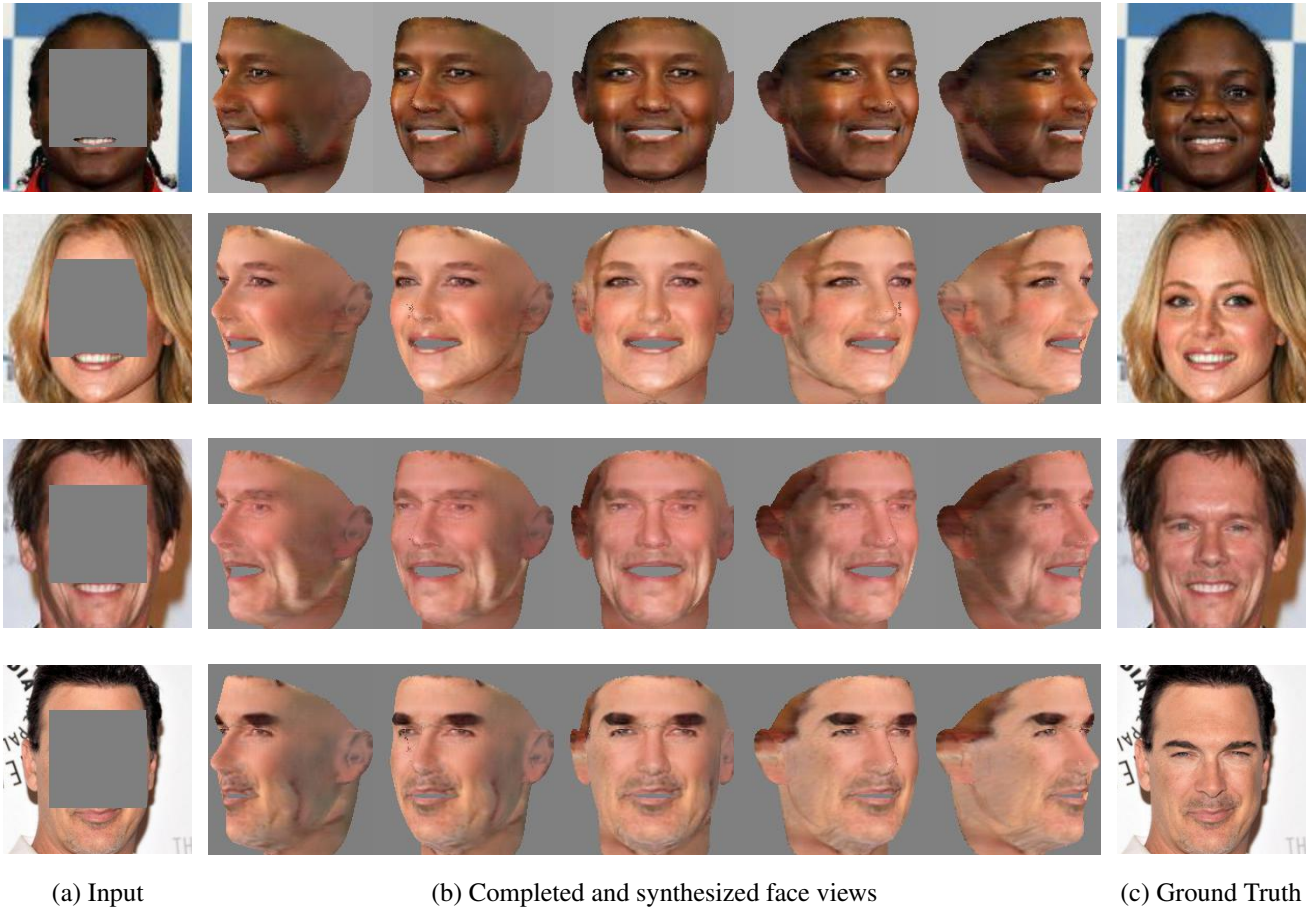
\begin{figure*}
    \centering
    \begin{tikzpicture}
        \node[] (a0) {\includegraphics[width=0.30\columnwidth,trim={20 20 20 20},clip]{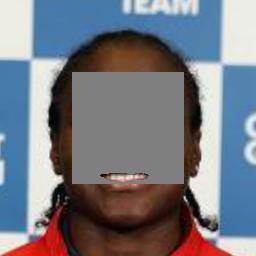}};
        \node[right of=a0, node distance=2.8cm] (a1) {\includegraphics[width=0.32\columnwidth,trim={10 20 20 20},clip]{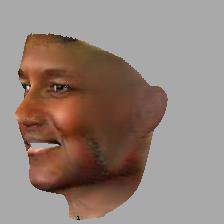}};
        \node[right of=a1, node distance=2.2cm] (a2) {\includegraphics[width=0.32\columnwidth,trim={15 20 15 20},clip]{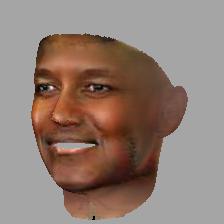}};
        \node[right of=a2, node distance=2.4cm] (a3) {\includegraphics[width=0.32\columnwidth,trim={15 20 15 20},clip]{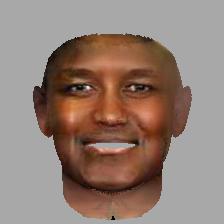}};
        \node[right of=a3, node distance=2.4cm] (a4) {\includegraphics[width=0.32\columnwidth,trim={15 20 15 20},clip]{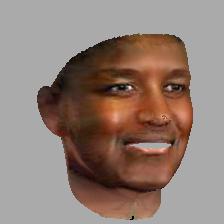}};
        \begin{scope}[on background layer]
            \node[right of=a4, node distance=2.2cm] (a5) {\includegraphics[width=0.32\columnwidth,trim={20 20 10 20},clip]{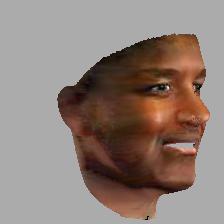}};
        \end{scope}
        \node[right of=a5, node distance=2.8cm] (a6) {\includegraphics[width=0.30\columnwidth,trim={20 20 20 20},clip]{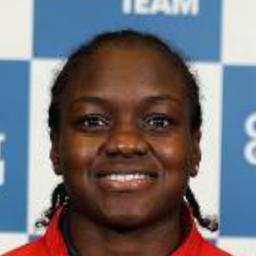}};
        
        \node[below of=a0, node distance=2.8cm] (c0) {\includegraphics[width=0.30\columnwidth,trim={20 20 20 20},clip]{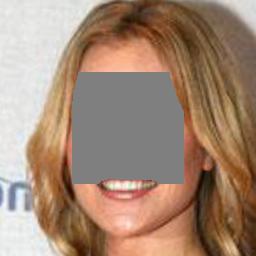}};
        \node[right of=c0, node distance=2.8cm] (c1) {\includegraphics[width=0.32\columnwidth,trim={10 20 20 20},clip]{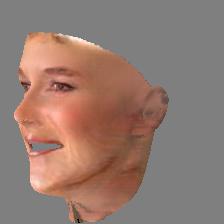}};
        \node[right of=c1, node distance=2.2cm] (c2) {\includegraphics[width=0.32\columnwidth,trim={15 20 15 20},clip]{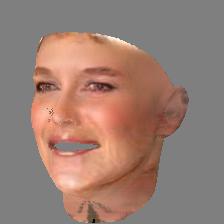}};
        \node[right of=c2, node distance=2.4cm] (c3) {\includegraphics[width=0.32\columnwidth,trim={15 20 15 20},clip]{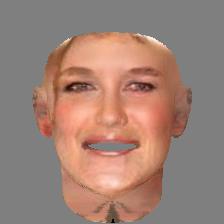}};
        \node[right of=c3, node distance=2.4cm] (c4) {\includegraphics[width=0.32\columnwidth,trim={15 20 15 20},clip]{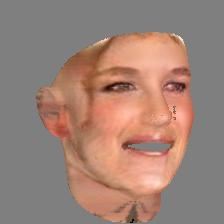}};
        \begin{scope}[on background layer]
            \node[right of=c4, node distance=2.2cm] (c5) {\includegraphics[width=0.32\columnwidth,trim={20 20 10 20},clip]{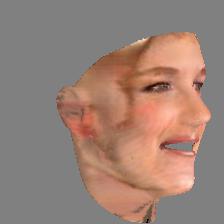}};
        \end{scope}
        \node[right of=c5, node distance=2.8cm] (c6) {\includegraphics[width=0.30\columnwidth,trim={20 20 20 20},clip]{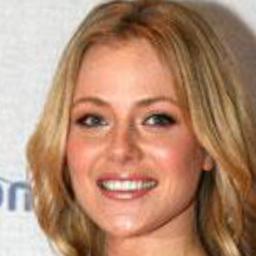}};
        
        \node[below of=c0, node distance=3.0cm] (e0) {\includegraphics[width=0.30\columnwidth,trim={20 20 20 20},clip]{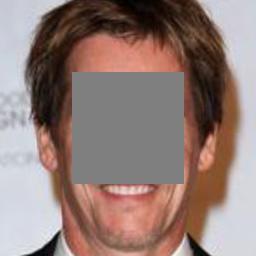}};
        \node[right of=e0, node distance=2.8cm] (e1) {\includegraphics[width=0.32\columnwidth,trim={10 20 20 20},clip]{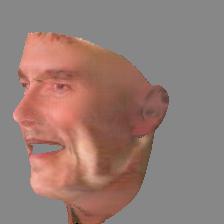}};
        \node[right of=e1, node distance=2.2cm] (e2) {\includegraphics[width=0.32\columnwidth,trim={15 20 15 20},clip]{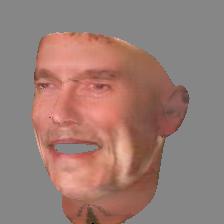}};
        \node[right of=e2, node distance=2.4cm] (e3) {\includegraphics[width=0.32\columnwidth,trim={15 20 15 20},clip]{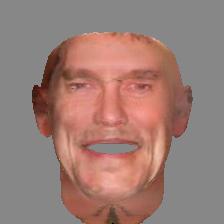}};
        \node[right of=e3, node distance=2.4cm] (e4) {\includegraphics[width=0.32\columnwidth,trim={15 20 15 20},clip]{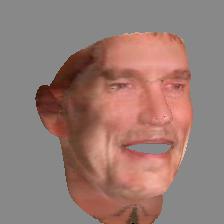}};
        \begin{scope}[on background layer]
            \node[right of=e4, node distance=2.2cm] (e5) {\includegraphics[width=0.32\columnwidth,trim={20 20 10 20},clip]{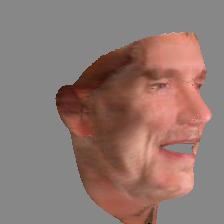}};
        \end{scope}
        \node[right of=e5, node distance=2.8cm] (e6) {\includegraphics[width=0.30\columnwidth,trim={20 20 20 20},clip]{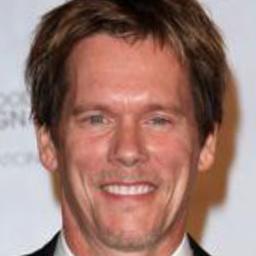}};
        
        \node[below of=e0, node distance=3.0cm] (f0) {\includegraphics[width=0.30\columnwidth,trim={20 20 20 20},clip]{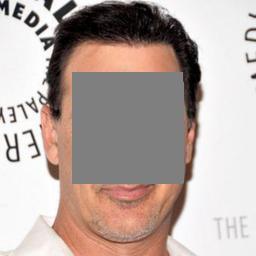}};
        \node[right of=f0, node distance=2.8cm] (f1) {\includegraphics[width=0.32\columnwidth,trim={10 20 20 20},clip]{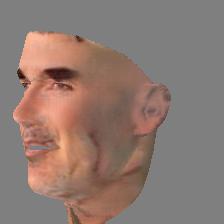}};
        \node[right of=f1, node distance=2.2cm] (f2) {\includegraphics[width=0.32\columnwidth,trim={15 20 15 20},clip]{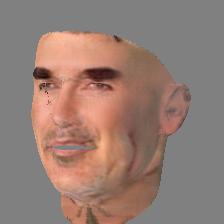}};
        \node[right of=f2, node distance=2.4cm] (f3) {\includegraphics[width=0.32\columnwidth,trim={15 20 15 20},clip]{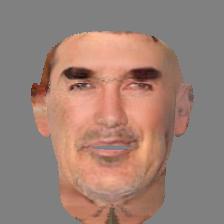}};
        \node[right of=f3, node distance=2.4cm] (f4) {\includegraphics[width=0.32\columnwidth,trim={15 20 15 20},clip]{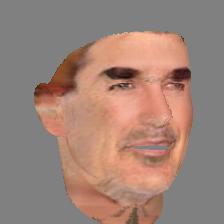}};
        \begin{scope}[on background layer]
            \node[right of=f4, node distance=2.2cm] (f5) {\includegraphics[width=0.32\columnwidth,trim={20 20 10 20},clip]{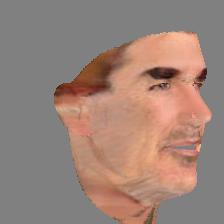}};
        \end{scope}
        \node[right of=f5, node distance=2.8cm] (f6) {\includegraphics[width=0.30\columnwidth,trim={20 20 20 20},clip]{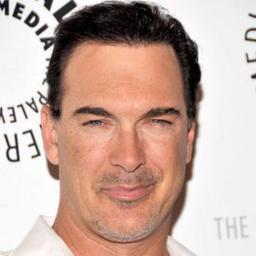}};
        
        \node[below of=f0, node distance=1.8cm] (h0) {(a) Input};
        \node[below of=f3, node distance=1.8cm] (h3) {(b) Completed and synthesized face views};
        \node[below of=f6, node distance=1.8cm] (h6) {(c) Ground Truth};
    \end{tikzpicture}
    \caption{\textbf{3D Face View Synthesis.} \ourmethod{} has the unique ability to not just complete masked faces realistically, but also synthesize new views from them.}
    \label{fig:multiviews}
\end{figure*}

\ourmethod{} has a unique advantage over other face completion approaches, in that unlike existing methods, our method can not only complete partial faces, but also render new views of the completed face from different view-points. In Fig.~\ref{fig:multiviews}, we show this through examples of face views rendered from five different viewpoints by completing the missing albedo and self-occluded regions in the masked faces.

\subsection{Evaluation in terms of SSIM}\label{subsec:quant}
\begin{figure*}
    \centering
    \begin{subfigure}{.33\textwidth}
        \centering
        \includegraphics[width=\linewidth,trim={0 10 60 70},clip]{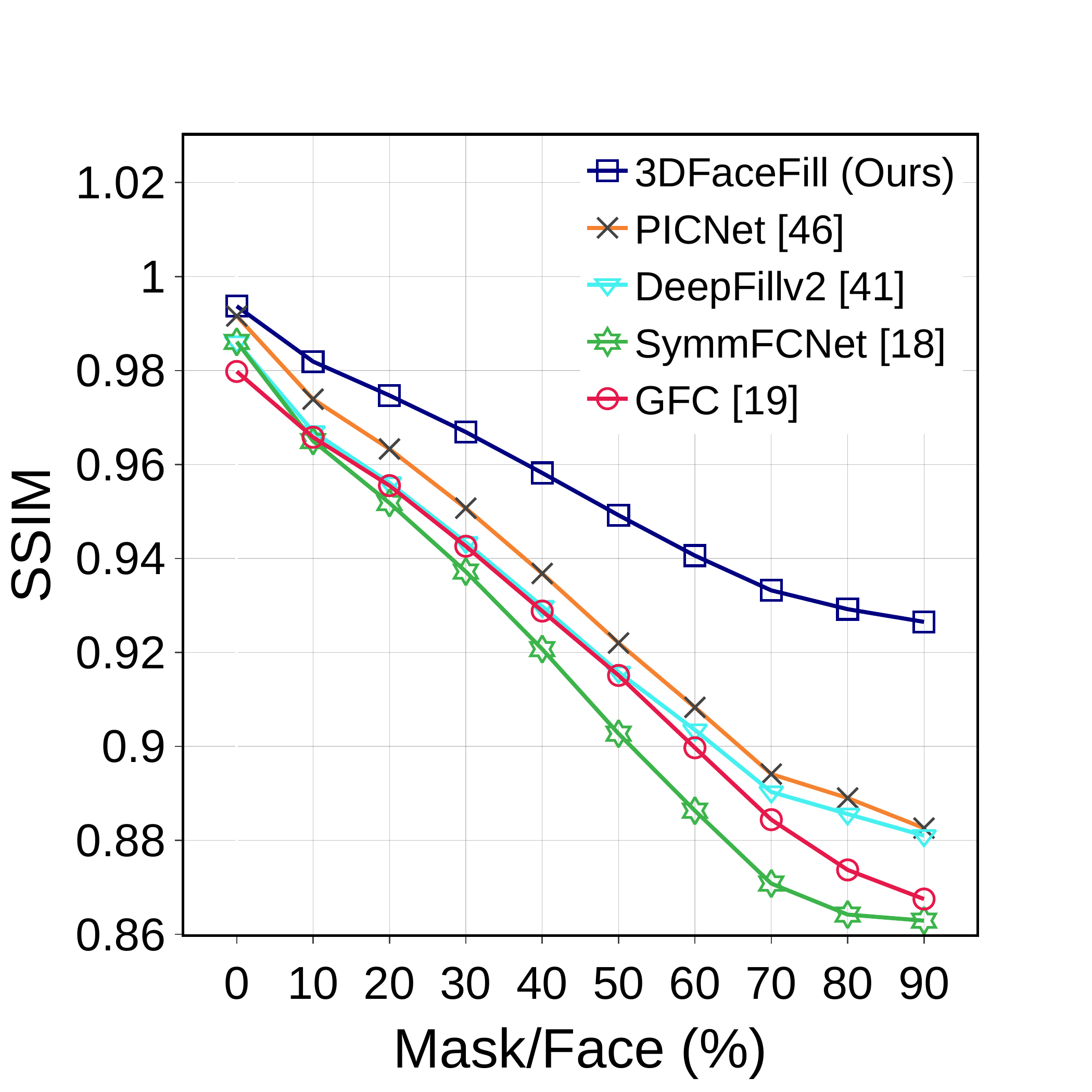}
        \caption{CelebA dataset \cite{celeba}}
        \label{fig:ssim_celeba}
    \end{subfigure}
    \begin{subfigure}{.33\textwidth}
        \centering
        \includegraphics[width=\linewidth,trim={0 10 60 70},clip]{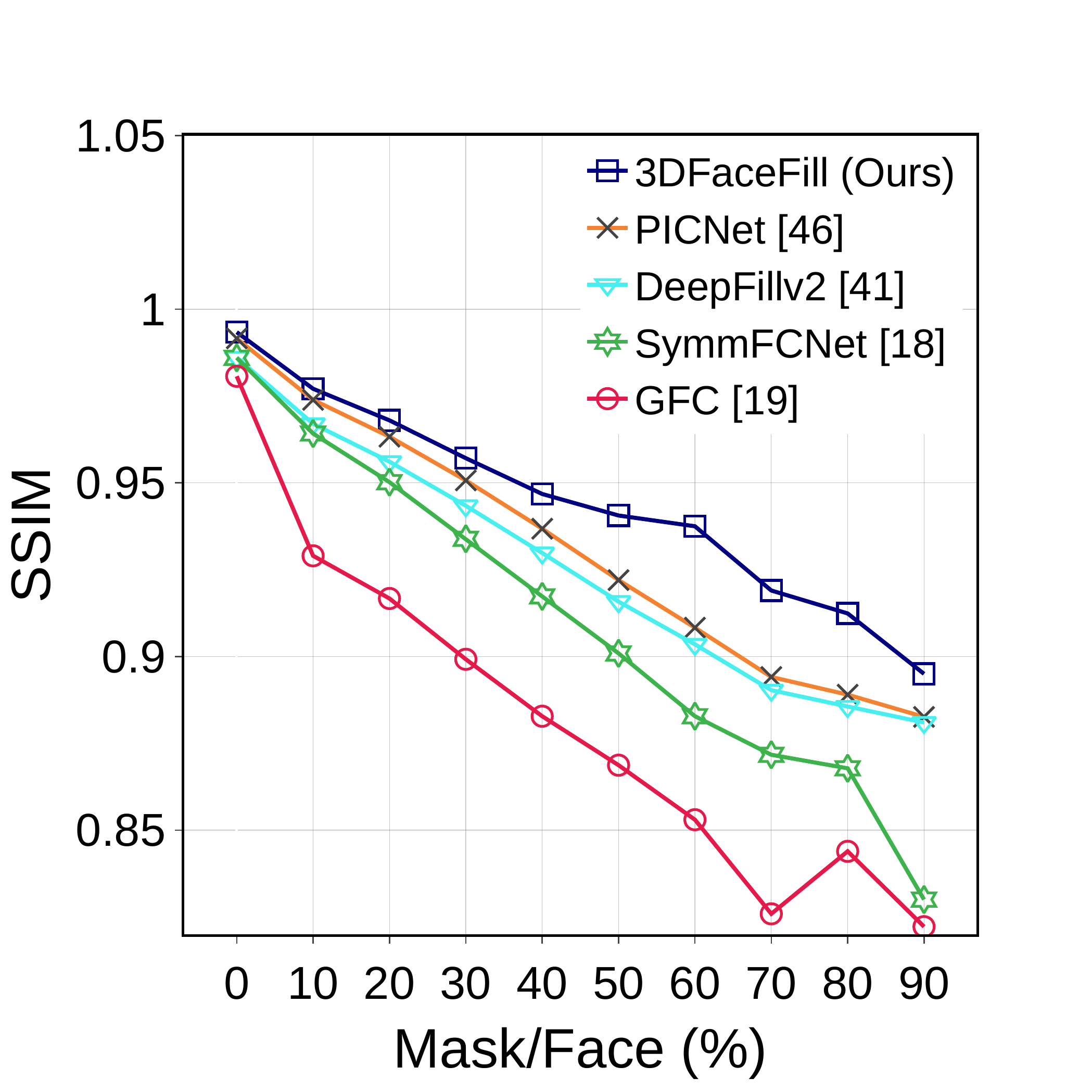}
        \caption{CelebA-HQ dataset \cite{CelebAMask-HQ}}
        \label{fig:ssim_celebahq}
    \end{subfigure}
    \begin{subfigure}{.33\textwidth}
        \centering
        \includegraphics[width=\linewidth,trim={0 10 60 70},clip]{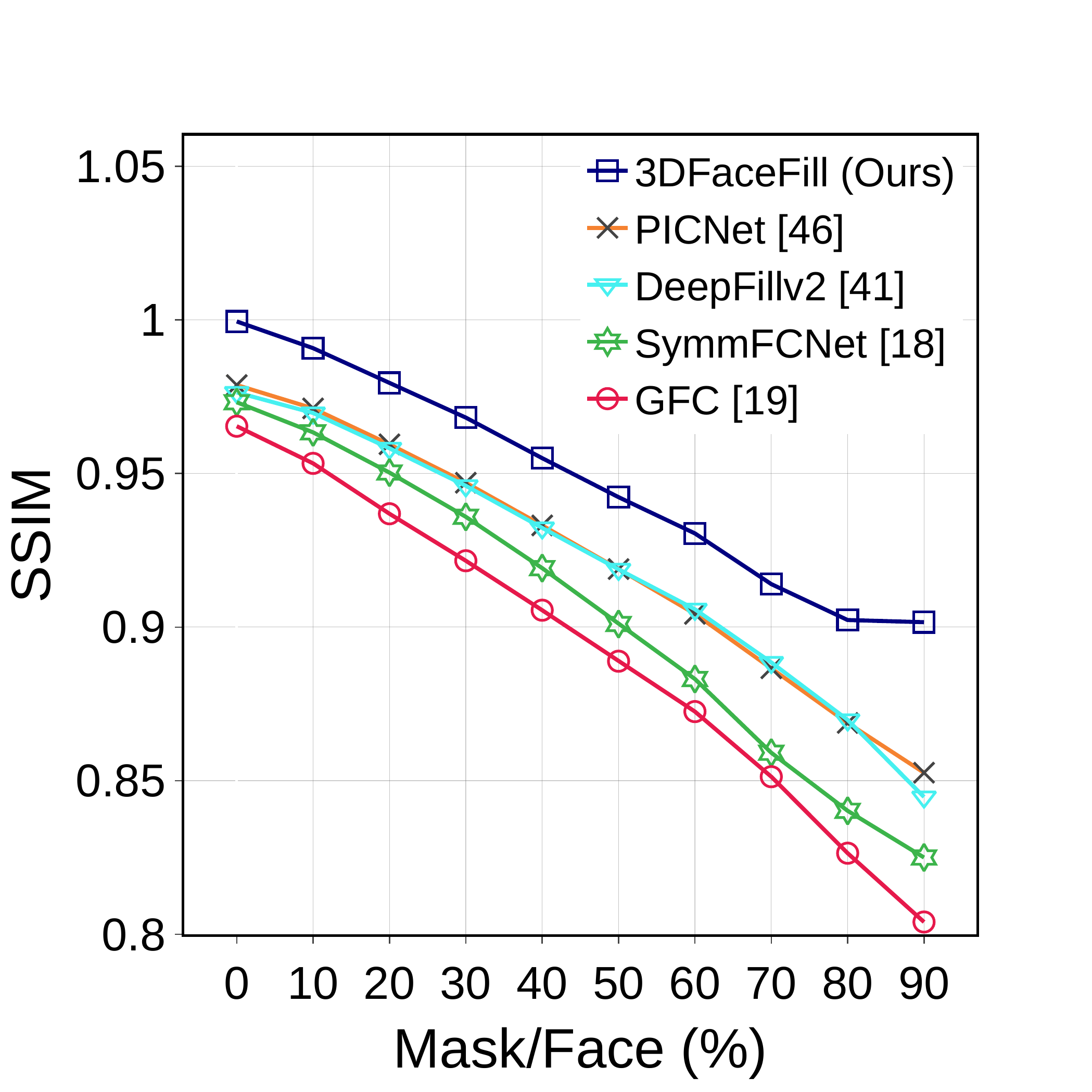}
        \caption{MultiPIE dataset \cite{multipie}}
        \label{fig:ssim_multipie}
    \end{subfigure}
    \caption{Quantitative evaluation of \ourmethod{} \textit{vs.} the baselines in terms of SSIM. \ourmethod{} consistently and significantly outperforms the baselines across all the Mask-to-Face area ratios and across all the datasets \textit{viz.} CelebA \cite{celeba}, CelebA-HQ \cite{CelebAMask-HQ} and MultiPIE \cite{multipie}.}
    \label{fig:ssim}
\end{figure*}

In addition to the PSNR/LPIPS \versus{} mask ratio analysis reported in the main paper, we also report similar comparison in terms of SSIM \versus{} mask ratio on the CelebA \cite{celeba}, CelebA-HQ \cite{CelebAMask-HQ} and MultiPIE \cite{multipie} datasets in Fig.~\ref{fig:ssim_celeba}, Fig.~\ref{fig:ssim_celebahq} and Fig.~\ref{fig:ssim_multipie}, respectively. \ourmethod{} consistently out-performs the baselines in terms of SSIM too for all the mask ratios. Moreover, it can be observed that the comparative gain by \ourmethod{} \versus{} the closest baseline increases as the mask ratio increases, which we attribute to the advantage of using explicit 3D priors for completion.

\section{Analysis}\label{sec:analysis}

\begin{figure*}
    \centering
    \captionsetup[sub]{font=footnotesize, labelformat = empty}
    \begin{subfigure}{\linewidth}
        \centering
        \includegraphics[width=.134\textwidth,trim={20 20 20 20},clip]{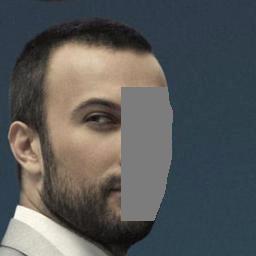}
        \includegraphics[width=.134\textwidth,trim={20 20 20 20},clip]{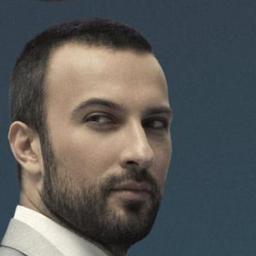}
        \includegraphics[width=.134\textwidth,trim={20 20 20 20},clip]{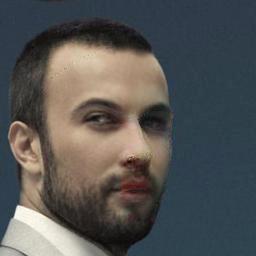}
        \includegraphics[width=.134\textwidth,trim={20 20 20 20},clip]{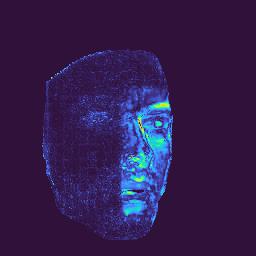}
        \includegraphics[width=.134\textwidth,trim={20 20 20 20},clip]{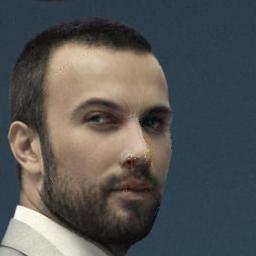}
        \includegraphics[width=.134\textwidth,trim={20 20 20 20},clip]{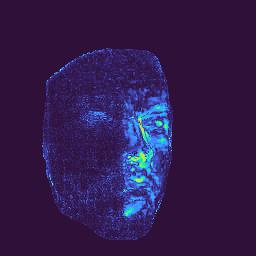}
        \includegraphics[width=.134\textwidth,trim={20 20 20 20},clip]{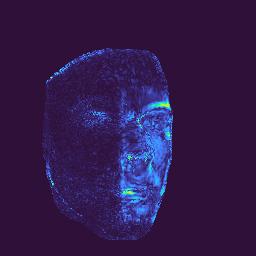}
    \end{subfigure}
    
    \begin{subfigure}{.134\textwidth}
        \centering
        \includegraphics[width=\textwidth,,trim={20 20 20 20},clip]{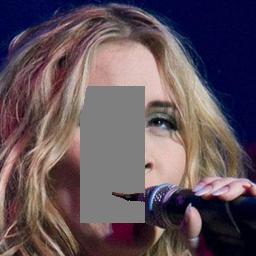}
        \caption{Input}
    \end{subfigure}
    \begin{subfigure}{.134\textwidth}
        \centering
        \includegraphics[width=\textwidth,,trim={20 20 20 20},clip]{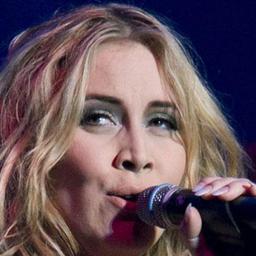}
        \caption{Ground Truth}
    \end{subfigure}
    \begin{subfigure}{.134\textwidth}
        \centering
        \includegraphics[width=\textwidth,,trim={20 20 20 20},clip]{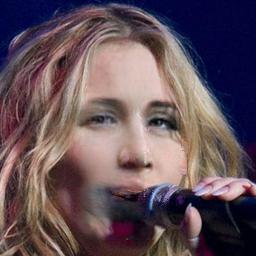}
        \caption{Iter1}
    \end{subfigure}
    \begin{subfigure}{.134\textwidth}
        \centering
        \includegraphics[width=\textwidth,,trim={20 20 20 20},clip]{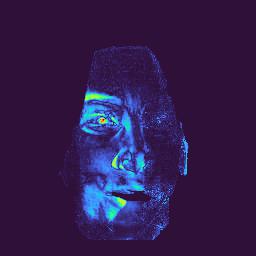}
        \caption{Iter1 - GT}
    \end{subfigure}
    \begin{subfigure}{.134\textwidth}
        \centering
        \includegraphics[width=\textwidth,,trim={20 20 20 20},clip]{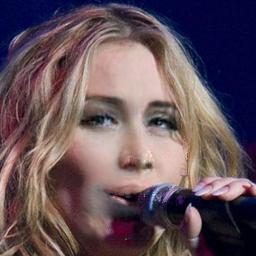}
        \caption{Iter2}
    \end{subfigure}
    \begin{subfigure}{.134\textwidth}
        \centering
        \includegraphics[width=\textwidth,,trim={20 20 20 20},clip]{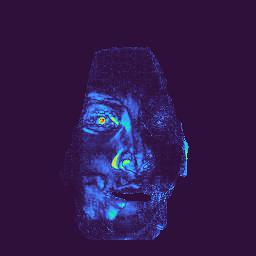}
        \caption{Iter2 - GT}
    \end{subfigure}
    \begin{subfigure}{.134\textwidth}
        \centering
        \includegraphics[width=\textwidth,,trim={20 20 20 20},clip]{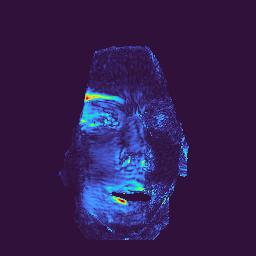}
        \caption{Iter2 - Iter1}
    \end{subfigure}
    \vspace{-2mm}
    \caption{\textbf{Effect of Iterative Finetuning}. We show raw completions (without blending) at iterations 1 and 2 along with the difference heatmaps. Note the improvements in \textit{Iter2} over \textit{Iter1} and the corresponding heatmap activations around eyes, eye-brows and other edges on the face.}
    \label{fig:ablation_iter}
\end{figure*}

\begin{figure*}
    \centering
    \captionsetup[sub]{font=footnotesize, labelformat = empty}
    \begin{subfigure}{\linewidth}
        \centering
        \includegraphics[width=.15\textwidth,,trim={20 20 20 20},clip]{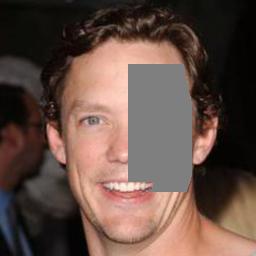}\hspace{6pt}
        \includegraphics[width=.15\textwidth,,trim={20 20 20 20},clip]{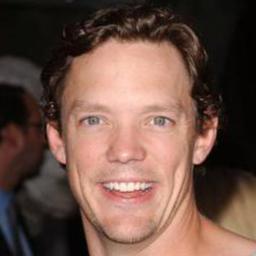}\hspace{6pt}
        \includegraphics[width=.15\textwidth,,trim={20 20 20 20},clip]{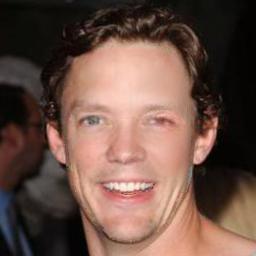}\hspace{6pt}
        \includegraphics[width=.15\textwidth,,trim={20 20 20 20},clip]{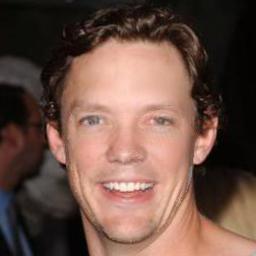}\hspace{6pt}
        \includegraphics[width=.15\textwidth,,trim={20 20 20 20},clip]{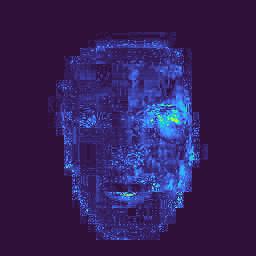}
    \end{subfigure}\vspace{6pt}
    
    \begin{subfigure}{.15\linewidth}
        \centering
        \includegraphics[width=\textwidth,,trim={20 20 20 20},clip]{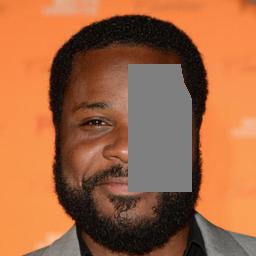}
        \caption{Input}
    \end{subfigure}\hspace{6pt}
    \begin{subfigure}{.15\linewidth}
        \centering
        \includegraphics[width=\textwidth,,trim={20 20 20 20},clip]{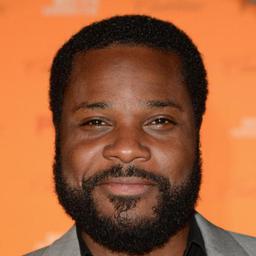}
        \caption{Ground Truth}
    \end{subfigure}\hspace{6pt}
    \begin{subfigure}{.15\linewidth}
        \centering
        \includegraphics[width=\textwidth,,trim={20 20 20 20},clip]{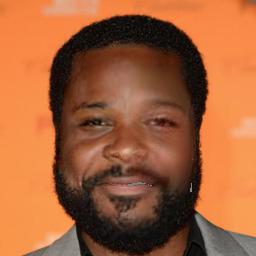}
        \caption{NoSym Model}
    \end{subfigure}\hspace{6pt}
    \begin{subfigure}{.15\linewidth}
        \centering
        \includegraphics[width=\textwidth,,trim={20 20 20 20},clip]{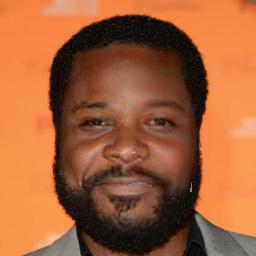}
        \caption{Full Model}
    \end{subfigure}\hspace{6pt}
    \begin{subfigure}{.15\linewidth}
        \centering
        \includegraphics[width=\textwidth,,trim={20 20 20 20},clip]{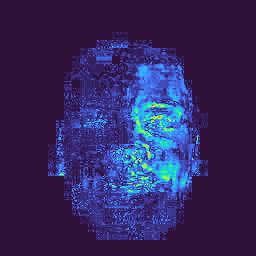}
        \caption{Full-NoSym}
    \end{subfigure}
    \vspace{-2mm}
    \caption{\textbf{Effect of using Symmetry}. The full model includes Sym-UNet and symmetry loss (during training) and can copy symmetric features when available. The absolute difference heatmaps (Full-NoSym) shows that most difference is coming from components such as eyes, eye-brows, \etc \vspace{-5pt}}
    \label{fig:ablation_symmetry}
\end{figure*}

\begin{figure*}
    \centering
    \begin{subfigure}{0.14\linewidth}
        \includegraphics[width=\textwidth]{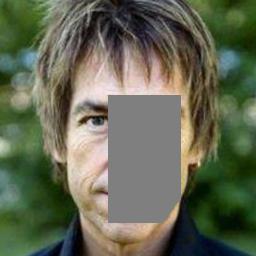}
    \end{subfigure}\hspace{6pt}
    \begin{subfigure}{0.15\linewidth}
        \includegraphics[width=\textwidth,trim={0.84cm 0 0.84cm 0},clip]{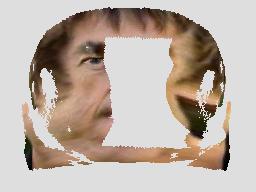}
    \end{subfigure}\hspace{6pt}
    \begin{subfigure}{0.15\linewidth}
        \includegraphics[width=\textwidth,trim={0.42cm 0 0.42cm 0},clip]{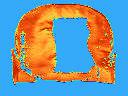}
    \end{subfigure}\hspace{6pt}
    \begin{subfigure}{0.15\linewidth}
        \includegraphics[width=\textwidth,trim={0.21cm 0 0.21cm 0},clip]{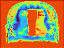}
    \end{subfigure}\hspace{6pt}
    \begin{subfigure}{0.15\linewidth}
        \includegraphics[width=\textwidth,trim={0.84cm 0 0.84cm 0},clip]{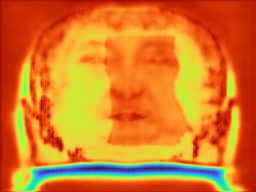}
    \end{subfigure}\hspace{6pt}
    \begin{subfigure}{0.15\linewidth}
        \includegraphics[width=\textwidth,trim={0.84cm 0 0.84cm 0},clip]{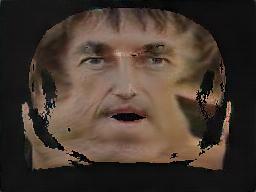}
    \end{subfigure}
    \vspace{6pt}
    
    \begin{subfigure}{0.14\linewidth}
        \includegraphics[width=\textwidth]{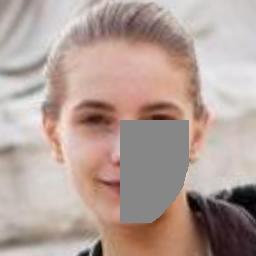}
    \end{subfigure}\hspace{6pt}
    \begin{subfigure}{0.15\linewidth}
        \includegraphics[width=\textwidth,trim={0.84cm 0 0.84cm 0},clip]{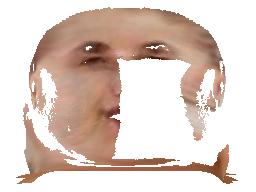}
    \end{subfigure}\hspace{6pt}
    \begin{subfigure}{0.15\linewidth}
        \includegraphics[width=\textwidth,trim={0.42cm 0 0.42cm 0},clip]{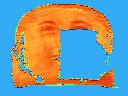}
    \end{subfigure}\hspace{6pt}
    \begin{subfigure}{0.15\linewidth}
        \includegraphics[width=\textwidth,trim={0.21cm 0 0.21cm 0},clip]{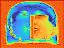}
    \end{subfigure}\hspace{6pt}
    \begin{subfigure}{0.15\linewidth}
        \includegraphics[width=\textwidth,trim={0.84cm 0 0.84cm 0},clip]{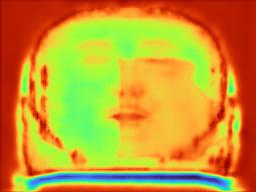}
    \end{subfigure}\hspace{6pt}
    \begin{subfigure}{0.15\linewidth}
        \includegraphics[width=\textwidth,trim={0.84cm 0 0.84cm 0},clip]{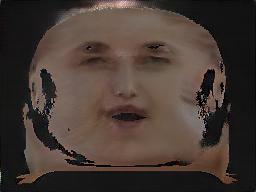}
    \end{subfigure}
    \vspace{6pt}
    
    \begin{subfigure}{0.14\linewidth}
        \includegraphics[width=\textwidth]{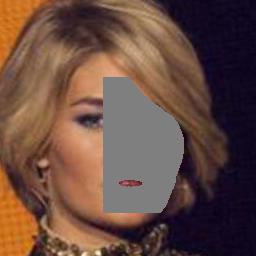}
        \caption{Input}
    \end{subfigure}\hspace{6pt}
    \begin{subfigure}{0.15\linewidth}
        \includegraphics[width=\textwidth,trim={0.84cm 0 0.84cm 0},clip]{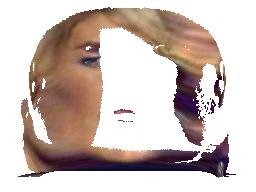}
        \caption{Input Albedo}
    \end{subfigure}\hspace{6pt}
    \begin{subfigure}{0.15\linewidth}
        \includegraphics[width=\textwidth,trim={0.42cm 0 0.42cm 0},clip]{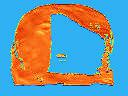}
        \caption{Gate 1}
    \end{subfigure}\hspace{6pt}
    \begin{subfigure}{0.15\linewidth}
        \includegraphics[width=\textwidth,trim={0.21cm 0 0.21cm 0},clip]{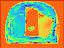}
        \caption{Gate 2}
    \end{subfigure}\hspace{6pt}
    \begin{subfigure}{0.15\linewidth}
        \includegraphics[width=\textwidth,trim={0.84cm 0 0.84cm 0},clip]{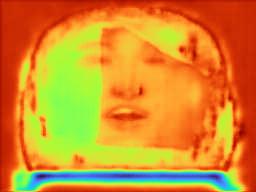}
        \caption{Uncertainty $\sigma$}
    \end{subfigure}\hspace{6pt}
    \begin{subfigure}{0.15\linewidth}
        \includegraphics[width=\textwidth,trim={0.84cm 0 0.84cm 0},clip]{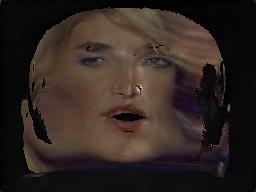}
        \caption{Output Albedo}
    \end{subfigure}
    
    \caption{\textbf{Visualizing the Gating Activations and the Uncertainty-Maps}. Observe that, while \textit{Gate 1} activates for the visible regions, \textit{Gate 2} activates for the masked regions to propagate useful features from the visible symmetric parts to their masked counterparts. The uncertainty map captures the model's uncertainty around the masked regions and the facial components such as the eyes, thus incurring higher losses for these  regions. (\textit{Note:} higher values are represented by warmer (redish) colors in the gating and uncertainty heatmaps).}
    \label{fig:gating}
\end{figure*}

\subsection{Iterative Refinement}\label{subsec:iterative}
As explained in Main:Sec \textcolor{red}{3.2}, we adopt an iterative refinement procedure whereby 3D factorization aids in completion and vice versa. We show heatmap-visualization of the difference between the pre-blended completions at iterations 1 and 2 in Fig. \ref{fig:ablation_iter}. A relative comparison of the distribution of Iter1-GT and Iter2-GT shows that Iter2 is closer to the GT (ground truth) image, which indicates improved 3D pose estimation in the second iteration. The visualization of Iter2-Iter1 shows that these differences are manifesting at the detailed face components such as eyes, nose, \etc., as well as the masked regions.

\subsection{Effect of Symmetry}\label{subsec:symmetry}
We further analyze the advantages of enforcing symmetry-consistency. We show completions by the Full model (Sym-UNet + symmetry loss) and the NoSym model (UNet) in Fig. \ref{fig:ablation_symmetry}. The completions by the NoSym variant look slightly asymmetric because of blurry and symmetry-agnostic completion around the eyes and other masked regions whereas our Full model leverages symmetry to generate sharper and symmetry-consistent completions. This can be further visualized in the Full-NoSym difference heatmaps shown in the figure that show that the difference is mostly concentrated around the eyes and eye-brows and spreads further through the masked regions.

\vspace{2pt}
\noindent\textbf{Symmetry Gating Activation:} We further visualize the intermediate gating maps used in our model that control the flow of information in the network (ref Fig.~\ref{fig:gating}). We visualize two (out of 64) gating activations (1st - Gate1 and 33rd - Gate2) from the second layer of our Sym-UNet network. As can be seen in Fig.~\ref{fig:gating}, while Gate1 activates for the visible regions in the input albedo, Gate2 activates for the masked regions to propagate useful features from the horizontally flipped albedo map to the symmetric side. This enables Sym-UNet to leverage and maintain facial symmetry for inpainting. We also visualize the estimated uncertainty map ($\sigma$) in Fig.~\ref{fig:gating} that is learned by the inpainter $\mathcal{G}$ in an unsupervised way. Note that the uncertainty is usually higher around important facial components like the eyes and the masked regions, which increases the loss incurred in these regions.
\section{Implementation Details}\label{sec:impl}
In this section, we provide further implementation details regarding the proposed approach. In sub-section \ref{subsec:network_arch}, we give detailed network architectures for the modules used in \ourmethod{}. In sub-section \ref{subsec:3dmm}, we provide details of the loss functions used to train the 3D factorization module. Lastly, we give full training details of the different components in sub-section \ref{subsec:training}.

\subsection{Network Architectures}\label{subsec:network_arch}
We report the detailed network architectures for the 3DMM Encoder $\mathcal{E}$, the Albedo Decoder $\mathcal{D}_A$, the Sym-UNet module, the PyramidGAN discriminator and the Face Segmenter $\mathcal{S}$ in Tables \ref{tab:3dmm_enc} to \ref{tab:face_seg}. Our network architectures for the 3DMM modules are based on the architectures used in \cite{tran2019learning} for the corresponding modules. Insipired by Miyato \etal \cite{spectralnorm}, we use spectral normalization in all our convolution layers. The abbreviated operators used are defined as follows:
\begin{itemize}
    \item Conv$(c_{in}, c_{out}, k, s, p)$: 2D convolution with $c_{in}$ input channels, $c_{out}$ output channels, kernel size $k$, stride $s$ and padding $p$.
    \item Deconv$(c_{in}, c_{out}, k, s, p)$: 2D transposed convolution (deconvolution) with $c_{in}$ input channels, $c_{out}$ output channels, kernel size $k$, stride $s$ and padding $p$.
    \item GN$(n)$: Group normalization \cite{groupnorm} with $n$ groups
    \item ELU: Exponential linear unit \cite{elu} activation, LReLU$(\alpha)$: Leaky ReLU \cite{lrelu} with a negative slope of $\alpha$
    \item ResUnit$(c_{in}, c_{out}, k, s, p)$: Residual unit \cite{resnet} with $c_{in}$ input channels, $c_{out}$ output channels, kernel size $k$, stride $s$, padding $p$ with group normalization \cite{groupnorm} and ELU activation \cite{lrelu}
    \item SigGNConv$(c_{in}, c_{out}, k, s, p)$: 2D convolution with $c_{in}$ input channels, $c_{out}$ output channels, kernel size $k$, stride $s$ and padding $p$ followed by group normalization \cite{groupnorm} and sigmoid activation
    \item SigGNDeconv$(c_{in}, c_{out}, k, s, p)$: 2D transposed convolution with $c_{in}$ input channels, $c_{out}$ output channels, kernel size $k$, stride $s$ and padding $p$ followed by group normalization \cite{groupnorm} and sigmoid activation
    \item SpectralConv$(c_{in}, c_{out}, k, s, p)$: 2D convolution with $c_{in}$ input channels, $c_{out}$ output channels, kernel size $k$, stride $s$, padding $p$ and spectral normalization \cite{spectralnorm}
    \item Upsample$(s_h, s_c)$: Upsamples height by $s_h$ and width by $s_w$ using nearest neighbour interpolation.
\end{itemize}
\begin{table}[]
    \caption{Network architecture of the 3DMM Encoder $\mathcal{E}$. The $Pose_{1}$ corresponds to the scale, $Pose_{2:4}$ correspond to the yaw, roll and pitch angles normalized by $\pi/2$ and $Pose_{5:6}$ correspond to the X and Y translations normalized by the input image size.}
    \label{tab:3dmm_enc}
    \resizebox{\columnwidth}{!}{%
    \begin{tabular}{lc}
         \thickhline
         3DMM Encoder & Output size\\
         \hline
         \textit{Image} $\rightarrow$ SpectralConv(3, 32, 7, 2, 3) + GN(8) + ELU & 112x112\\
         SpectralConv(32, 64, 3, 1, 1) + GN(16) + ELU & 112x112\\
         SpectralConv(64, 64, 3, 2, 1) + GN(16) + ELU & 56x56\\
         SpectralConv(64, 96, 3, 1, 1) + GN(24) + ELU & 56x56\\
         SpectralConv(96, 128, 3, 1, 1) + GN(32) + ELU & 56x56\\
         SpectralConv(128, 128, 3, 2, 1) + GN(32) + ELU & 28x28\\
         SpectralConv(128, 196, 3, 1, 1) + GN(48) + ELU & 28x28\\
         SpectralConv(196, 256, 3, 1, 1) + GN(64) + ELU & 28x28\\
         SpectralConv(256, 256, 3, 2, 1) + GN(64) + ELU & 14x14\\
         SpectralConv(256, 256, 3, 1, 1) + GN(64) + ELU & 14x14\\
         SpectralConv(256, 256, 3, 1, 1) + GN(64) + ELU & 14x14\\
         SpectralConv(256, 512, 3, 2, 1) + GN(128) + ELU & 7x7\\
         SpectralConv(512, 512, 3, 1, 1) + GN(128) + ELU $\rightarrow$ \textit{feats} & 7x7\\
         & \\
         \textit{feats} $\rightarrow$ SpectralConv(512, 160, 3, 1, 1) + GN(40) + ELU & 7x7\\
         AvgPool(7,7) & 1x1\\
         Linear(160, 6) + Tanh $\rightarrow$ \textit{Pose}\\
         & \\
         \textit{feats} $\rightarrow$ SpectralConv(512, 160, 3, 1, 1) + GN(40) + ELU & 7x7\\
         AvgPool(7,7) & 1x1\\
         Linear(160, 27) $\rightarrow$ \textit{Illumination}\\
         & \\
         \textit{feats} $\rightarrow$ SpectralConv(512, 512, 3, 1, 1) + GN(128) + ELU & 7x7\\
         SpectralConv(512, 512, 3, 1, 1) + GN(128) + ELU & 7x7\\
         AvgPool(7,7) & 1x1\\
         Linear(512, 199+29) $\rightarrow$ \textit{199 Shape + 29 Expression coefficients}\\
         & \\
         \textit{feats} $\rightarrow$ SpectralConv(512, 512, 3, 1, 1) + GN(128) + ELU & 7x7\\
         AvgPool(7,7) $\rightarrow$ \textit{Albedo features} & 1x1\\
         \hline
         \textbf{Model Complexity} & 17.4M\\
         \thickhline
    \end{tabular}%
    }
\end{table}

\begin{table}[]
    \caption{Network architecture of the Albedo Decoder $\mathcal{D}_A$ that decodes the 512 dimensional Albedo features from the 3DMM Encoder $\mathcal{E}$ into $3\times 192\times 256$ dimensional Albedo representation in the UV space.}
    \label{tab:alb_dec}
    \resizebox{\columnwidth}{!}{
    \begin{tabular}{lc}
         \thickhline
         Albedo Decoder & Output size\\
         \hline
         $\textit{Albedo features}$ $\rightarrow$ Upsample(3,4) & 3x4\\
         SpectralConv(512, 512, 3, 1, 1) + GN(128) + ELU & 3x4\\
         SpectralConv(512, 256, 3, 1, 1) + GN(64) + ELU & 3x4\\
         Upsample(2,2) & 6x8\\ 
         SpectralConv(256, 256, 3, 1, 1) + GN(64) + ELU & 6x8\\
         SpectralConv(256, 128, 3, 1, 1) + GN(32) + ELU & 6x8\\
         SpectralConv(128, 128, 3, 1, 1) + GN(32) + ELU & 6x8\\
         Upsample(2,2) & 12x16\\
         SpectralConv(128, 160, 3, 1, 1) + GN(40) + ELU & 12x16\\
         SpectralConv(160, 96, 3, 1, 1) + GN(32) + ELU & 12x16\\
         SpectralConv(96, 128, 3, 1, 1) + GN(32) + ELU & 12x16\\
         Upsample(2,2) & 24x32\\
         SpectralConv(128, 128, 3, 1, 1) + GN(32) + ELU & 24x32\\
         SpectralConv(128, 64, 3, 1, 1) + GN(16) + ELU & 24x32\\
         SpectralConv(64, 96, 3, 1, 1) + GN(24) + ELU & 24x32\\
         Upsample(2,2) & 48x64\\
         SpectralConv(96, 96, 3, 1, 1) + GN(32) + ELU & 48x64\\
         SpectralConv(96, 64, 3, 1, 1) + GN(16) + ELU & 48x64\\
         SpectralConv(64, 64, 3, 1, 1) + GN(16) + ELU & 48x64\\
         Upsample(2,2) & 96x128\\
         SpectralConv(64, 64, 3, 1, 1) + GN(16) + ELU & 96x128\\
         SpectralConv(64, 32, 3, 1, 1) + GN(8) + ELU & 96x128\\
         SpectralConv(32, 32, 3, 1, 1) + GN(8) + ELU & 96x128\\
         Upsample(2,2) & 192x256\\
         SpectralConv(32, 32, 3, 1, 1) + GN(8) + ELU & 192x256\\
         SpectralConv(32, 16, 3, 1, 1) + GN(4) + ELU & 192x256\\
         SpectralConv(16, 16, 3, 1, 1) + GN(4) + ELU & 192x256\\
         Conv(16, 3, 1, 1, 0) + Tanh $\rightarrow$ \textit{Albedo}\\
         \hline
         \textbf{Model Complexity} & 5.54M\\
         \thickhline
    \end{tabular}}
\end{table}

\begin{table}[]
    \caption{Network architecture of the Albedo Inpainter $\mathcal{G}$ (Sym-UNet). The input to the network is the concatenation of the masked Albedo $\mathbf{A}^{uv}_m$ and the mask $\mathbf{M}^{uv}$ in the UV space $X = (\mathbf{A}^{uv}_m, \mathbf{M}^{uv})$. Outputs are the completed Albedo $\hat{\mathbf{A}}^{uv}$ and the uncertainty map $\sigma^{uv}$.}
    \label{tab:alb_comp}
    \resizebox{\columnwidth}{!}{
    \begin{tabular}{clc}
         \thickhline
         Input & Layer & Output\\
         \hline
         $X$ & ResUnit(4, 32, 3, 2, 1) & $f1$\\
         $X$ & SigGNConv(4, 32, 3, 2, 1) & $g1$\\
         $hflip(X)$ & ResUnit(4, 32, 3, 2, 1) & $f1'$\\
         $hflip(X)$ & SigGNConv(4, 32, 3, 2, 1) & $g1'$\\
         $(f1\odot g1, f1'\odot g1')$ & ResUnit(64, 64, 3, 2, 1) & $f2$\\
         $(f1\odot g1, f1'\odot g1')$ & SigGNConv(64, 64, 3, 2, 1) & $g2$\\
         $f2\odot g2$ & ResUnit(64, 128, 3, 2, 1) & $f3$\\
         $f2\odot g2$ & SigGNConv(64, 128, 3, 2, 1) & $g3$\\
         $f3\odot g3$ & ResUnit(128, 256, 3, 2, 1) & $f4$\\
         $f3\odot g3$ & SigGNConv(128, 256, 3, 2, 1) & $g4$\\
         $f4\odot g4$ & ResUnit(256, 512, 3, 2, 1) & $f5$\\
         $f4\odot g4$ & SigGNConv(256, 512, 3, 2, 1) & $g5$\\
         $f5\odot g5$ & ResUnit(512, 256, 3, 1, 1) & $f5^1$\\
         $f5\odot g5$ & SigGNConv(512, 256, 3, 1, 1) & $g5^1$\\
         & \\
         $f5^1\odot g5^1$ & Upsample(2,2) & $x4$\\
         $(x4, f4\odot g4)$ & ResUnit(512, 128, 3, 1, 1) & $f4^1$\\
         $f5^1\odot g5^1$ & SigGNDeconv(256, 128, 4, 2, 1) & $g4^1$\\
         
         $f4^1\odot g4^1$ & Upsample(2,2) & $x3$\\
         $(x3, f3\odot g3)$ & ResUnit(256, 64, 3, 1, 1) & $f3^1$\\
         $f4^1\odot g4^1$ & SigGNDeconv(128, 64, 4, 2, 1) & $g3^1$\\
         
         $f3^1\odot g3^1$ & Upsample(2,2) & $x2$\\
         $(x2, f2\odot g2)$ & ResUnit(128, 64, 3, 1, 1) & $f2^1$\\
         $f3^1\odot g3^1$ & SigGNDeconv(128, 64, 4, 2, 1) & $g2^1$\\
         
         $f2^1\odot g2^1$ & Upsample(2,2) & $x1$\\
         $(x1, f1\odot g1)$ & ResUnit(128, 64, 3, 1, 1) & $f1^1$\\
         $f2^1\odot g2^1$ & SigGNDeconv(128, 64, 4, 2, 1) & $g1^1$\\
         
         $f1^1\odot g1^1$ & Upsample(2,2) & $x0$\\
         $x0$ & ResUnit(64, 32, 3, 1, 1) & $f0^1$\\
         $f1^1\odot g1^1$ & SigGNDeconv(64, 32, 4, 2, 1) & $g0^1$\\
         
         $f0^1\odot g0^1$ & Conv(32, 4, 1, 1, 0) & $(\hat{\mathbf{A}}^{uv}, \sigma^{uv})$\\
         \hline
         \multicolumn{2}{c}{\textbf{Model Complexity}} & 11.7M\\
         \thickhline
    \end{tabular}}
\end{table}

\begin{table}[]
    \caption{Network architecture of the PyramidGAN discriminator $\mathcal{D}$.}
    \label{tab:alb_disc}
    \resizebox{\columnwidth}{!}{
    \begin{tabular}{clc}
         \thickhline
         Input & Layer & Output\\
         \hline
         $\mathbf{I}_{gt}/\hat{\mathbf{I}}$ & SpectralConv(3, 32, 4, 2, 1) + GN(8) + LReLU(.2) & $x0$\\
         
         $x0$ & SpectralConv(32, 64, 4, 2, 1) + GN(16) + LReLU(.2) & $x1$\\
         $x1$ & SpectralConv(64, 1, 1, 1, 0) & $out1$\\
         
         $x1$ & SpectralConv(64, 128, 4, 2, 1) + GN(32) + LReLU(.2) & $x2$\\
         $x2$ & SpectralConv(128, 1, 1, 1, 0) & $out2$\\
         
         $x2$ & SpectralConv(128, 256, 4, 2, 1) + GN(64) + LReLU(.2) & $x3$\\
         $x3$ & SpectralConv(256, 1, 1, 1, 0) & $out3$\\
         
         $x3$ & SpectralConv(256, 512, 4, 2, 1) + GN(128) + LReLU(.2) & $x4$\\
         $x4$ & SpectralConv(512, 1, 1, 1, 0) & $out4$\\
         \hline
         \multicolumn{2}{c}{\textbf{Model Complexity}} & 2.79M\\
         \thickhline
    \end{tabular}}
\end{table}

\begin{table}[]
    \caption{Network architecture of the face segmenter $\mathcal{S}$. $(x, y)$ represents the concatenation of tensors $x$ and $y$ along the channel dimension. The output of the network consist of a face mask $\mathbf{M}_f$, an occlusion mask $\mathbf{M}_o$ and a background mask $\mathbf{M}_b$.}
    \label{tab:face_seg}
    \resizebox{\columnwidth}{!}{
    \begin{tabular}{clc}
         \thickhline
         Input & Layer & Output\\
         \hline
         \textit{Image} & ResUnit(3, 32, 3, 1, 1) & $x1$\\
         $x1$ & ResUnit(32, 64, 3, 2, 1) & $x2$\\
         $x2$ & ResUnit(64, 128, 3, 2, 1) & $x3$\\
         $x3$ & ResUnit(128, 256, 3, 2, 1) & $x4$\\
         $x4$ & ResUnit(256, 256, 3, 2, 1) & $x5$\\
         $x5$ & ResUnit(256, 256, 3, 2, 1) & $x6$\\
         &\\
         $x6$ & Upsample(2,2) & $x5^1$\\
         $(x5^1, x5)$ & ResUnit(512, 256, 3, 1, 1) & $x5^2$\\
         $x5^2$ & Upsample(2,2) & $x4^1$\\
         $(x4^1, x4)$ & ResUnit(512, 128, 3, 1, 1) & $x4^2$\\
         $x4^2$ & Upsample(2,2) & $x3^1$\\
         $(x3^1, x3)$ & ResUnit(256, 64, 3, 1, 1) & $x3^2$\\
         $x3^2$ & Upsample(2,2) & $x2^1$\\
         $(x2^1, x2)$ & ResUnit(128, 32, 3, 1, 1) & $x2^2$\\
         $x2^2$ & Upsample(2,2) & $x1^1$\\
         $(x1^1, x1)$ & ResUnit(64, 32, 3, 1, 1) & $x1^2$\\
         &\\
         $x1^2$ & Conv(32, 3, 1, 1, 0) + Softmax2d & $(\mathbf{M}_f, \mathbf{M}_o, \mathbf{M}_b)$\\
         \hline
         \multicolumn{2}{c}{\textbf{Model Complexity}} & 7.18M\\
         \thickhline
    \end{tabular}}
\end{table}

\subsection{3DMM Module Losses}\label{subsec:3dmm}
The 3DMM module is trained using a combination of supervised, reconstruction and regularization losses: 
\begin{equation}
    \mathcal{L}_{3DMM} = \lambda_{sup}\mathcal{L}_{sup} + \lambda_{rec}\mathcal{L}_{rec} + \lambda_{reg}\mathcal{L}_{reg},
\end{equation}
where, $\mathcal{L}_{sup}=\lambda_S \mathcal{L}(\mathbf{S}, \mathbf{\Tilde{S}}) + \lambda_p \mathcal{L}(\mathbf{p}, \mathbf{\Tilde{p}}) + \lambda_T \mathcal{L}(\mathbf{T}^{uv}, \mathbf{\Tilde{T}}^{uv}) + \lambda_{lmark}\mathcal{L}_{lmark}$ use the groundtruth shape, pose, texture and 2D landmarks when available, $\mathcal{L}_{rec}$ enforces similarity between the rendered and grountruth images and $\mathcal{L}_{reg}=\lambda_{3dsym}\mathcal{L}_{3dsym}+\lambda_{const}\mathcal{L}_{const}$ are regularization losses to enforce bilateral symmetry of albedo and effective separation of shade and albedo. All loss coefficients $\lambda$'s are set to have equal weightage for all the loss terms. We now define these losses:

\vspace{5pt}
\noindent \textit{- Shape loss is defined as:}
\begin{align*}
    \mathcal{L}(\mathbf{S}, \mathbf{\Tilde{S}}) = \mathbb{E}\left[ ||\mathbf{f}_S - \Tilde{\mathbf{f}}_S||_2^2 \right],
\end{align*}
where $\mathbf{f}_S$ and $\Tilde{\mathbf{f}}_S$ are the output and groundtruth shape and expression coefficients, respectively. 

\vspace{5pt}
\noindent \textit{- Pose loss is defined as a combination of scale, translation and rotation losses:}
\begin{align*}
    \mathcal{L}(\mathbf{p}, \mathbf{\Tilde{p}}) = \lambda_s \mathbb{E}\left[ (s-\Tilde{s})^2 \right] + \lambda_t \mathbb{E}\left[ ||\mathbf{t}_{x,y} - \Tilde{\mathbf{t}}_{x,y}||_2^2 \right] + \lambda_r\mathcal{L}_R,
\end{align*}
where $s$ represents scale, $\mathbf{t}_{x,y}$ represents the X and Y translations, and $\mathcal{L}_R = \mathbb{E}\left[ ||quat(\mathbf{R}) - quat(\Tilde{\mathbf{R}})||_2^2 \right]$ is the rotation loss with $\mathbf{R}$ representing the rotation along the X, Y and Z axes and $quat(.)$ gives its quaternion representation.

\vspace{5pt}
\noindent \textit{- Texture loss is defined as:}
\begin{align*}
    \mathcal{L}(\mathbf{T}^{uv}, \mathbf{\Tilde{T}}^{uv}) = \mathbb{E}\left[ ||\mathbf{T}^{uv} - \mathbf{\Tilde{T}}^{uv}||_2^2 \right],
\end{align*}
where $\mathbf{T}^{uv}$ is the texture represented in UV space.

\vspace{5pt} \noindent \textit{- Landmark loss is defined as:}
\begin{align*}
    \mathcal{L}_{lmark} = \left\lVert \mathbf{M(p)} * \left[ \begin{matrix}\mathbf{S(:,d)} \\ \mathbf{1} \end{matrix} \right] - \mathbf{U} \right \rVert_2^2,
\end{align*}
where $\mathbf{M}$ is the camera projection matrix obtained from the pose $\mathbf{p}$, $\mathbf{d}$ selects 68 indices corresponding to sparse 2D landmarks on the 3D face mesh $\mathbf{S}$ and $\mathbf{U}\in \mathbb{R}^{2\times 68}$ are the groundtruth locations of 2D facial landmarks.

\vspace{5pt} \noindent \textit{- Reconstruction is defined as:}
\begin{align*}
    \mathcal{L}_{rec} = \mathbb{E}\left[||\mathbf{I} - \mathbf{\hat{I}}||_1\right],
\end{align*}
where $\mathbf{I}$ and $\Tilde{\mathbf{I}}$ are the rendered and ground-truth images, respectively.

\vspace{5pt} \noindent \textit{- Albedo symmetry loss is defined as:}
\begin{align*}
    L_{3dsym}(\mathbf{A}) = \lVert\mathbf{A}^{uv} - hflip(\mathbf{A}^{uv}) \rVert_1,
\end{align*}
where $\mathbf{A}^{uv}$ is the UV representation of albedo and hflip() is the horizontal image flipping operation.

\vspace{5pt} \noindent \textit{- Albedo constancy loss is defined as:}
\begin{align*}
    L_{const}(\mathbf{A}) = \sum_{\mathbf{v}^{uv}_j \in \mathcal{N}_i} \omega(\mathbf{v}^{uv}_i, \mathbf{v}^{uv}_j) \lVert \mathbf{A}^{uv}(\mathbf{v}^{uv}_i) -\\ \mathbf{A}^{uv}(\mathbf{v}^{uv}_j) \rVert_2^p,
\end{align*}
where $\mathcal{N}_i$ denotes the 4-neighborhood around $\mathbf{v}^{uv}_i$ and the weight $\omega(\mathbf{v}^{uv}_i, \mathbf{v}^{uv}_j) = exp(-\alpha \lVert c(\mathbf{v}^{uv}_i) - c(\mathbf{v}^{uv}_j) \rVert)$ enforce that pixels with similar chromaticity should have similar albedo.


\subsection{Training Details}\label{subsec:training}
\noindent \textbf{3DMM Module:}
We train the 3DMM module in two stages. First, we train it using the 300W-3D dataset \cite{facealignzhu}, which has ground-truth shape, pose, texture and landmark annotations, for 100k iterations in a supervised way. Then, we further train it on the CelebA dataset \cite{celeba} with 1/10th of the original learning rate for further 30k iterations in an unsupervised way, whereby we use only the reconstruction loss, 2D landmark loss and the regularization losses. During this stage, we use landmark detections from HRNet \cite{hrnet} as groundtruth for the landmark loss. To make the 3DMM encoder robust to partial face images, we introduce artificial occlusions in the training images using random rectangular masks of varying sizes and locations. In addition, we also use random horizontal flipping as a data augmentation. During inference, occlusions are removed from the input image using the occlusion mask and passed through the 3DMM encoder to obtain occlusion-robust factorization.

\vspace{5pt}
\noindent \textbf{Albedo Inpainting Module:}
The albedo inpainting module is trained on the CelebA dataset \cite{celeba} for 30k iterations. To obtain the UV representations of the partial albedo and the mask, we re-project the 3D mesh obtained from the pretrained 3DMM module on the partial image and mask, respectively as shown in Main:Fig.\textcolor{red}{3.b}. On the GAN loss (Main:Eqn.\textcolor{red}{2}), we update the inpainter $\mathcal{G}$ and the discriminator $\mathcal{D}$ alternatively using a ratio of 1:1. On all the other completion losses, we update the inpainter $\mathcal{G}$ continuously. Other than the random face masks, we use random horizontal flipping as the only data augmentation to train the albedo inpainter.

\vspace{5pt}
\noindent \textbf{Face Segmentation Module:}
Since our method inpaints only the facial region in the UV domain, we restrict the image masks to lie on the face region too. For this, we train a UNet \cite{unet} based face segmentation model that separates the face region from the background, hair and inner mouth. The face segmenter predicts segmentation masks for (a) the face, (b) hair and other occlusions and (c) the background. We train the face segmentation module on the CelebAMask-HQ dataset \cite{CelebAMask-HQ} for a total of 50k iterations using the ground-truth annotations provided by the dataset. We use Focal loss \cite{focalloss} to train this module.

For all the modules, except the discriminator $\mathcal{D}$, we use the Adam optimizer with an initial learning rate of $10^{-4}$ and a step-decay of 0.98 per epoch, while for the PyramidGAN discriminator, we use an initial learning rate of $3\times 10^{-4}$. The input images are first aligned to $256\times256$ using the method suggested in \cite{CelebAMask-HQ}, which is the alignment used in the CelebA-HQ dataset. For training, we randomly crop the images to a size of $224\times224$ while during inference we use central crop. The full training takes 2 days on an Intel Xeon E5-2650 machine with two NVIDIA RTX 2080 GPUs, while inference takes 0.1 sec per image on a single GPU.

{\small
\bibliographystyle{ieee}
\bibliography{egbib}
}

\end{document}